\documentclass{article}
\usepackage{amc_arxiv_layout,times}

\usepackage{amsmath,amsfonts,bm}

\def\eqref#1{equation~\ref{#1}}

\def\1{\bm{1}}

\DeclareMathAlphabet{\mathsfit}{\encodingdefault}{\sfdefault}{m}{sl}
\SetMathAlphabet{\mathsfit}{bold}{\encodingdefault}{\sfdefault}{bx}{n}

\usepackage[T1]{fontenc}
\usepackage{amsmath,amssymb,amsthm}
\usepackage{graphicx,booktabs,multirow,array}
\usepackage[table]{xcolor}
\usepackage{flafter,placeins}
\usepackage{enumitem,pifont}
\usepackage{needspace}
\usepackage[nopatch=footnote]{microtype}
\usepackage[hidelinks]{hyperref}
\usepackage{url}
\usepackage[capitalize,noabbrev]{cleveref}
\usepackage[normalem]{ulem}
\hypersetup{
  pdftitle={How Many Futures Are Enough? Redundancy-Aware Effective Multimodality for End-to-End Autonomous Driving},
  pdfauthor={Yongzhi Liu, Xiao Yang, Yushu Qiu, Dian Zhou, Jinchang Xu, Zibo Ku, Dehui Chen, Guodong Yin, Weichao Zhuang}
}

\graphicspath{{figures/}}
\DeclareGraphicsExtensions{.jpg,.png}

\newtheorem{proposition}{Proposition}
\newcommand{\method}{AMC-Drive}
\newcommand{\keff}{K_{\mathrm{eff}}}

\newcommand{\tr}{\operatorname{tr}}
\newcommand{\cmark}{\ding{51}}

\definecolor{methodrow}{RGB}{245,243,248}
\definecolor{metriccol}{RGB}{238,238,238}
\newcolumntype{M}{>{\columncolor{metriccol}}r}
\newcolumntype{N}{>{\columncolor{metriccol}}c}
\setlist[itemize]{leftmargin=*,nosep}

\title{
How Many Futures Are Enough?\\
Redundancy-Aware Effective Multimodality \\
for End-to-End Autonomous Driving
}
\author{
\small\normalfont
Yongzhi Liu\textsuperscript{*}, Xiao Yang\textsuperscript{*}, Yushu Qiu,
Dian Zhou, Jinchang Xu\\
Zibo Ku, Dehui Chen, Guodong Yin,
Weichao Zhuang\textsuperscript{$\dagger$}\\
\footnotesize School of Mechanical Engineering, Southeast University\\
\footnotesize \textsuperscript{*}Equal contribution.\quad
\textsuperscript{$\dagger$}Corresponding author: \texttt{wezhuang@seu.edu.cn}
}
\date{}
\iclrfinalcopy
\begin{document}

\maketitle
\begin{abstract}

Multimodal end-to-end planners can generate multiple plausible trajectories
for the same driving scene, yet increasing the number of candidates does not
necessarily increase the diversity of represented futures.
Different trajectories may describe similar driving decisions and induce
similar future evolution, resulting in \emph{multimodal redundancy}.
We introduce \emph{effective multimodality} to distinguish the nominal
candidate count from the effective future structure represented by the set.
We propose \method{}
(\textbf{Action-conditioned Multimodal Consolidation for Driving}),
which uses a frozen action-conditioned predictive model to construct
candidate relations beyond trajectory geometry.
AMC-Drive calibrates the candidate structure by reducing redundant
representations while preserving distinct decision modes, without changing
the physical output cardinality.
On NAVSIM v1, \method{} achieves \textbf{93.9 PDMS} under the reported
protocol, and further obtains \textbf{89.2 EPDMS} on NAVSIM v2 and
\textbf{0.4308 HD-Score} on HUGSIM without HUGSIM-specific fine-tuning.
AMC is compatible with proposal-, diffusion-, and flow-matching-based
planners. A preliminary Push-T case study improves the manipulation score from
76.0 to 77.8, providing initial evidence beyond autonomous driving.
Demonstrations are available on the
\href{https://qyshu5356-ui.github.io/amc-drive-project-page/}{\uline{project page}},
and a partial NAVSIM v1 release is in the
\href{https://github.com/lumen2023/AMC-Drive}{\uline{code repository}}.
We plan to organize and release the remaining code and reproducibility
materials after acceptance for publication.

\end{abstract}

\section{Introduction}
\label{sec:introduction}

\begin{figure}[!htbp]
    \centering 
    \includegraphics[width=0.97\linewidth]{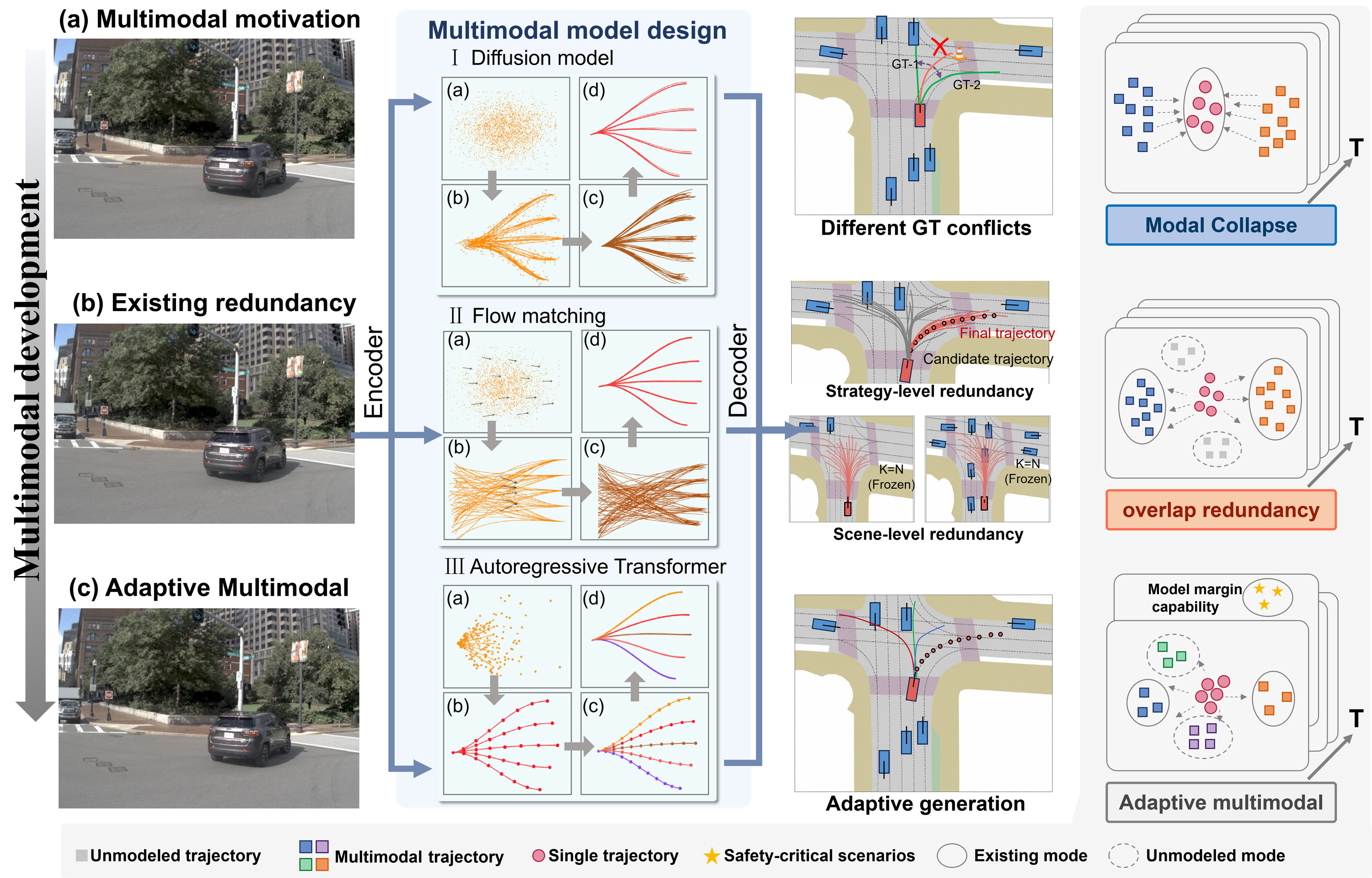}
    \caption{
    \textbf{From multiple trajectories to effective multimodality.}
    Modern multimodal planners can generate multiple plausible trajectories,
    yet their candidate sets may still exhibit mode collapse or redundant
    overlap.
    AMC-Drive reduces repeated modeling of closely related predictive futures
    and promotes a more effective organization of multimodal candidates.
    }
    \label{fig:intro_overview}
    \vspace{-2mm} 
\end{figure}

Autonomous driving involves inherently multimodal decision making and planning.
Given the same sensory observation and route intention, an ego vehicle may
admit multiple valid behaviors, including different speed profiles, path
geometries, yielding decisions, and interaction strategies.
Point-wise objectives such as $\ell_2$ regression tend to average these
valid futures and suppress distinct behavioral modes
\citep{phanminh2020covernet,chai2020multipath}.
Recent end-to-end planners therefore employ proposal-based, probabilistic,
diffusion, and flow-matching models to explicitly generate multiple
trajectories
\citep{li2024hydramdp,liao2025diffusiondrive,
chen2024vadv2,xing2025goalflow}.
Multimodal trajectory generation has consequently become a standard
capability of modern driving systems
(Fig.~\ref{fig:intro_overview}).

Across these formulations, multimodal planners typically generate a
predefined number of candidate trajectories for each scene, providing a
practical representation of multiple plausible driving behaviors.
The number of meaningful futures required by these candidates, however,
is inherently scene-dependent.
More importantly, generating $K$ different trajectories does not imply
that a planner represents $K$ distinct futures.
Multiple candidates can be numerically different while repeatedly
describing closely related driving decisions and inducing similar future
evolution.

This discrepancy motivates a more precise question:
\begin{center}
\setlength{\fboxsep}{10pt}
\fbox{
\parbox{0.82\linewidth}{
\centering
\vspace{1pt}
\emph{\textbf{
How much effective future structure is represented\\[3pt]
under a fixed candidate budget?
}}
\vspace{1pt}
}}
\end{center}

To formalize this distinction, we introduce
\emph{effective multimodality}, which separates the nominal candidate
count from the scene-dependent future structure represented by the candidate
set.
For a planner producing
$\mathcal{T}(s)=\{\tau_i\}_{i=1}^{K}$, we define
\begin{equation}
    K_{\mathrm{generated}} = K,
    \qquad
    K_{\mathrm{eff}}(s) \leq K .
\end{equation}
Here, $K$ remains fixed, while $K_{\mathrm{eff}}(s)$ characterizes the
effective future structure represented within this candidate budget.
We refer to repeated representation of closely related predictive futures
as \emph{multimodal redundancy}.
%
%
%
%
%

Identifying redundancy requires understanding candidate-specific future
dynamics beyond trajectory geometry. Action-conditioned predictive world
models provide such a representation by characterizing future evolution
under different candidate actions
\citep{ha2018worldmodels,assran2025vjepa2}.
We therefore instantiate a pretrained and frozen action-conditioned JEPA
interface (AC-JEPA), following the action-conditioned predictive formulation
of V-JEPA~2~\citep{assran2025vjepa2}, to encode each candidate action into a
predictive representation.
Based on these action-conditioned relations, we introduce
\textbf{AMC-Drive}
(\textbf{Action-conditioned Multimodal Consolidation for Driving}),
which identifies redundant predictive futures and calibrates the candidate
structure while preserving distinct decision modes.
AMC-Drive reduces repeated representation of near-equivalent futures
without changing the physical output cardinality, enabling more effective
use of a fixed multimodal candidate budget.

AMC-Drive achieves \textbf{93.9 PDMS} on NAVSIM v1,
\textbf{89.2 EPDMS} on NAVSIM v2
\citep{dauner2024navsim}, and
\textbf{0.4308 HD-Score} on HUGSIM
\citep{zhou2024hugsim} without HUGSIM-specific fine-tuning.
AMC-Drive is compatible with proposal-, diffusion-, and flow-matching-based
planners. A Push-T adaptation further improves the score from $76.0$ to
$77.8$, providing initial evidence that candidate-level redundancy
calibration can extend beyond autonomous driving.


\Needspace{9\baselineskip}
\paragraph{Contributions.}
Our contributions are threefold:

\begin{itemize}[leftmargin=*,topsep=4pt,itemsep=2pt,parsep=0pt]
    \item We introduce \emph{effective multimodality} and, to our knowledge,
    provide the first end-to-end driving framework that regularizes candidate
    redundancy during planner training through action-conditioned predictive
    representations.
    
    \item We propose an action-conditioned predictive representation framework
    that models future-aware relations among candidates through a frozen
    predictive world model, enabling redundancy identification beyond trajectory
    geometry.

    \item We develop a redundancy-aware optimization strategy with
    topology-aware mode correction and a cross-mode guard mechanism, which
    explicitly optimizes effective multimodality during end-to-end planner
    training.
    

\end{itemize}

\section{Related Work}
\label{sec:related}

\subsection{Multimodal Trajectory Generation and Planning}

Multimodal driving planners represent alternative behaviors through
multiple trajectory hypotheses or distributions over future trajectories.
Proposal-, anchor-, and probabilistic formulations provide explicit
mechanisms for maintaining competing driving hypotheses
\citep{phanminh2020covernet,chai2020multipath,breuer2021quovadis,
varadarajan2022multipathpp,gu2021densetnt,li2024hydramdp,chen2024vadv2}.
More recent generative planners directly model richer trajectory
distributions: diffusion-based methods generate diverse candidates through
iterative denoising, while flow-matching approaches provide continuous
generation with goal- or constraint-conditioned guidance
\citep{liao2025diffusiondrive,xing2025goalflow,liu2026guideflow}.

A complementary direction improves an existing candidate set through
refinement, scoring, and selection.
Proposal-centric planners iteratively update trajectory hypotheses or learn
to discriminate among competing candidates
\citep{guo2025ipad,yao2025drivesuprim}.
These advances have made multimodal trajectory generation a standard
component of modern end-to-end driving systems.

These prior systems establish that producing, selecting, and aggregating
multiple driving hypotheses is valuable, but they do not by themselves imply
that the hypotheses are non-redundant in a predictive or behavioral sense.
In particular, CoverNet and MultiPath-style methods organize a fixed set of
anchors or trajectory hypotheses, DenseTNT emphasizes dense goal coverage,
and Quo Vadis explicitly asks whether multiple hypotheses are meaningful
\citep{phanminh2020covernet,chai2020multipath,gu2021densetnt,
breuer2021quovadis,varadarajan2022multipathpp}.

\subsection{World Models and Predictive Representations for Driving}

World models learn predictive dynamics for future reasoning, planning, and
control
\citep{ha2018worldmodels,hafner2019planet,hafner2020dreamer,
hafner2023dreamerv3}.
In autonomous driving, this paradigm has been extended to model future
scene evolution from visual observations, latent states, and ego actions,
supporting planning through predicted observations or latent dynamics
\citep{hu2023gaia1,wang2023drivedreamer,
li2025law,zheng2025world4drive}.

Predictive learning has also shifted from raw-observation reconstruction
toward structured latent representations.
I-JEPA and V-JEPA establish joint-embedding prediction for images and
videos, while V-JEPA~2 further introduces action-conditioned prediction
for physical interaction and planning
\citep{assran2023ijepa,bardes2024vjepa,assran2025vjepa2}.
Recent methods such as DINO-WM, ACT-JEPA, Delta-JEPA, and WA-JEPA further
connect latent prediction with actions and future state transitions
\citep{zhou2024dinowm,vujinovic2025actjepa,
zhang2026deltajepa,wang2026wajepa}.
These works highlight the growing role of action-conditioned predictive
representations in relating alternative actions to future dynamics.

\subsection{Diversity and Structure of Multimodal Candidate Sets}

Beyond individual trajectory quality, multimodal planning depends on the
structure of the candidate set.
Prior works improve multimodal coverage and diversity through structured
sampling and set-level objectives, such as DPP-based formulations and
correlated latent sampling in DLow
\citep{kulesza2012dpp,yuan2020dlow}.
These studies establish diversity as a key property of multimodal
trajectory representations.

\label{sec:method}
\begin{figure}[t]
    \centering
    \includegraphics[width=0.95\linewidth]{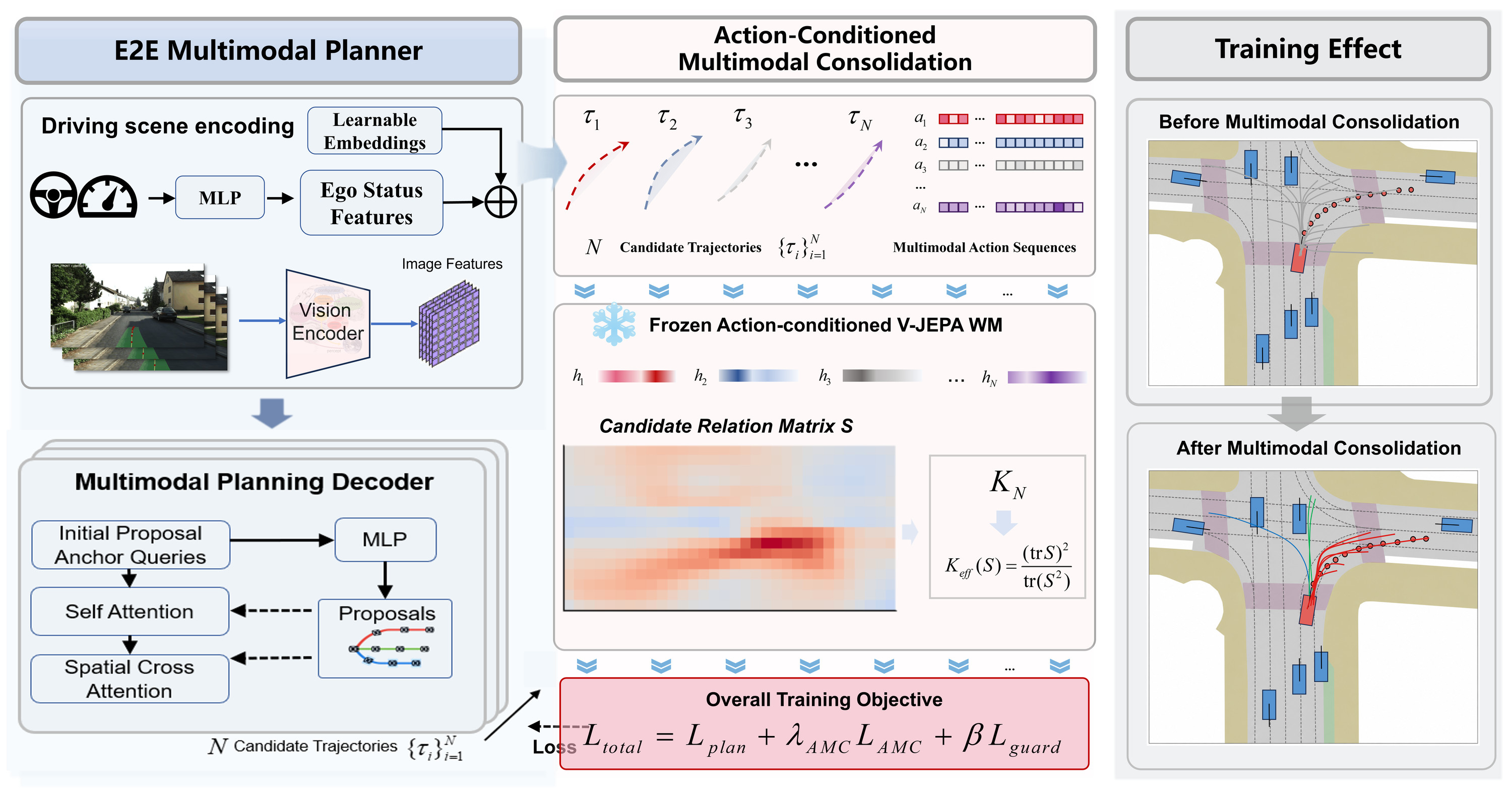}
    \caption{
    \textbf{AMC-Drive overview.}
    A frozen action-conditioned AC-JEPA provides candidate-level predictive
    relations for identifying multimodal redundancy.
    AMC-Drive optimizes effective multimodality by reducing repeated futures
    and preserving distinct modes under a fixed candidate budget.
    }
    \label{fig:architecture}
\end{figure}

Recent studies extend multimodal diversity beyond geometric separation toward
behavioral relevance and hypothesis organization.
Trajectory prediction methods such as CoverNet, MultiPath, and Trajectron++
improve future coverage through structured multimodal distributions
\citep{phanminh2020covernet,chai2020multipath,salzmann2020trajectron}.
Planning-oriented works further consider response-based diversity, where
LookOut distinguishes futures by induced ego responses
\citep{cui2021lookout}, while winner-take-all analyses reveal challenges
in mode partitioning and interpretation
\citep{wu2026rethinking}.
These studies shift attention from hypothesis quantity toward candidate
organization; however, redundancy among generated candidates and its impact
on effective multimodal representation remain underexplored.

\section{Method}

Given a driving scene $s$, a multimodal planner produces a fixed-size
candidate set $\mathcal{T}(s)=\{\tau_i\}_{i=1}^{K}$.
AMC-Drive keeps the output cardinality $K$ unchanged and operates only
during training to improve how these candidates organize multimodal futures.
A frozen action-conditioned predictive model provides candidate-specific
future representations, from which we characterize the effective
multimodality of the candidate set
\citep{assran2023ijepa,assran2025vjepa2}.
AMC-Drive then calibrates this structure toward a scene-dependent reference,
while a one-sided cross-mode guard prevents excessive contraction between
distinct decision modes.
The resulting objective augments the original planning loss without
changing the planner's inference interface. An overview of AMC-Drive is provided in
Fig.~\ref{fig:architecture}.

\subsection{Effective Multimodality}
\label{sec:effective_multimodality}

The number of generated trajectories specifies the cardinality of a
multimodal prediction, while leaving its internal structure unspecified.
Candidate sets with the same size $K$ can distribute their hypotheses very
differently across possible futures, ranging from highly overlapping
variations to well-separated alternatives.
This motivates a set-level measure of the effective multimodal structure
represented by a fixed candidate budget.

We represent pairwise candidate relations with a symmetric affinity matrix
$S\in[0,1]^{K\times K}$, where $S_{ij}$ denotes the relation strength
between trajectories $\tau_i$ and $\tau_j$, and $S_{ii}=1$.
Larger affinity indicates stronger relation between two candidates,
whereas smaller affinity reflects greater separation.
The construction of $S$ from action-conditioned predictive representations
is described in Sec.~\ref{sec:amc}.

We define the \emph{effective multimodality} of a candidate set by
\begin{equation}
    \boxed{
    K_{\mathrm{eff}}(S)
    =
    \frac{
        \left(\operatorname{tr}S\right)^2
    }{
        \operatorname{tr}\left(S^2\right)
    }}
    .
    \label{eq:keff}
\end{equation}
The definition admits a direct interpretation through its limiting cases.
For mutually separated candidates, $S=I$ and
$K_{\mathrm{eff}}=K$.
For maximally related candidates,
$S=\mathbf{1}\mathbf{1}^{\top}$ and
$K_{\mathrm{eff}}=1$.
Consequently,
\begin{equation}
    1 \leq K_{\mathrm{eff}}(S) \leq K .
    \label{eq:keff_bound}
\end{equation}

$K_{\mathrm{eff}}$ provides a continuous, permutation-invariant, and
differentiable characterization of candidate-set structure.
It does not require a discrete partition or a predefined number of
behavioral modes, which allows multimodal structure to vary continuously
across scenes.
We use $K_{\mathrm{eff}}$ as a descriptor of the organization induced by a
given candidate set, without assigning it the interpretation of a
ground-truth behavioral-mode count.
The remaining question is therefore how to construct candidate relations
that are informative for driving, which we address next using
action-conditioned predictive representations.

\subsection{Action-Conditioned Multimodal Consolidation}
\label{sec:amc}

\begin{figure}[t]
    \centering
    \includegraphics[
        width=0.62\linewidth,
        height=0.34\textheight,
        keepaspectratio
    ]{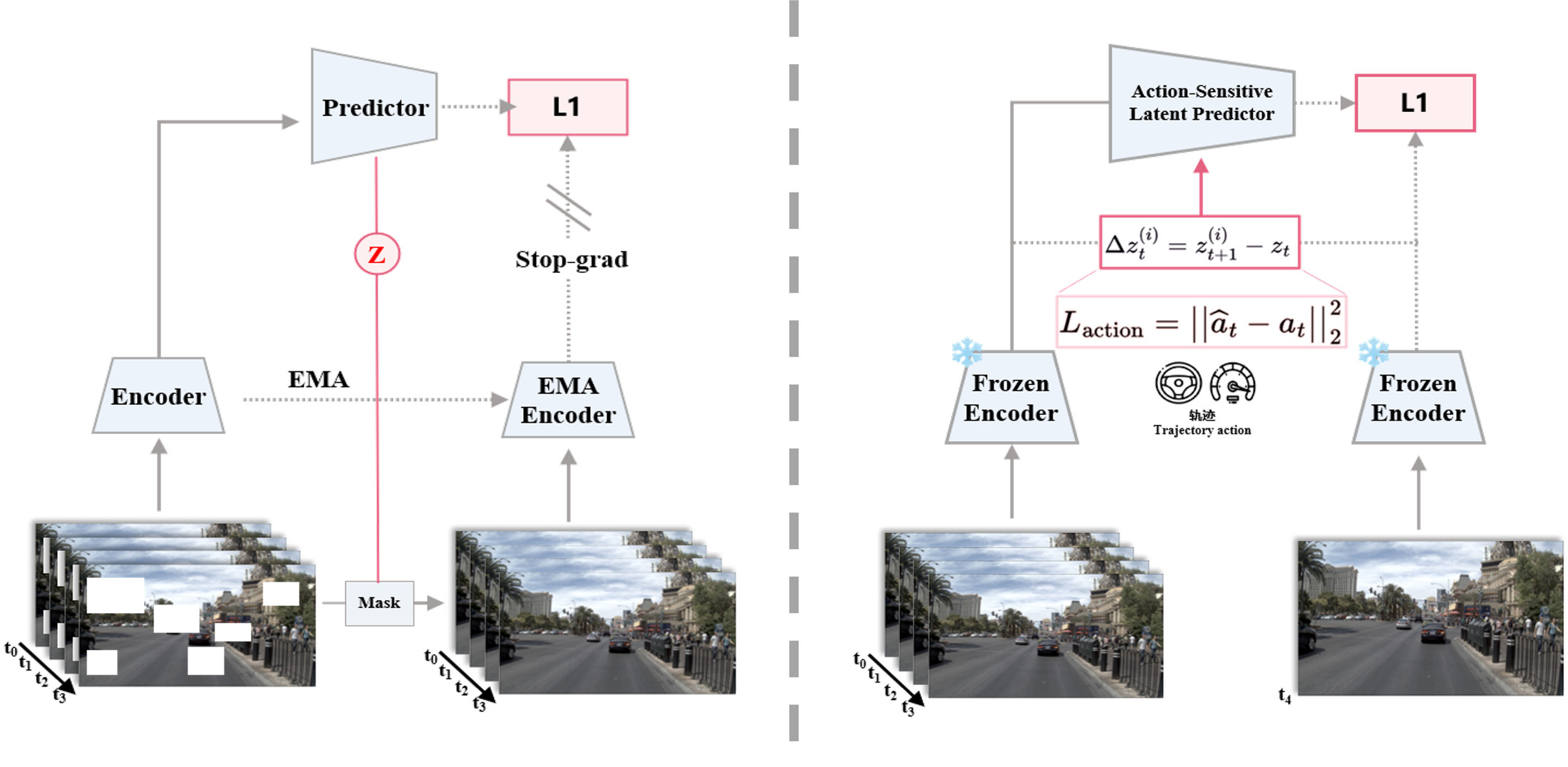}
    \caption{
\textbf{Action-conditioned predictive model.}
AC-JEPA extends JEPA prediction with action-conditioned future representations.
During AMC training, it is frozen to encode each candidate action into its
predictive future representation.
}
\vspace{-2mm} 
    \label{fig:acjepa}
\end{figure}

The effective multimodality defined in
Sec.~\ref{sec:effective_multimodality}
depends on the relations encoded by $S$.
To obtain relations that reflect the scene-conditioned effect of each
candidate, we associate trajectory $\tau_i$ with an action sequence
$a_i=\mathcal{A}(\tau_i)$ and evaluate it using a frozen
action-conditioned predictive model,
\begin{equation}
    z_i = F_{\mathrm{AC}}(s,a_i).
    \label{eq:acjepa_embedding}
\end{equation}
The parameters of $F_{\mathrm{AC}}$ remain fixed throughout AMC training,
while gradients through its input pathway are retained for planner
optimization.
For the ``+ Predictive'' control in Table~\ref{tab:ablation}, the same
planner-side feature interface is retained while
$\lambda_{\mathrm{AMC}}=\lambda_{\mathrm{guard}}=0$; consequently,
$\mathcal{L}_{\mathrm{plan}}$ still trains the planner, whereas AC-JEPA acts
only as a fixed pretrained feature extractor.
The predictive model is illustrated in
Fig.~\ref{fig:acjepa}. 

For a pair of candidates $(\tau_i,\tau_j)$, trajectory information enters
AMC only through the action mapping $A(\tau)$ and the frozen predictive
rollout. The relation distance $d_{ij}$ is computed solely in the resulting
action-conditioned predictive space.
The corresponding affinity is
\begin{equation}
    S_{ij}
    =
    \exp\left[
        -\left(
        \frac{d_{ij}}{\tau_s}
        \right)^2
    \right],
    \qquad
    S_{ii}=1 ,
    \label{eq:affinity}
\end{equation}
where $\tau_s$ controls the affinity scale.
This construction yields a continuous relation over the candidate set
without introducing a discrete mode assignment.
Implementation details of the relation function and feature normalization
are provided in the appendix.

Let $S^P$ be the planner affinity and $S^R$ the affinity of the
scene-indexed real-MTD teacher proposal set, both constructed through
the same frozen predictive model. Their effective multimodalities are
\begin{equation}
    K_{P}=K_{\mathrm{eff}}(S^{P}),
    \qquad
    K_{R}=K_{\mathrm{eff}}(S^{R}).
    \label{eq:planner_reference_keff}
\end{equation}
We match the planner to the detached teacher structure with
\[
\mathcal{L}_{\mathrm{AMC}}
=
\left(
\log K^P-\mathrm{sg}[\log K^R]
\right)^2,
\]
where $\mathrm{sg}[\cdot]$ denotes stop-gradient.
The log-domain objective measures the relative discrepancy in effective
multimodality and supports scene-dependent adjustment in either direction.

A set-level calibration alone does not determine how its correction should
be distributed among individual candidate pairs.
We therefore use a topology-aware realization of AMC, denoted TA-AMC,
which concentrates the optimization signal on strongly related candidates.
This selective credit assignment favors local consolidation of redundant
variations while limiting diffuse changes across the full candidate set.
The topology statistic and its exact optimization form are given in the
appendix.

TA-AMC determines where multimodal consolidation is applied,
but pairwise relation strength alone does not explicitly constrain
contraction across distinct decision modes.
Sec.~\ref{sec:cross_mode_guard} introduces a one-sided constraint for this
remaining case.
\subsection{Cross-Mode Guard and Training Objective}
\label{sec:cross_mode_guard}

The selective calibration in Sec.~\ref{sec:amc} concentrates structural
updates on strongly related candidates.
Pairwise relation strength, however, does not provide an explicit bound on
contraction across distinct decision modes.
We denote the cross-mode relation burden of the planner and the training
reference by $B_{\times}^{P}$ and $B_{\times}^{R}$, respectively.
The exact construction of the cross-mode relation set and its normalization
is provided in the appendix.

We use the reference burden to define a scene-conditioned upper bound,
\begin{equation}
    B_{\times}^{P}
    \leq
    B_{\times}^{R}.
    \label{eq:cross_bound}
\end{equation}
Let
\begin{equation}
    e_{\times}
    =
    \log B_{\times}^{P}
    -
    \log B_{\times}^{R}.
    \label{eq:cross_residual}
\end{equation}
The corresponding guard is implemented with a one-sided quadratic penalty,
\begin{equation}
    \boxed{
    \mathcal{L}_{\mathrm{guard}}
    =
    [e_{\times}]_{+}^{2}
    } ,
    \label{eq:cross_guard}
\end{equation}
where $[x]_{+}=\max(x,0)$.
When the planner remains within the reference bound,
$\mathcal{L}_{\mathrm{guard}}=0$;
the penalty becomes active only when the cross-mode burden exceeds this
level.
This formulation permits stronger mode separation when it already arises
from the planner, while constraining excessive cross-mode contraction.

The complete training objective combines the original planning objective
with multimodality calibration and cross-mode control:
\begin{equation}
    \boxed{
    \mathcal{L}
    =
    \mathcal{L}_{\mathrm{plan}}
    +
    \lambda_{\mathrm{AMC}}
    \mathcal{L}_{\mathrm{AMC}}
    +
    \lambda_{\mathrm{guard}}
    \mathcal{L}_{\mathrm{guard}}
    } .
    \label{eq:total_objective}
\end{equation}
Here, $\mathcal{L}_{\mathrm{plan}}$ retains the original task supervision,
$\mathcal{L}_{\mathrm{AMC}}$ calibrates the effective multimodal structure,
and $\mathcal{L}_{\mathrm{guard}}$ controls cross-mode contraction.
Together, these terms preserve the original planning objective while
regularizing the organization of its fixed-size multimodal candidate set.

\section{Experiments}
\label{sec:experiments}

\subsection{Experimental Setup}
\label{sec:experimental_setup}

We evaluate \method{} on NAVSIM v1, NAVSIM v2, and HUGSIM
\citep{dauner2024navsim,zhou2024hugsim}.
Our comparisons cover proposal-based, diffusion-based, and flow-matching
planners, with results grouped by comparable backbone configurations.
Metric definitions and the available implementation and provenance details are
provided in \cref{app:experiments}; fields without a verifiable run record are
marked explicitly there.
For the AMC results, we report the mean over three independent random-seed
runs. The three-seed aggregate remains above the matched comparator, including
for the small margins reported in the main tables. Thus, the comparisons are
not based on a single favorable run. We do not claim bootstrap
confidence intervals or formal scene-level paired significance in this version.

\begin{table}[t]
\caption{
Comparison with representative multimodal planners and
JEPA-based end-to-end driving methods on NAVSIM v1.
We report source-reported results under comparable standard-data
settings and exclude variants using additional training data or
test-time oracle selection.
}
\label{tab:navsim_v1_main}

\centering
\small
\setlength{\tabcolsep}{3.8pt}
\renewcommand{\arraystretch}{1.05}

\begin{tabular}{lccrrrrrM}
\toprule
Method
& Year
& Inputs
& NC $\uparrow$
& DAC $\uparrow$
& EP $\uparrow$
& C $\uparrow$
& TTC $\uparrow$
& PDMS $\uparrow$ \\
\midrule

\multicolumn{9}{l}{\textit{Multimodal end-to-end planners}} \\

DiffusionDrive~\citep{liao2025diffusiondrive}
& 2025
& C+L
& 98.2
& 96.2
& 82.2
& 100.0
& 94.7
& 88.1 \\

GoalFlow~\citep{xing2025goalflow}
& 2025
& C+L
& 98.4
& 98.3
& 85.0
& 100.0
& 94.6
& 90.3 \\

ReCogDrive~\citep{li2025recogdrive}
& 2025
& C
& 97.9
& 97.3
& 87.3
& 100.0
& 94.9
& 90.8 \\

iPad~\citep{guo2025ipad}
& 2025
& C
& 99.2
& 97.4
& 87.8
& 99.7
& 96.3
& 91.7 \\

DrivoR~\citep{kirby2026drivor}
& 2026
& C
& 98.9
& 98.3
& 89.1
& 100.0
& 96.2
& \underline{93.1} \\

DriveSuprim~\citep{yao2025drivesuprim}
& 2025
& C
& 98.6
& 98.6
& 91.3
& 100.0
& 95.5
& 93.5 \\

\midrule
\multicolumn{9}{l}{\textit{JEPA-based end-to-end driving}} \\

Auto-JEPA~\citep{yang2026autojepa}
& 2026
& C
& 98.4
& 98.3
& 87.1
& 100.0
& 95.0
& 91.3 \\

WA-JEPA~\citep{wang2026wajepa}
& 2026
& C
& 99.5
& 98.3
& 85.0
& 100.0
& 97.7
& 91.8 \\

Drive-JEPA~\citep{wang2026drivejepa}
& 2026
& C
& 99.1
& 98.2
& 91.4
& 99.9
& 96.2
& 93.3 \\

DA-WAM~\citep{zhong2026dawam}
& 2026
& C
& 99.1
& \textbf{98.9}
& 90.0
& 99.8
& 96.8
& 93.7 \\

Drive-HWM~\citep{fan2026drivehwm}
& 2026
& C
& \textbf{99.6}
& 99.0
& 89.0
& \textbf{100.0}
& 98.5
& \underline{93.8} \\

\rowcolor{methodrow}
\textbf{\method{}}
& \textbf{2026}
& \textbf{C}
& {99.1}
& \textbf{98.6}
& \textbf{92.2}
& \textbf{99.8}
& \textbf{95.8}
& \textbf{93.9} \\

\bottomrule
\end{tabular}

\par\vspace{3pt}
\begin{minipage}{0.94\linewidth}
\footnotesize
C denotes camera input and C+L camera and LiDAR.
Bold and underlined values denote the best and second-best results within each group;
results using additional training data or test-time oracle selection are excluded,
and Drive-HWM follows its main NAVSIM-v1 table. The \method{} row reports the
mean over three independent random-seed runs; the other rows retain their
source-reported values.
\end{minipage}

\end{table}

\begin{table}[t]
\caption{
Comparison with representative multimodal planners and
JEPA-based end-to-end driving methods on NAVSIM v2.
We report source-reported results under the original NAVSIM-v2
evaluation protocol for a consistent comparison.
}
\label{tab:navsim_v2_main}

\centering
\small
\setlength{\tabcolsep}{5.2pt}
\renewcommand{\arraystretch}{1.06}

\begin{tabular}{lcccM}
\toprule
Method
& Year
& Inputs
& Planning Paradigm
& EPDMS $\uparrow$ \\
\midrule

\multicolumn{5}{l}{\textit{Multimodal and end-to-end planners}} \\

Hydra-MDP++~\citep{li2025hydramdpp}
& 2025
& C
& Multimodal Distillation
& 85.6 \\

iPad~\citep{guo2025ipad}
& 2025
& C
& Proposal-based Planning
& 85.8 \\

DiffusionDriveV2~\citep{liao2025diffusiondrive}
& 2025
& C
& Diffusion Planning
& 85.5 \\

DriveSuprim~\citep{yao2025drivesuprim}
& 2025
& C
& Proposal / Selection
& 87.1 \\

\midrule
\multicolumn{5}{l}{\textit{JEPA-based end-to-end driving}} \\

Auto-JEPA~\citep{yang2026autojepa}
& 2026
& C
& Trajectory-latent JEPA
& 85.6 \\

Drive-HWM~\citep{fan2026drivehwm}
& 2026
& C
& Hierarchical World Modeling
& 86.4 \\

DA-WAM~\citep{zhong2026dawam}
& 2026
& C
& Decision-aligned Future Latents
& 87.7 \\

Drive-JEPA~\citep{wang2026drivejepa}
& 2026
& C
& V-JEPA + Proposal
& 87.8 \\

WA-JEPA~\citep{wang2026wajepa}
& 2026
& 4C
& V-JEPA World--Action Modeling
& \underline{88.0} \\

\rowcolor{methodrow}
\textbf{\method{}}
& \textbf{2026}
& \textbf{C}
& \textbf{AC-JEPA + AMC}
& \textbf{89.2} \\

\bottomrule
\end{tabular}

\par\vspace{3pt}
\begin{minipage}{0.94\linewidth}
\footnotesize
C denotes camera-only input, 4C four camera views, and C+L camera and LiDAR.
All EPDMS results use the original NAVSIM-v2 evaluator; bold and underlined
values indicate the best and second-best results within each group. The
\method{} row reports the mean over three independent random-seed runs; the
other rows retain their source-reported values.
\end{minipage}

\end{table}

\begin{table}[t]
\centering
\caption{Zero-shot closed-loop results on HUGSIM. Higher is better; bold marks the best value per row. \method{} is evaluated without HUGSIM-specific fine-tuning.}
\label{tab:hugsim}
\footnotesize
\setlength{\tabcolsep}{3pt}
\begin{tabular}{lrrrrrr}
\toprule
Metric & \cellcolor{methodrow}\method{}$^{\dagger}$ & WA-JEPA & LTF & DrivoR & UniAD & VAD \\
\midrule
\multicolumn{7}{l}{\textit{All scenarios}} \\
NC$\uparrow$ & \textbf{0.7258} & 0.6856 & 0.4428 & 0.5217 & 0.6555 & 0.4117 \\
DAC$\uparrow$ & 0.9578 & \textbf{0.9635} & 0.9275 & 0.9559 & 0.9320 & 0.9028 \\
TTC$\uparrow$ & \textbf{0.6672} & 0.6120 & 0.3751 & 0.4620 & 0.5156 & 0.2798 \\
Comf.$\uparrow$ & 0.8567 & 0.6620 & 0.9478 & 0.9390 & 0.6633 & \textbf{0.9534} \\
PDMS$\uparrow$ & \textbf{0.6374} & 0.5717 & 0.3653 & 0.4475 & 0.4940 & 0.2831 \\
RC$\uparrow$ & 0.5312 & \textbf{0.5689} & 0.3804 & 0.4721 & 0.4383 & 0.3006 \\
\rowcolor{metriccol}
HD-Score$\uparrow$ & 0.4308 & \textbf{0.4462} & 0.2310 & 0.3252 & 0.3124 & 0.1393 \\
\midrule
\multicolumn{7}{l}{\textit{HD-Score by difficulty}} \\
Easy & \cellcolor{methodrow}0.7777 & \textbf{0.7977} & 0.6608 & 0.7799 & 0.6395 & 0.4197 \\
Medium & \cellcolor{methodrow}0.5387 & \textbf{0.5563} & 0.1547 & 0.2911 & 0.3718 & 0.0849 \\
Hard & \cellcolor{methodrow}0.2629 & \textbf{0.3060} & 0.1204 & 0.2000 & 0.2099 & 0.0770 \\
Extreme & \cellcolor{methodrow}\textbf{0.1476} & 0.1362 & 0.1167 & 0.1407 & 0.0632 & 0.0626 \\
\bottomrule
\end{tabular}
\par\vspace{3pt}\begin{minipage}{\linewidth}\scriptsize $^{\dagger}$\method{} uses the NAVSIM-trained planner without HUGSIM-specific fine-tuning. Comparison methods follow WA-JEPA~\citep{wang2026wajepa}. Difficulty-split HD-Scores retain the previously logged local evaluation; the all-scenario HD-Score is $0.4308$.\end{minipage}
\end{table}

\paragraph{NAVSIM v1.}
NAVSIM v1 is derived from OpenScene and nuPlan and contains approximately
$103$k training scenarios (navtrain) and $12$k evaluation scenarios
(navtest)~\citep{dauner2024navsim}.
It evaluates planning with non-reactive pseudo-simulation under an
open-loop protocol, and we report the Predictive Driver Model Score
(PDMS) as the primary metric.

\paragraph{NAVSIM v2.}
NAVSIM v2 follows the same benchmark setting while extending the evaluation
criteria and reports the Extended PDM Score (EPDMS).
Compared with v1, EPDMS additionally accounts for driving-direction,
traffic-light, lane-keeping, and temporal-comfort criteria.
These protocol and metric extensions follow the NAVSIM benchmark definition
\citep{dauner2024navsim}.

\paragraph{HUGSIM.}
HUGSIM provides photorealistic closed-loop simulation for end-to-end
driving~\citep{zhou2024hugsim}.
We use HD-Score as the primary metric and report complementary driving
metrics in the corresponding table.
Unless otherwise stated, all HUGSIM results are obtained zero-shot,
without HUGSIM-specific fine-tuning.
\subsection{Main Results}
\label{sec:main_results}

We evaluate the complete \method{} across standard NAVSIM planning,
the revised NAVSIM v2 evaluation, and zero-shot closed-loop transfer
to HUGSIM.

Across NAVSIM v1 and v2, AMC-Drive consistently improves planning performance.
On NAVSIM v1, AMC-Drive achieves 93.9 PDMS with a ViT/L backbone,
compared with 93.3 for Drive-JEPA~\citep{wang2026drivejepa}.
The gain persists on NAVSIM v2, where AMC-Drive reaches 89.2 EPDMS
versus 87.8 for Drive-JEPA (Tables~1 and~2).
Without HUGSIM-specific fine-tuning, AMC-Drive obtains an HD-Score of
0.4308 under closed-loop evaluation (Table~3), providing evidence of
zero-shot transfer beyond NAVSIM.

\subsection{Mechanism and Ablation}
\label{sec:mechanism_ablation}

With a fixed candidate budget, effective multimodality still varies across
scenes, indicating that output cardinality alone does not characterize the
organization of the candidate set.
Our structural diagnostics further show that AMC reduces the discrepancy
between planner and reference effective multimodality, while the one-sided
guard suppresses excessive cross-mode violations without reducing candidate
coverage.
Qualitative scene-level diagnostics are provided in
\cref{app:mechanism_diagnostics}; the current evaluation does not include a
candidate-budget sweep or a stratified scene-type table of $K_{\mathrm{eff}}$.

\begin{table}[t]
\centering
\caption{\textbf{Ablation of \method{} modules on NAVSIM.}
Pred. Backbone denotes replacing the native backbone with the frozen
pretrained predictive backbone; $\mathcal{L}_K$, Topology, and Guard denote
the effective-capacity, topology-preservation, and semantic-preservation
objectives, respectively.}

\label{tab:ablation}
\small
\setlength{\tabcolsep}{4.5pt}
\renewcommand{\arraystretch}{1.05}
\begin{tabular}{lccccM}
\toprule
Config. & Pred. Backbone & $\mathcal{L}_K$ & Topology & Guard & PDMS $\uparrow$ \\
\midrule
Baseline & & & & & 91.7 \\
+ Predictive & \cmark & & & & \textbf{93.5} \\
+ Capacity & \cmark & \cmark & & & 89.2 \\
+ Topology & \cmark & \cmark & \cmark & & 91.6 \\
\rowcolor{methodrow}
+ Guard & \cmark & \cmark & \cmark & \cmark & \textbf{93.9} \\
\bottomrule
\end{tabular}

\par\vspace{3pt}
\begin{minipage}{0.96\linewidth}
\scriptsize
The first row is the original iPad baseline.
The ``+ Predictive'' row is a backbone-substitution control:
the native backbone is replaced by the frozen pretrained predictive backbone,
while only $\mathcal{L}_{\mathrm{plan}}$ is retained and all AMC objectives are disabled.
The remaining rows progressively introduce the AMC objectives.
Their three-seed means show that capacity consolidation alone degrades
planning, while topology preservation and Guard recover and improve performance.
\end{minipage}
\end{table}

The ablation in \cref{tab:ablation} clarifies why both selective calibration
and cross-mode protection are required.
The ``+ Predictive'' row is a representation-transfer control: it keeps the
original planning loss $\mathcal{L}_{\mathrm{plan}}$ active, uses the
pretrained AC-JEPA representation as an additional predictive feature, and
freezes the AC-JEPA parameters. Thus, its improvement from $91.7$ to $93.5$
does not come from $\mathcal{L}_K$, topology, or the guard; it measures the
information already present in the pretrained action-conditioned features.
Set-level AMC alone reduces PDMS from $93.5$ to $89.2$, showing that
calibrating global effective multimodality without controlling how the
correction is distributed can interfere with the planning objective.
Selective credit assignment recovers performance to $91.6$, and the
cross-mode guard further raises it to $93.9$.
The full design therefore couples scene-level calibration with localized
structural updates and explicit cross-mode protection.

\begin{figure}[!htbp]
  \centering
  \includegraphics[
    width=\linewidth,
    height=0.28\textheight,
    keepaspectratio
    ]{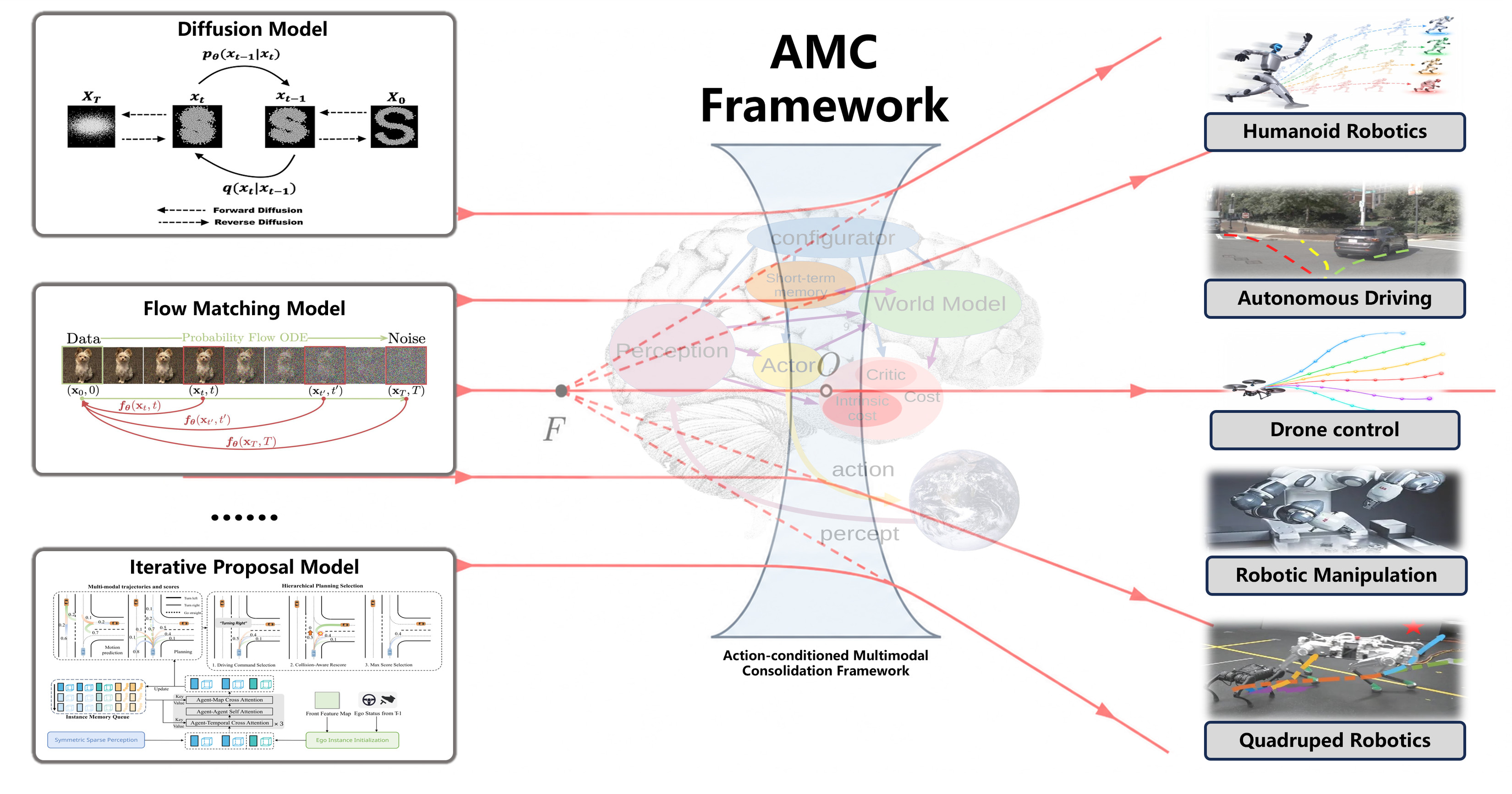}
  \caption{
  \textbf{Cross-generator portability and broader decision scope.}
  AMC provides a shared candidate-level calibration interface after candidate
  generation, applicable to diffusion-, flow-matching-, and proposal-based
  generators.
  Driving and manipulation are evaluated settings, while other robotics
  domains indicate potential extensions.
  }
  \vspace{-2mm} 
\label{fig:applications}
\end{figure}

\subsection{Generalization and Outlook}
\label{sec:generalization}

AMC operates on the generated candidate set rather than the internal
parameterization of a specific generator.
Therefore, it provides a unified candidate-level calibration interface
across diffusion-, flow-matching-, and proposal-based planners.
Experiments on DiffusionDrive, ReCogDrive, GoalFlow, and Drive-JEPA show
consistent improvements under their corresponding evaluation settings,
demonstrating that AMC is not tied to a specific generation paradigm.
Detailed implementation and provenance information are provided in
Appendix~\ref{app:transfer}.

As a preliminary cross-domain case study, we adapt AMC to Diffusion
Policy on Push-T; the reported score changes from 76.0 to 77.8.
Given its unverified paired provenance, we treat this only as initial
evidence beyond driving; details are provided in Appendix~C.3.
Figure~\ref{fig:applications} summarizes the evaluated settings and broader
potential applications.
Many multimodal decision problems face a finite candidate budget, where
the key challenge is not only generating alternatives but allocating
representative candidates among distinct futures.
These observations motivate broader evaluation of redundancy-aware
effective multimodality beyond driving.
Humanoid, aerial, and legged robotics are potential extensions beyond the
evaluated tasks.


\section{Conclusion}
\label{sec:conclusion}
Our results demonstrate the importance of effective multimodality under
a fixed candidate budget, AMC-Drive preserves output cardinality while
reducing intra-modal redundancy and retaining useful inter-modal distinctions.
AMC-Drive reaches 93.9 PDMS on NAVSIM v1, the highest reported PDMS
among the compared methods under the listed protocol, and attains
89.2 EPDMS on NAVSIM v2 and transfers zero-shot to HUGSIM, where it obtains
a 0.4308 HD-Score without HUGSIM-specific fine-tuning.
Across proposal-, diffusion-, and flow-based planners, AMC remains compatible
with different generator families. Push-T provides initial evidence beyond
driving, but does not establish broad cross-domain generality.

\label{page:main_end}
\clearpage
\bibliography{references}
\bibliographystyle{amc_arxiv_refs}
\clearpage
\appendix

\paragraph{AI use statement.}
Generative AI tools assisted with manuscript editing, LaTeX/table formatting,
reference checks, and code debugging. The authors verified all text, analyses,
and reported results and take responsibility for the final submission.

\paragraph{Reproducibility Statement.}
The public \method{} resources are available at:
\par\noindent Code: \url{https://github.com/lumen2023/AMC-Drive}\par
\noindent Project page: \url{https://qyshu5356-ui.github.io/amc-drive-project-page/}\par
The code repository currently provides a partial release: the NAVSIM v1
implementation, checkpoint access, submission script, and supporting evidence.
The project page hosts demonstrations and visualizations. Full release packages
for NAVSIM v2, HUGSIM, and cross-domain adapters are not yet available. We plan
to complete, organize, and publicly release the remaining implementation and
reproducibility materials after the paper is accepted for publication.
Benchmark protocols and provenance for the additional results are documented
in Appendix~\ref{app:experiments}.
\par\medskip
\noindent\textbf{Appendix organization.}
The supplementary material follows the order of the claims in the main text.
Appendix~\ref{app:theory} gives the effective-capacity and warm-start analysis;
Appendix~\ref{app:experiments} defines the benchmark protocols and records the
provenance of the reported numbers; Appendix~\ref{app:transfer} collects the
cross-generator and cross-domain transfer results; and
Appendix~\ref{app:qualitative} provides qualitative mechanism and
scene-level diagnostics. Qualitative panels are used to inspect observable
behavior and are not treated as substitutes for matched quantitative
evaluation.
\smallskip
\section{Analysis of Effective Capacity and Warm-Start Refinement}
\label{app:theory}
\subsection{Capacity bounds and occupancy}
\begin{proposition}\label{prop:capacity}
For a symmetric affinity matrix $S$ with unit diagonal and positive
semidefinite spectrum, the effective capacity satisfies
$1\leq K_{\mathrm{eff}}(S)\leq \operatorname{rank}(S)\leq K$.
If $S$ is composed of all-ones blocks with sizes $n_m$, then
$K_{\mathrm{eff}}(S)=K^2/\sum_m n_m^2$ and therefore
$K_{\mathrm{eff}}(S)\leq M$, with equality for equally occupied blocks.
\end{proposition}

For a nonzero positive-semidefinite $S$ with eigenvalues $\lambda_i\geq0$,
\begin{equation}
 1\leq\frac{(\sum_i\lambda_i)^2}{\sum_i\lambda_i^2}
 \leq\operatorname{rank}(S)\leq K.
\end{equation}
The lower bound follows by expanding the numerator; the upper bound is Cauchy--Schwarz on the nonzero eigenvalues. Symmetry and $S_{ii}=1$ give
$\tr(S^2)=\sum_{i,j}S_{ij}^2=K+2\sum_{i<j}S_{ij}^2$, proving \cref{prop:capacity}. If $S$ consists of all-ones blocks of sizes $n_m$, then $\tr(S^2)=\sum_m n_m^2$. Thus $\keff=K^2/\sum_m n_m^2\leq M$, with equality for equally occupied blocks.

\subsection{Relation to participation ratio and effective rank}
\label{app:keff_participation}

The quantity in Eq.~\ref{eq:keff} has the same spectral form as the
participation ratio (equivalently, the inverse participation ratio written in
its effective-number form). For a positive-semidefinite affinity matrix with
eigenvalues $\{\lambda_i\}$,
\begin{equation}
  K_{\mathrm{eff}}(S)
  = \frac{(\operatorname{tr}S)^2}{\operatorname{tr}(S^2)}
  = \frac{\left(\sum_i\lambda_i\right)^2}
         {\sum_i\lambda_i^2}.
  \label{eq:keff_participation_form}
\end{equation}
Thus, $K_{\mathrm{eff}}$ belongs to the broader family of effective-rank
measures, while remaining distinct from entropy-based effective-rank
definitions that use a different spectral functional. The bounds and the
block-occupancy calculation above are therefore standard consequences of this
spectral concentration form, rather than a claim that the ratio itself is a
new mathematical invention.

Our use of the ratio is specific to the relation structure in AMC-Drive. The
matrix $S$ is a scene-conditioned candidate-relation matrix: each entry is
induced by the distance between frozen action-conditioned predictive rollouts,
rather than by an arbitrary covariance matrix or an unconditioned similarity
kernel. In the ideal block case, $K_{\mathrm{eff}}=K^2/\sum_m n_m^2$ is an
occupancy-weighted effective number of relation groups; it should not be read
as a ground-truth count of semantic driving modes. This interpretation also
explains why unequal block occupancy lowers $K_{\mathrm{eff}}$ even when the
number of blocks is unchanged.

The methodological contribution is to use this scene-conditioned statistic as
a differentiable planner regularizer. In particular, the student statistic
$K_{\mathrm{eff}}^{P}$ is matched to the detached reference statistic
$K_{\mathrm{eff}}^{R}$ through
\begin{equation}
  L_K=\left(\log K_{\mathrm{eff}}^{P}
  -\operatorname{sg}\!\left[\log K_{\mathrm{eff}}^{R}\right]\right)^2,
  \label{eq:keff_regularizer_appendix}
\end{equation}
so gradients flow through the student relation matrix and its frozen
action-conditioned predictive model to the planner parameters, but not into the
reference branch or the predictive-model parameters. The novelty claim is
therefore about this scene-conditioned, action-conditioned use and its
integration into a differentiable planner objective, not about inventing the
participation-ratio formula itself.

\subsection{Equal capacity, different edge concentration}
Consider
\begin{equation}
 S_A=\begin{pmatrix}1&1/2&0\\1/2&1&0\\0&0&1\end{pmatrix},\qquad
 S_B=\begin{pmatrix}1&r&r\\r&1&r\\r&r&1\end{pmatrix},\qquad r=1/\sqrt{12}.
\end{equation}
Both are positive definite. Their total squared off-diagonal edge weight is $1/4$, giving $\keff=9/(3+1/2)=18/7$ in both cases. However, $S_A$ has one nonzero edge, with $E_{\mathrm{eff}}=1$ and $\rho=1/3$, while $S_B$ has three equally weighted edges, with $E_{\mathrm{eff}}=3$ and $\rho=1$. This proves non-identifiability for the stated class of affinity matrices. The zero entries are idealized; the example does not assert exact realizability by a finite-bandwidth Gaussian kernel.

\subsection{Meaning of the semantic guard}
For positive stabilized masses, $[e]_+^2=0$ if and only if $B_\times(S)\leq B_\times(\bar S)$. Hence arbitrarily negative log-ratios satisfy the constraint; equality error is not the matching estimand. Moreover, this is an aggregate constraint. Increased similarity on one cross-mode pair can be offset by a decrease elsewhere. Pairwise false-merge rates and task safety metrics remain necessary diagnostics.

\subsection{Proof of the local warm-start result}
Assume $L_0,R$ are three times continuously differentiable in a neighborhood of $\theta_0$. Define
\begin{equation}
 F(\theta,\epsilon)=\nabla L_0(\theta)+\epsilon\nabla R(\theta)+\mu(\theta-\theta_0).
\end{equation}
At $(\theta_0,0)$, $F=0$ and $\partial F/\partial\theta=H_\mu\succ0$. The implicit function theorem gives a local smooth stationary branch. Differentiating $F(\theta_\epsilon,\epsilon)=0$ at zero yields
$\theta'_0=-H_\mu^{-1}g$, where $g=\nabla R(\theta_0)$. Taylor expansion gives
\begin{equation}
  \theta_\epsilon=\theta_0-\epsilon H_\mu^{-1}g+O(\epsilon^2),
  \label{eq:warmstart}
\end{equation}
and
\begin{align}
 R(\theta_\epsilon)-R(\theta_0)&=-\epsilon g^\top H_\mu^{-1}g+O(\epsilon^2),\\
 L_0(\theta_\epsilon)-L_0(\theta_0)&=
 \frac{\epsilon^2}{2}g^\top H_\mu^{-1}\nabla^2L_0(\theta_0)H_\mu^{-1}g+O(\epsilon^3).
\end{align}
Positive definiteness implies $g^\top H_\mu^{-1}g>0$ when $g\neq0$. The result concerns an ideal local stationary solution. If the initial baseline is not stationary, a first-order change in $L_0$ need not vanish. Assignment boundaries and the hinge-square kink also require a locally smooth region or a smooth surrogate.

\section{Experimental Details and Data Provenance}
\label{app:experiments}

This appendix restores the experimental information omitted from
Sec.~\ref{sec:experimental_setup}. We separate benchmark definitions from
reported values and from implementation details. Numerical results are not
recomputed here.

\paragraph{Implementation availability.}
The code repository provides a partial \method{} release, including the NAVSIM
v1 implementation, checkpoint access, submission script, and supporting
evidence:
\par\noindent\url{https://github.com/lumen2023/AMC-Drive}\par
The project page provides demonstrations and visualizations:
\par\noindent\url{https://qyshu5356-ui.github.io/amc-drive-project-page/}\par
Full release packages for NAVSIM v2, HUGSIM, and the cross-domain adapters are
not yet included. We plan to complete, organize, and release the remaining
implementation and reproducibility materials after acceptance for publication.
This appendix records the benchmark protocols, implementation choices, and
evidence boundaries needed to interpret the reported results.

\subsection{Benchmark Protocols and Metrics}
\label{app:benchmark_metrics}

\paragraph{NAVSIM v1.}
NAVSIM v1 is derived from OpenScene and nuPlan and uses approximately
$103$k training scenarios (\textit{navtrain}) and $12$k evaluation scenarios
(\textit{navtest})~\citep{dauner2024navsim}. Evaluation uses non-reactive
pseudo-simulation under an open-loop protocol. The official evaluation
computes the subscores after a 4-second non-reactive simulation and aggregates
them into the Predictive Driver Model Score (PDMS). The subscores used here are
No at-fault Collision (NC), Drivable Area Compliance (DAC), Ego Progress (EP),
Time to Collision within bounds (TTC), and Comfort (C):

\begin{equation}
    \mathrm{PDMS}
    = \mathrm{NC}\times\mathrm{DAC}\times
    \frac{5\,\mathrm{EP}+5\,\mathrm{TTC}+2\,\mathrm{C}}{12}.
    \label{eq:pdms_appendix}
\end{equation}

NC and DAC are multiplicative penalty terms. EP, TTC, and C form the weighted
average in Eq.~\ref{eq:pdms_appendix}. The main-text NAVSIM v1 table reports
PDMS as its primary metric; the subscore columns retain the labels supplied by
the source table.

\paragraph{NAVSIM v2.}
NAVSIM v2 keeps the same data-split philosophy while extending the evaluation
criteria. Relative to v1, the Extended Predictive Driver Model Score (EPDMS)
adds Driving Direction Compliance (DDC), Traffic Light Compliance (TLC;
reported as \texttt{TL} in the table), Lane Keeping (LK), History Comfort
(HC), and Extended Comfort (EC). DDC and TLC are multiplier terms. EP and TTC
retain weight $5$, while LK, HC, and EC each have weight $2$ in the weighted
part. The corresponding benchmark definition is

\begin{equation}
\begin{aligned}
\Pi_{\mathrm{v2}}
&=
\prod_{m\in\{\mathrm{NC},\mathrm{DAC},\mathrm{DDC},\mathrm{TLC}\}}
\operatorname{filter}_{m},\\
S_{\mathrm{v2}}
&=
5\,\operatorname{filter}_{\mathrm{EP}}
+5\,\operatorname{filter}_{\mathrm{TTC}}
+2\,\operatorname{filter}_{\mathrm{LK}}\\
&\quad
+2\,\operatorname{filter}_{\mathrm{HC}}
+2\,\operatorname{filter}_{\mathrm{EC}},\\
\mathrm{EPDMS}
&=
\Pi_{\mathrm{v2}}\,\frac{S_{\mathrm{v2}}}{16}.
\end{aligned}
\label{eq:epdms_appendix}
\end{equation}

where the benchmark's human-reference filter sets a metric to $1$ when the
human reference receives a zero score for that metric; otherwise it retains the
agent score. DDC checks legal driving direction, TLC checks traffic-light
compliance, LK penalizes sustained deviation from the lane centerline (and is
disabled at intersections), HC evaluates continuity with recent motion history,
and EC evaluates consistency across consecutive plans and their dynamic states.

The NAVSIM v2 table reports EPDMS values from the original NAVSIM-v2
evaluation implementation, before the corrected human-reference
penalty-filter aggregation. The table footnote records this implementation
version. We retain the reported values and do not mix them with values
recomputed using the corrected evaluator associated with
Eq.~\ref{eq:epdms_appendix}.

\paragraph{HUGSIM.}
HUGSIM is a photorealistic closed-loop simulator for end-to-end driving
\citep{zhou2024hugsim}. The reported table contains NC, DAC, TTC, Comfort
(COM), PDMS, Route Completion (RC), and HD-Score. In the HUGSIM benchmark,
NC denotes no collision, DAC denotes drivable-area compliance, TTC denotes
time-to-collision compliance, COM denotes comfort, and RC is the fraction of
the route completed. The per-timestamp HD-Score is formed from NC and DAC as
multiplicative driving-policy terms and TTC and COM as contributory terms:

\begin{equation}
\begin{aligned}
\mathrm{HD\text{-}Score}_{t}
&= \mathrm{NC}_{t}\times\mathrm{DAC}_{t}\times
\frac{w_{\mathrm{TTC}}\mathrm{TTC}_{t}
      +w_{\mathrm{COM}}\mathrm{COM}_{t}}
     {w_{\mathrm{TTC}}+w_{\mathrm{COM}}},\\
\mathrm{HD\text{-}Score}
&=\mathrm{RC}\times\frac{1}{T}\sum_{t=1}^{T}
\mathrm{HD\text{-}Score}_{t}.
\end{aligned}
\label{eq:hd_score_appendix}
\end{equation}

The main-text HUGSIM results use a zero-shot protocol: the NAVSIM-trained
planner is evaluated without HUGSIM-specific fine-tuning, as stated in the
table footnote. The table's auxiliary PDMS column is retained from the supplied
local result file; it is not treated as an alternative definition of
HD-Score.

\paragraph{HUGSIM scene-count provenance.}
The supplied local baseline contains 80 easy, 149 medium, 94 hard, and 99
extreme scenes, for 422 scenes in total. The comparison protocol lists 80, 157,
96, and 103 scenes, for 436 scenes. Because these scene sets differ, the
all-scenario values are retained as supplied rather than reconstructed from
rounded difficulty means. The descriptive difference table uses the supplied
rounded values and uses WA-JEPA as the denominator for relative differences.

\paragraph{HUGSIM comparison reproducibility and scene-set reconciliation.}
The scene count is a provenance check and is kept separate from the evaluation
implementation. The supplied AMC and WA-JEPA runs use the same HUGSIM
evaluation standard and configuration. In a matched rerun, both methods are
given the identical ordered list of scene IDs and the same scene filtering,
simulator configuration, evaluator implementation, input preprocessing,
trajectory horizon, and metric aggregation. We regard a comparison as
scene-matched only when the two manifests have identical IDs and cardinality.
Accordingly, the 422-scene local log and the 436-scene comparison protocol are
reported with their original provenance and are not presented as a scene-ID
matched head-to-head result. The HUGSIM difference table below is therefore
descriptive; it does not support a causal attribution of each metric difference
to AMC. The common-manifest protocol is the one to use for any exact rerun or
future numerical update.
\begin{table}[t]
\centering
\caption{
Descriptive local-baseline differences relative to WA-JEPA on HUGSIM. Evaluation scene sets differ.}
\label{tab:hugsim_delta}
\footnotesize
\setlength{\tabcolsep}{3pt}
\begin{tabular}{lrrrr}
\toprule
Metric & \cellcolor{methodrow}Local baseline & WA-JEPA & Absolute difference & Relative difference \\
\midrule
NC$\uparrow$ & 0.7258 & 0.6856 & +0.0402 & +5.9\% \\
DAC$\uparrow$ & 0.9578 & 0.9635 & -0.0057 & -0.6\% \\
TTC$\uparrow$ & 0.6672 & 0.6120 & +0.0552 & +9.0\% \\
Comf.$\uparrow$ & 0.8567 & 0.6620 & +0.1947 & +29.4\% \\
PDMS$\uparrow$ & 0.6374 & 0.5717 & +0.0657 & +11.5\% \\
RC$\uparrow$ & 0.5312 & 0.5689 & -0.0377 & -6.6\% \\
\rowcolor{metriccol}
HD-Score$\uparrow$ & 0.4308 & 0.4462 & -0.0154 & -3.5\% \\
\bottomrule
\end{tabular}
\end{table}

\subsection{Table~4: Predictive-only Control and Gradient Path}
\label{app:predictive_ablation}

The ``+ Predictive'' entry in Table~\ref{tab:ablation} is a
predictive-representation control, not an AMC run. In this control, the
planner's native end-to-end image backbone (either its ViT or ResNet variant)
is replaced by the ViT encoder from the pretrained JEPA-v2.1 model. The
JEPA-v2.1 encoder is loaded after pretraining and kept frozen throughout the
ablation; it supplies the visual representation to the otherwise trainable
planning head and trajectory decoder. Thus, the ``Predictive Rep.'' column
denotes a backbone substitution, rather than an additional AMC objective or
a post-hoc feature concatenation.
The phrase ``additional predictive feature'' used as shorthand in the main
text refers to making this predictive representation available to the planning
head; it does not mean concatenating it with a simultaneously retained native
ViT/ResNet stream.

For this row, no candidate-relation calibration branch is instantiated. The
training objective is therefore only the original planning objective,
\begin{equation}
  \mathcal{L}_{\mathrm{total}}^{+\,\mathrm{Predictive}}
  = \mathcal{L}_{\mathrm{plan}},
  \qquad
  \lambda_{\mathrm{AMC}}=0,
  \qquad
  \lambda_{\mathrm{guard}}=0,
\end{equation}
and the capacity, topology, and Guard terms are disabled. The corresponding
forward and backward path is
\begin{equation}
  I
  \xrightarrow{\;E_{\mathrm{JEPA\text{-}v2.1}}^{\mathrm{frozen}}\;}
  h
  \xrightarrow{\;G_{\theta}\;}
  \hat{\tau}
  \xrightarrow{\;\mathcal{L}_{\mathrm{plan}}\;}
  \nabla_{\theta}\mathcal{L}_{\mathrm{plan}},
  \qquad
  \nabla_{\phi_{\mathrm{JEPA}}}\mathcal{L}_{\mathrm{plan}}=0.
  \label{eq:predictive_only_gradient_path}
\end{equation}
Here $E_{\mathrm{JEPA\text{-}v2.1}}$ is the frozen encoder and
$G_{\theta}$ denotes the trainable planning head/trajectory decoder (including
any trainable adapter after the encoder). Consequently, the planning loss
still provides a non-zero gradient to the trainable planner parameters even
though no gradient updates the JEPA-v2.1 encoder. In particular, there is no
AMC gradient path through $\mathcal{L}_{K}$, topology, or the cross-mode Guard
because those branches are absent in this row.

The reported $93.5$ is the NAVSIM-v1 PDMS obtained by training and evaluating
this frozen-encoder, planning-loss-only configuration under the same benchmark
protocol as the $91.7$ baseline. It is an endpoint score of the ablation, not
the result of adding a fixed ``Predictive'' increment to $91.7$.
The subsequent ``+ Capacity'', ``+ Topology'', and
``+ Guard'' rows retain the frozen JEPA-v2.1 encoder and add their respective
AMC terms progressively.
For comparison, the baseline uses the planner's native image backbone in the
same planning-loss-only path, with that native backbone trainable; the
backbone replacement and freezing are the only intended changes in the
``+ Predictive'' control before AMC terms are enabled.

\subsection{AMC Relation and Calibration Details}
\label{app:amc_relation_details}

\paragraph{Distance and affinity.}
\paragraph{Distance and affinity.}
Trajectory $\tau_i$ enters the relation construction only through
$a_i=A(\tau_i)$. Let $z_{i,h}$ denote the frozen AC-JEPA predictive tensor
at rollout horizon $h$.
With $H=2$ and $D$ the number of scalars in one flattened horizon tensor, the relation distance is
\begin{equation}
q_{ij}=\frac{1}{H}\sum_{h=1}^{H}\frac{1}{D}
\left\|\operatorname{vec}(z_{i,h})-\operatorname{vec}(z_{j,h})\right\|_2^2,
\qquad d_{ij}=\sqrt{q_{ij}}.
\label{eq:amc_pilot_distance}
\end{equation}
The code computes $q_{ij}$ by a nonnegative-clamped Gram-matrix identity;
the square-root notation makes Eq.~\ref{eq:affinity} in the main text
identical to $S_{ij}=\exp(-q_{ij}/\tau_s^2)$, with
$\tau_s=0.5014539957046509$ in this mechanism implementation. The matrix is
symmetrized and its diagonal set to one. There is no additive waypoint-geometry
term or separately tuned geometry/latent mixture weight in this
affinity: trajectory geometry enters through $\mathcal A(\tau)$ and the
action-conditioned rollout. The $1/D$ and $1/H$ factors are its explicit
feature and horizon normalizations.

\paragraph{Topology objective.}
For either planner ($P$) or reference ($R$), let $N$ be its trajectory count
and $\mathcal E_N=\{(i,j):1\leq i<j\leq N\}$. All strictly upper-triangular
pairs are included; there is no additional topology mask or learned pair
weight. The affinity itself supplies the selective weight:
\begin{equation}
w_{ij}=S_{ij}^{2},\quad A=\sum_{\mathcal E_N}w_{ij},\quad
C=\max\!\left(\sum_{\mathcal E_N}w_{ij}^{2},10^{-12}\right),\quad
K_{\rm eff}=\frac{N^2}{N+2A},\quad
\rho=\frac{A^2}{C\binom N2}.
\label{eq:amc_pilot_topology_stat}
\end{equation}
The scene losses are
\begin{equation}
L_K=\bigl(\log K_{\rm eff}^{P}-\operatorname{sg}[\log K_{\rm eff}^{R}]\bigr)^2,
\qquad
L_{\rm topo}=\left[\log\rho^{P}-\operatorname{sg}[\log\rho^{R}]\right]_{+}^{2}.
\label{eq:amc_pilot_topology_loss}
\end{equation}
The implementation clamps $K_{\rm eff}$ below at $10^{-6}$ and $\rho$ below
at $10^{-12}$ before taking logarithms. A reference set with $N<2$ has no
topology target and is excluded.

\paragraph{Cross-mode Guard.}
Each of the $M\geq2$ real-MTD teacher trajectories $T_m$ is treated as one
reference mode. For a planner proposal $\tau_i$, define the mean position error
to teacher mode $m$ over the same eight 0.5-s waypoints by
\begin{equation}
c_{im}=\frac18\sum_{t=1}^{8}
\left\|(x_{it},y_{it})-(x^R_{mt},y^R_{mt})\right\|_2,\quad
y_i=\arg\min_m c_{im},\quad
e_i=\mathbf 1\!\left[c_{i,(2)}-c_{i,(1)}\geq0.5\,\mathrm m\right],
\label{eq:amc_pilot_assignment}
\end{equation}
where $c_{i,(1)}$ and $c_{i,(2)}$ are the two smallest errors; ties use the
first minimum. The assignment and eligibility mask are detached from the
proposal-gradient path. For $i<j$, let
$V_{ij}=e_i e_j\mathbf 1[y_i\ne y_j]$ and $N_{\times}^{P}=\sum_{i<j}V_{ij}$.
Every distinct teacher pair is cross-mode, so $N_{\times}^{R}=\binom M2$.
The code compares \emph{per-edge} squared-affinity burdens,
\begin{equation}
B_{\times}^{P}=\frac{\sum_{i<j}V_{ij}(S_{ij}^{P})^2}{N_{\times}^{P}},
\qquad
B_{\times}^{R}=\frac{\sum_{a<b}(S_{ab}^{R})^2}{\binom M2},
\label{eq:amc_pilot_cross_burden}
\end{equation}
when $N_{\times}^{P}>0$; a scene without eligible cross edges is skipped.
The zero-margin one-sided penalty is
\begin{equation}
L_{\times}=\left[\log(B_{\times}^{P}+10^{-8})-
\operatorname{sg}\{\log(B_{\times}^{R}+10^{-8})\}\right]_{+}^{2}.
\label{eq:amc_pilot_cross_loss}
\end{equation}
The additive $10^{-8}$ is a numerical stabilizer for the main-text
log-burden expression.

\paragraph{Reference construction.}
The reference is the scene-indexed \texttt{real\_target\_trajectories} array
from the real-MTD teacher proposal cache, not the single ground-truth future,
an historical planner checkpoint, or a hand-specified affinity. For each
scene, the reference path reads its cached $M\times8\times3$ teacher
poses and the corresponding cached AC-JEPA context. Under no gradient, it
applies exactly the student action adapter, train-split action normalization,
frozen two-horizon AC-JEPA rollout, and Eq.~\ref{eq:amc_pilot_distance} to
construct $S^R$, $K_{\rm eff}^R$, and $\rho^R$. These teacher topology scalars
are computed once before the pilot update loop and detached in the losses.
The teacher cross-pair affinity sum is precomputed and divided by
$\binom M2$ in Eq.~\ref{eq:amc_pilot_cross_burden};
it is not recomputed from student proposals at each update. Student affinity
is recomputed online each step and remains differentiable through the frozen
predictive model to the planner.

\paragraph{Trajectory-to-action conversion.}
Write the eight future poses in the current ego frame as
$p_t=(x_t,y_t,\psi_t)$ at times $t/2$ seconds, $t=1,\ldots,8$, and prepend
$p_0=(0,0,0)$. The deterministic adapter produces adjacent local-frame
$\mathrm{SE}(2)$ increments, not steering/throttle or curvature controls:
\begin{equation}
\mathcal A(\tau)_t=
\begin{pmatrix}
\cos\psi_{t-1}\,\Delta x_t+\sin\psi_{t-1}\,\Delta y_t\\
-\sin\psi_{t-1}\,\Delta x_t+\cos\psi_{t-1}\,\Delta y_t\\
\operatorname{wrap}_{[-\pi,\pi)}(\psi_t-\psi_{t-1})
\end{pmatrix},\quad
(\Delta x_t,\Delta y_t)=(x_t-x_{t-1},y_t-y_{t-1}).
\label{eq:amc_pilot_action}
\end{equation}
For the stride-one AC-JEPA checkpoint, each action channel is normalized as
$(a_t-\mu_1)/\max(\sigma_1,10^{-6})$, where $\mu_1,\sigma_1$ are computed
channelwise over the training-split action manifest by
\texttt{action\_stats.py}; no validation/test trajectories enter these
statistics. The adapter emits eight actions, of which the first two 0.5-s
actions condition the predictive rollout.

\subsection{Audited AMC Mechanism Pilot}
\label{app:amc_mechanism_pilot}

The P0 evidence package contains one internally consistent, density-normalized
teacher-conditioned configuration for auditing the AMC mechanism. It uses
Adam with learning rate $10^{-5}$, 48 optimizer steps over eight fixed scenes,
$(\lambda_K,\lambda_{\rm topo},\lambda_{\times})=(1,0.1,0.1)$, two frozen
AC-JEPA rollout horizons, and a fixed candidate budget of 32. The pilot is
intentionally reported as a mechanism check rather than as full planner
training or a leaderboard result.

\begin{figure}[!htbp]
  \centering
  \includegraphics[width=0.94\linewidth]{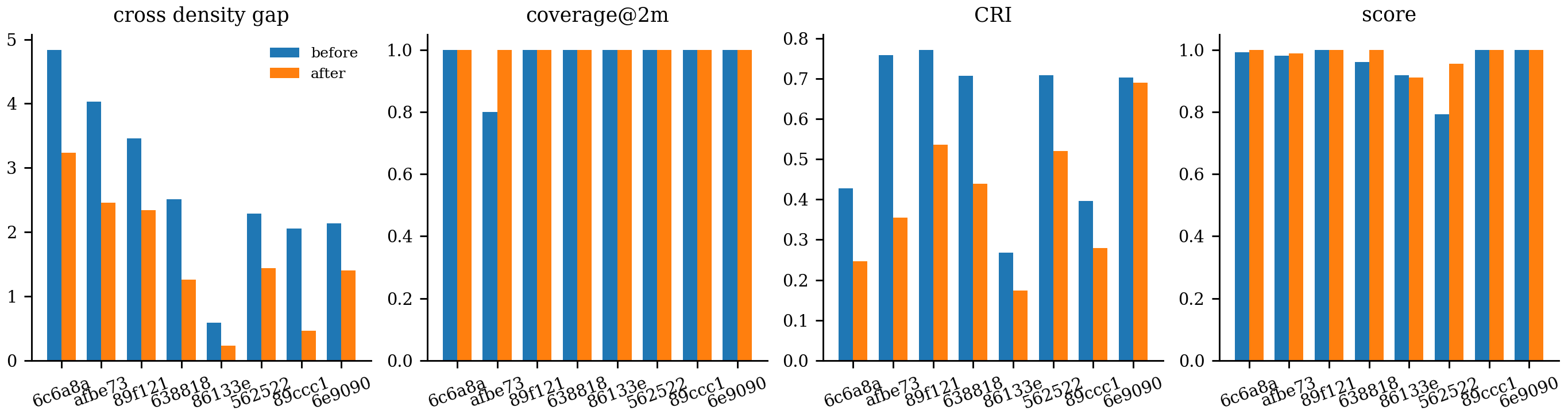}
  \caption{\textbf{A22.6 mechanism-pilot trace.} The auxiliary objective and
  cross-density gap decrease from $1.0571$ to $0.4701$ and from $2.7393$ to
  $1.6047$, respectively, while teacher coverage at $2$ m increases from
  $0.975$ to $1.000$. AC-JEPA remains frozen throughout.}
  \label{fig:a22_6_mechanism_pilot}
\end{figure}

The direction is consistent across all eight selected scenes: the auxiliary
loss and cross-mode density mismatch improve, with no coverage drop and no
AC-JEPA parameter update. This compact result is the strongest positive
mechanism evidence currently available in the package; it does not establish
that the pilot transfers to the released NAVSIM endpoint.

\subsection{Source Tables and Evidence Boundary}
\label{app:experimental_provenance}

The numerical tables are transcribed from
\texttt{NAVSIM\_tables.md} and
\texttt{hugsim\_table\_with\_baseline.md}. Empty numerical cells are not
imputed. Best and second-best marks in NAVSIM are recomputed within each
supplied setting/backbone group when the source emphasis marks disagree with
the numbers; the reported numerical values and their precision are unchanged.
HUGSIM row-wise marks identify the largest supplied value and do not imply a
matched evaluation protocol.

In the main NAVSIM v1, NAVSIM v2, and HUGSIM comparison tables, external
baseline results are source-reported values from the corresponding cited
papers. Our results are obtained using our implementation under the evaluation
protocols described above. The ablation and paired cross-generator tables have
separate provenance notes and are not classified as source-reported baseline
comparisons here.

The benchmark tables provide the quantitative comparisons. The qualitative
figures in \cref{app:qualitative} illustrate candidate selection, planner
comparisons, scene context, and closed-loop plan renderings; they are not used
as substitutes for matched quantitative evaluation.

\subsection{AC-JEPA Relation Audit: Geometry, Latent, and Combined Readouts}
\label{app:acjepa_relation_audit}
This subsection evaluates whether frozen AC-JEPA features contain relation
information beyond trajectory geometry. These geometry-only and combined
readouts are diagnostic controls only; they are not the AMC affinity used in
Sec.~3.2, which is computed solely from the frozen predictive tensors after
the action mapping $A(\tau)$.
This subsection isolates the evidence relevant to the claim that a predictive
AC-JEPA relation contains information not available from a purely geometric
relation. We distinguish three levels of evidence: (i) a matched feature
anatomy readout, (ii) a geometry- and raw-action-controlled conditional readout,
and (iii) a full-support held-out relation audit that reports both relation
discrimination and redundancy retention. All three analyses use frozen
representations, fixed pair labels, and scene/log-held-out evaluation; they do
not retrain the planner or AC-JEPA.
\begin{figure}[!htbp]
  \centering
  \includegraphics[width=0.88\linewidth]{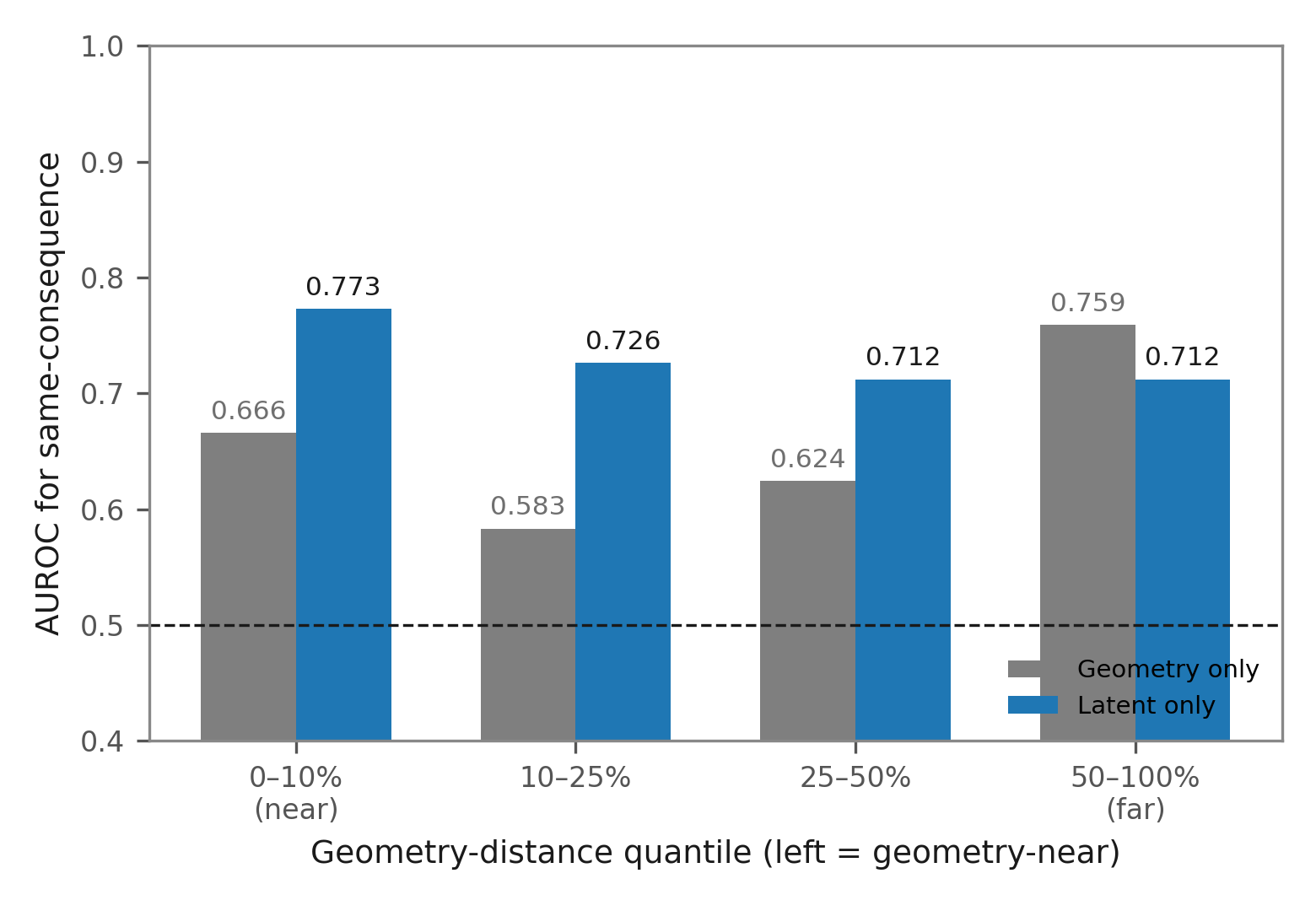}
  \caption{\textbf{Relation discrimination across geometry difficulty.} The
  held-out diagnostic shows that predictive and geometric readouts behave
  differently across geometry bins: the predictive readout is stronger in
  near and intermediate bins but is not globally dominant. The matched
  combined-readout gains are reported in Table~\ref{tab:acjepa_relation_anatomy};
  this plot is not an end-to-end planner ablation.}
  \label{fig:a2_auc_by_geometry}
\end{figure}

\paragraph{Matched feature anatomy.}
The first audit compares geometry-only ($G$), canonical latent-only ($Z$), and
their concatenation ($G+Z$); the horizon-latent variant ($G+H$) is included as
the multi-horizon predictive representation. The readout, pair labels, and
held-out split are held fixed across rows. The result is not a replacement of
geometry by latent space: latent-only is weaker globally, whereas adding the
latent features to geometry improves the held-out relation task.

\begin{table}[t]
  \centering
  \caption{
  Matched held-out relation readout anatomy (A2.1). AUROC and AP are
  computed on the same relation labels and held-out pairs for every feature
  group. The table supports an incremental contribution of predictive features,
  not global superiority of latent-only distance.}
  \label{tab:acjepa_relation_anatomy}
  \small
  \renewcommand{\arraystretch}{1.05}
  \begin{tabular}{lcc}
    \toprule
    Feature group & AUROC & AP \\
    \midrule
    Geometry only ($G$) & 0.891936 & 0.660612 \\
    Canonical latent only ($Z$) & 0.824435 & 0.626469 \\
    Geometry + canonical latent ($G+Z$) & 0.900970 & 0.691880 \\
    Geometry + horizon latent ($G+H$) & 0.901024 & 0.693624 \\
    \bottomrule
  \end{tabular}
\end{table}

At the very low false-merge-rate operating points used by the original A2.1
diagnostic, the redundancy-recall gain was zero for the combined readouts.
Accordingly, A2.1 alone is not sufficient to claim a redundancy improvement;
its defensible conclusion is that the combined predictive representation
improves global relation discrimination while latent-only does not dominate
geometry-only.

\paragraph{Conditional latent increment beyond geometry and raw action.}
The stronger test controls for geometric distance, raw action difference, raw
trajectory difference, endpoint distance, and maximum point distance in the
$G+A$ baseline. The same-capacity logistic or small-MLP readout is then
augmented with frozen AC-JEPA features ($G+A+Z$ or $G+A+H$). The primary
geometry-degenerate region is $\mathrm{ADE}_{\mathrm{geometry}}\leq0.3$ m;
bootstrap units are scenes or logs rather than individual correlated pairs.
Here $\Delta$log-loss is defined as (augmented $-$ baseline), so a negative
interval indicates an improvement.

\begin{table}[t]
  \centering
  \caption{
  Conditional latent increment in the geometry-degenerate held-out
  region (A2.4). All intervals exclude zero at both scene and log levels.}
  \label{tab:acjepa_conditional_increment}
  \small
  \resizebox{\linewidth}{!}{%
  \begin{tabular}{lccc}
    \toprule
    Comparison & Readout & Scene $\Delta$log-loss (95\% CI) & Log $\Delta$log-loss (95\% CI) \\
    \midrule
    $G+A+Z$ vs. $G+A$ & Logistic & $[-0.0481,-0.0236]$ & $[-0.0604,-0.0107]$ \\
    $G+A+H$ vs. $G+A$ & Logistic & $[-0.0642,-0.0378]$ & $[-0.0791,-0.0206]$ \\
    $G+A+Z$ vs. $G+A$ & Small MLP & $[-0.0650,-0.0371]$ & $[-0.0792,-0.0238]$ \\
    $G+A+H$ vs. $G+A$ & Small MLP & $[-0.1290,-0.0931]$ & $[-0.1402,-0.0799]$ \\
    \bottomrule
  \end{tabular}}
\end{table}

The corresponding scene-level information gains (negative $\Delta$log-loss)
are $+0.0357$ for logistic $G+A+Z$ and $+0.1121$ for small-MLP $G+A+H$.
These are conditional-information proxies, not estimates of mutual
information and not end-to-end planner scores.

\paragraph{Full-support relation task and redundancy retention.}
The A2.7 audit rebuilds the three readouts on the same frozen bank and evaluates
all 496 unordered candidate pairs per scene under a fixed scene/log split.
This provides the closest available matched evidence for the two requested
metric families: relation-task performance (log-loss, AUROC, AP, and
unsafe-veto recall) and redundancy retention at the same 95\% pair-retention
threshold.

\begin{table}[t]
  \centering
  \caption{
  Full-support held-out relation audit (A2.7). Rows use the same pair
  labels, split, and retention threshold; $Z$ and $H$ are frozen AC-JEPA latent
  features.}
  \label{tab:acjepa_full_support}
  \small
  \resizebox{\linewidth}{!}{%
  \begin{tabular}{lccccc}
    \toprule
    Feature group & log-loss & AUROC & AP & Unsafe-veto recall & Redundancy retention \\
    \midrule
    $G+A$ & 0.3918 & 0.9073 & 0.9771 & 63.38\% & 94.18\% \\
    $G+A+Z$ & 0.3634 & 0.9191 & 0.9798 & 64.97\% & 94.92\% \\
    $G+A+H$ & 0.3626 & 0.9193 & 0.9798 & 64.90\% & 94.98\% \\
    \midrule
    \multicolumn{5}{l}{\textit{$>1.0$ m geometry bin:} $G+A$: 66.81\% / 88.44\%;\quad $G+A+Z$: 68.46\% / 89.92\%;\quad $G+A+H$: 68.38\% / 90.04\%.}\\
    \bottomrule
  \end{tabular}}
\end{table}

The paired scene/log bootstrap remains positive for both latent-augmented
readouts in every reported geometry bin. In the widest-support bin, where the
geometry baseline misses the most oracle-safe edges, adding latent features
raises unsafe-veto recall by 1.57--1.65 percentage points and redundancy
retention by 1.48--1.60 percentage points. Thus the strongest current claim is
that predictive latent features provide conditional relation information beyond
geometry and raw action, with a small but measurable redundancy benefit in the
held-out relation audit.

\paragraph{Evidence boundary for the strict ablation criterion.}
The evidence above does not establish all three of the following stronger
claims: (1) latent-only globally outperforms geometry-only, (2) a matched
end-to-end planner trained with Geometry-only, Latent-only, and
Geometry+Latent objectives improves the planner task endpoint, and (3) the same
planner runs simultaneously improve task performance and redundancy metrics.
The current package contains no such equal-budget planner ablation. We
therefore use ``relation-task performance'' for the held-out readout metrics in
this appendix and do not relabel them as NAVSIM PDMS or end-to-end planning
performance. This boundary preserves the valid positive result---incremental
predictive information and relation-level redundancy retention---without
claiming a matched task-endpoint proof that has not been run.

\subsection{Action-Conditioned Predictive Relations: Prior Basis and Evidence Boundary}
\label{app:acjepa_action_sensitivity}

This subsection records the prior basis and the implementation-specific evidence
for using a frozen action-conditioned predictive representation as a candidate
relation signal. We keep three evidence levels separate: (i) the JEPA literature
provides the predictive-representation and action-conditioned modeling precedent,
(ii) Delta-JEPA provides an independent latent-transition argument and empirical
diagnostics for action-sensitive dynamics, and (iii) the experiments below test
the evaluated AC-JEPA checkpoint and the AMC relation readout itself. Results from
the prior papers are not treated as direct proofs about AMC-Drive.

\paragraph{Relation to the JEPA prior.}
V-JEPA~2 establishes the predictive latent representation setting used by the
V-JEPA family~\citep{assran2025vjepa2}. The V-JEPA~2-AC branch in the public
implementation provides the direct architectural precedent for injecting an
action sequence into the latent predictor and forecasting an action-conditioned
future representation. V-JEPA~2.1 is used here as a dense predictive-feature
reference; its representation role is distinct from the action-conditioned
precedent supplied by V-JEPA~2-AC. Delta-JEPA~\citep{zhang2026deltajepa}
provides a complementary transition-level argument. With
\begin{equation}
  \Delta z_t=z_{t+1}-z_t,\qquad g(\Delta z_t)\approx a_t,
  \label{eq:deltajepa_prior}
\end{equation}
an action-insensitive transition would make the executed action difficult to
recover from the latent displacement. Their ablations and action-sensitivity
analyses therefore motivate testing whether counterfactual action interventions
change future-latent prediction in a direction that is consistent with the
observed action. This is a prior mechanistic rationale, rather than a theorem
about the present driving checkpoint.

\paragraph{The weaker assumption used by AMC.}
For a fixed scene context $s$, AMC forms candidate-specific predictive
representations $z_i=F_{\mathrm{AC}}(s,a_i)$ and computes the relation distance
and affinity from the frozen representations, as specified in
Eq.~\ref{eq:amc_pilot_distance} and Eq.~\ref{eq:affinity}. AMC uses this affinity
only to characterize and regularize the organization of a fixed candidate set.
It does not require latent Euclidean distance to be a calibrated physical
outcome, a safety score, or a one-to-one action label. The relevant assumption
is weaker: the predictive representation should contain informative relational
variation across candidate actions. The A2.1, A2.4, and A2.7 audits above test
this relation-level assumption beyond trajectory geometry and raw action; they
are not direct action-identification benchmarks.

\paragraph{Paired action intervention (E0).}
We report the existing inference-only diagnostic on the fixed
action-stride-1 AC-JEPA checkpoint. It uses 256 validation samples (1,024 paired
rows), identical visual context, proprioception, and target future latent, and
changes only the input action sequence of shape $[8,3]$. The four rollout
horizons are 0.5, 1, 2, and 4 seconds (steps 1, 2, 4, and 8). The wrong-action
conditions are zero action, a fixed-seed derangement (Shuffle), and a temporal
shift. For each pair we define
\begin{equation}
  \mathrm{CAG}_h
  =\mathrm{MSE}_{h}(\text{wrong action})
   -\mathrm{MSE}_{h}(\text{ground-truth action}).
  \label{eq:acjepa_cag}
\end{equation}
A positive CAG means that the ground-truth action gives the lower
future-latent prediction error. Confidence intervals are percentile intervals
from 5,000 paired bootstrap resamples.

\begin{table}[t]
  \centering
  \caption{\textbf{Paired action intervention.} Shuffle CAG is the increase in
  future-latent MSE after replacing the ground-truth action by a fixed-seed
  derangement. All intervals are positive.}
  \label{tab:acjepa_action_perturbation}
  \small
  \setlength{\tabcolsep}{4pt}
  \begin{tabular}{lccc}
    \toprule
    Horizon & Shuffle CAG & 95\% bootstrap CI & Positive rate \\
    \midrule
    0.5 s & 0.136983 & [0.116567, 0.159578] & 94.14\% \\
    1 s   & 0.282042 & [0.241997, 0.325663] & 96.09\% \\
    2 s   & 0.601921 & [0.524839, 0.681142] & 95.31\% \\
    4 s   & 1.178901 & [1.054256, 1.304259] & 96.09\% \\
    \bottomrule
  \end{tabular}
\end{table}

The effect is positive at every evaluated horizon and grows with the rollout
horizon. At 2 seconds, it remains positive in all behavior buckets: LEFT
($n=59$, CAG $=0.696542$, 95\% CI $[0.538416,0.876427]$), RIGHT
($n=22$, $0.895216$, $[0.591664,1.227518]$), BRAKE ($n=110$, $0.476396$,
$[0.379782,0.579924]$), and ACCELERATE ($n=57$, $0.694297$,
$[0.529049,0.870787]$). The zero-action control gives 2-second CAG
$0.398265$, whereas Shuffle gives $0.601921$; the temporal-shift control is
smaller ($0.015042$), which is consistent with sensitivity to action identity
and weaker sensitivity to a small neighboring-step shift.

\begin{figure}[!htbp]
  \centering
  \includegraphics[width=0.96\linewidth]{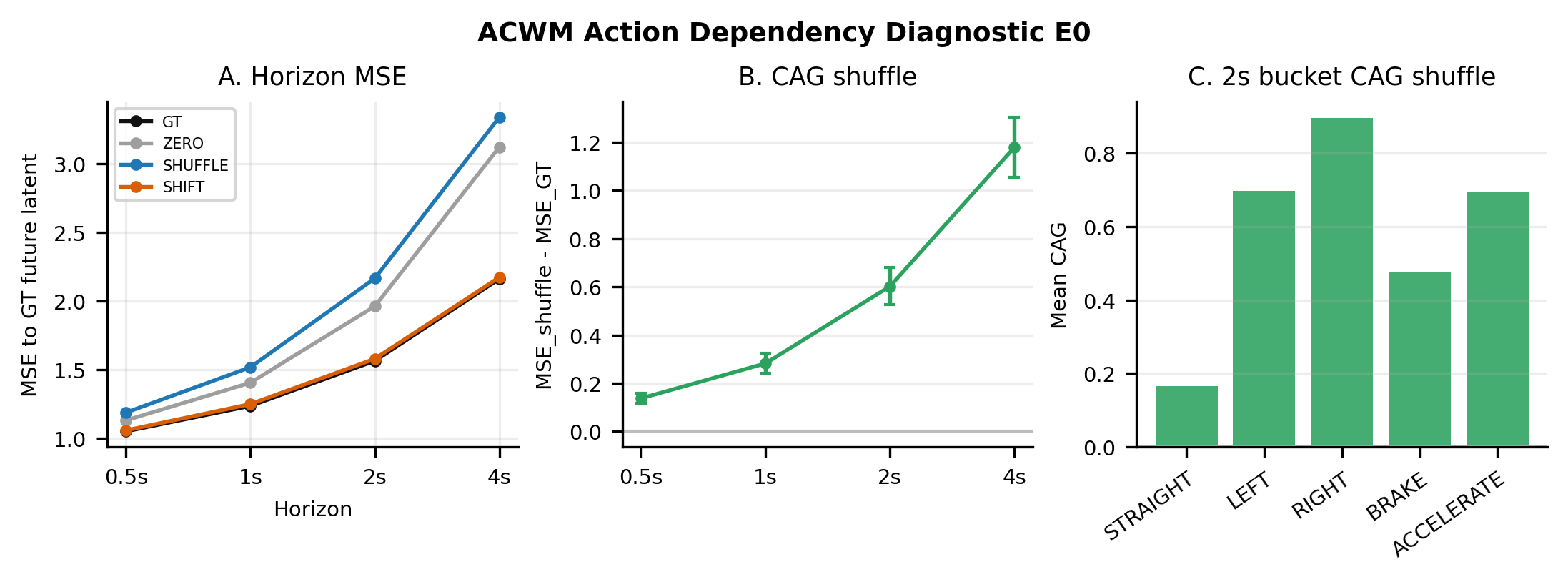}
  \caption{\textbf{Action perturbation diagnostic.} Fixed-context intervention
  results for ground-truth, zero, shuffled, and temporally shifted actions. The
  positive CAG intervals show that the evaluated checkpoint uses action
  information in future-latent prediction at all four horizons.}
  \label{fig:acjepa_action_dependency}
\end{figure}

\begin{figure}[!htbp]
  \centering
  \includegraphics[width=0.96\linewidth]{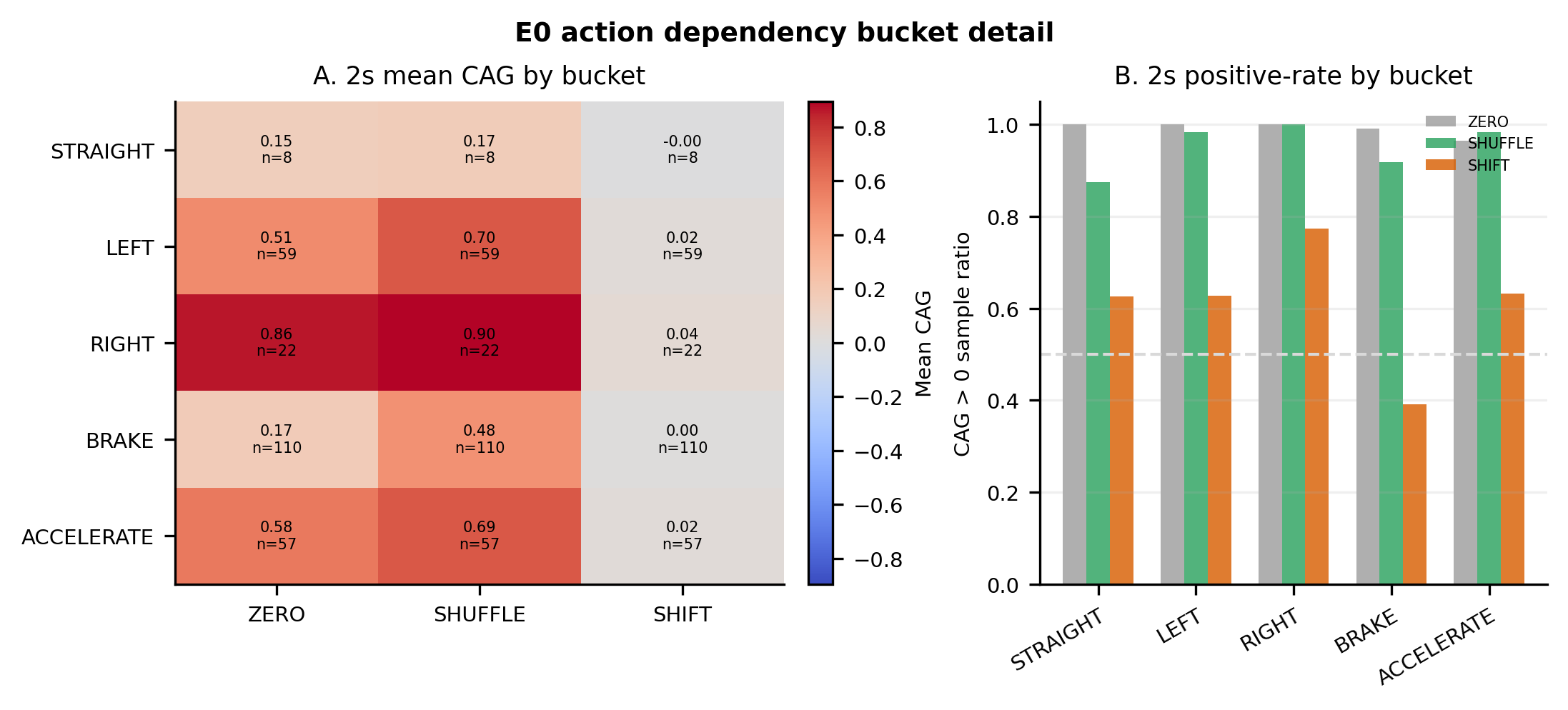}
  \caption{\textbf{Behavior-bucket action sensitivity.} At the 2-second
  horizon, the Shuffle intervention remains positive for LEFT, RIGHT, BRAKE,
  and ACCELERATE buckets. The plot is a checkpoint-level diagnostic and does
  not claim complete action identifiability.}
  \label{fig:acjepa_action_buckets}
\end{figure}

\paragraph{Temporal counterfactuals and frozen-path check.}
The structured temporal counterfactual diagnostic gives CAG values of
$0.015042$ (Shift$+1$, 95\% CI $[0.007628,0.023101]$), $0.063146$
(Shift$+2$, $[0.044433,0.084934]$), and $0.190634$ (Shift$+4$,
$[0.149107,0.234387]$); reversing the action sequence gives $0.195616$
($[0.152204,0.241112]$), and Shuffle gives $0.070473$
($[0.049983,0.094712]$). In the separate frozen-path smoke test, the mean
absolute gradient with respect to the action input is $0.0273527$ and the
maximum absolute gradient is $0.972636$, while the number of AC-JEPA parameter
gradient tensors is zero and no optimizer step is taken. This verifies that the
action input has a live differentiable path through the frozen predictor; it is
not a training-gain or planner-performance claim.

The extended diagnostic reports overall ground-truth top-1 action ranking of
$0.468750$ and top-3 ranking of $0.718750$, compared with a random top-1
baseline of $0.062500$. The corresponding top-1 rates are $0.661017$ for LEFT,
$0.727273$ for RIGHT, $0.345455$ for BRAKE, and $0.456140$ for ACCELERATE.
These values support checkpoint-level action sensitivity, while the broader
validation record remains scoped by long-horizon instability and proprioceptive
redundancy; the paired AP/NP control is not included in that diagnostic.

\paragraph{Evidence boundary and public resources.}
The combined evidence supports the following bounded statement: replacing the
ground-truth action changes future-latent prediction in a reproducible direction,
and frozen AC-JEPA features add conditional relation information beyond geometry
and raw action in the held-out relation audits. It does not establish that every
action is cleanly identifiable, that latent distance is a calibrated outcome
metric, or that AMC-Drive inherits a theorem or an end-to-end planning gain from
Delta-JEPA or V-JEPA. The external papers provide the mathematical and empirical
precedent; the intervention and relation audits above provide the
AMC-Drive-specific evidence.

For reproducibility, the public resources used as methodological references are:
\begin{table}[!htbp]
  \centering
  \caption{\textbf{Public methodological resources.} V-JEPA 2, V-JEPA 2-AC,
  and the V-JEPA 2.1 implementation are available in the official Meta
  repository. The Delta-JEPA paper and supplementary source are linked below;
  no author-maintained Delta-JEPA code repository is identified in the cited
  arXiv record, so no unverified mirror is labelled as official.}
  \label{tab:acjepa_public_resources}
  \small
  \setlength{\tabcolsep}{3pt}
  \begin{tabular}{p{0.25\linewidth}p{0.26\linewidth}p{0.39\linewidth}}
    \toprule
    Reference & Public resource & Role in this appendix \\
    \midrule
    V-JEPA 2 / 2-AC / 2.1
      & \url{https://github.com/facebookresearch/vjepa2}
      & Predictive-representation baseline, action-conditioned predictor precedent, and released configurations. \\
    Delta-JEPA
      & \url{https://arxiv.org/abs/2606.31232}
      & Latent-difference action-decoding rationale, action-sensitivity analyses, and ablation structure. \\
    AMC-Drive
      & \url{https://github.com/lumen2023/AMC-Drive}
      & Public NAVSIM v1 implementation, checkpoint, submission script, and supporting evidence. \\
    \bottomrule
  \end{tabular}
\end{table}

\FloatBarrier
\section{Cross-Generator and Cross-Domain Transfer}
\label{app:transfer}

We use the same candidate-level calibration interface with several driving
generator families and with a robotic manipulation policy. The table and
figures in this section record the corresponding settings and observations.
They are separate from the matched NAVSIM and HUGSIM evaluations reported in
the main text.

\subsection{Cross-Generator Implementations}
\label{app:cross_generator}

The interface receives a generated candidate set and does not depend on
whether the set comes from diffusion, flow matching, or proposal-based
generation. For driving, the compared generator families include DiffusionDrive,
GoalFlow, ReCogDrive, and Drive-JEPA~\citep{liao2025diffusiondrive,
xing2025goalflow,li2025recogdrive,wang2026drivejepa}. Their NAVSIM v1
comparison is listed in Table~\ref{tab:amc_transfer}.

The provenance audit for the cross-generator comparison is given below.  Each
driving row was produced by deploying the cited official implementation,
running its Base configuration, adding the AMC training branch, and then
training/evaluating the +AMC counterpart under the same NAVSIM v1 split and
evaluator.  Consequently, the four rows are paired local reproductions rather
than source-reported baselines combined with an AMC-only run.  The current
paper repository does not include the raw per-row checkpoint/log bundles;
the table therefore records author-confirmed provenance and keeps the
artifact-archival limitation explicit.

The shared interface should not be conflated with shared hyperparameters.
For these four driving rows, the paper repository does not expose verified
values for $\lambda_{\mathrm{AMC}}$, $\lambda_{\mathrm{guard}}$, or $\tau_s$,
nor a per-generator tuning manifest (search space, validation split, budget,
or selection rule).  The paired-run provenance above therefore does not by
itself establish hyperparameter invariance; that separate protocol question is
tracked in R21.

\subsection{Cross-Generator Calibration Principle of AMC}
\label{app:cross_generator_calibration}

AMC is an auxiliary structural regularizer.  It does not replace the original
planning objective.  For every candidate generator, the training objective is
the same form as in the main text,
\begin{equation}
  \mathcal{L}_{\mathrm{total}}
  =
  \mathcal{L}_{\mathrm{plan}}
  +
  \lambda_{\mathrm{AMC}}\mathcal{L}_{\mathrm{AMC}}
  +
  \lambda_{\mathrm{guard}}\mathcal{L}_{\mathrm{guard}} .
  \label{eq:cross_generator_calibration_objective}
\end{equation}
Here, $\mathcal{L}_{\mathrm{plan}}$ remains the primary supervision, while
the AMC terms regularize only the organization of the fixed-size candidate
set.

The generator-independent part of AMC is the candidate-level construction
itself.  Given candidates $\{\tau_i\}_{i=1}^{K}$, AMC uses the same
action-conditioned relation construction, affinity definition, effective
multimodality metric, topology formulation, and cross-mode guard for
proposal-based, diffusion-based, and flow-based planners.  These operations
consume the generated candidate set and its action-conditioned predictive
relations; they do not depend on the internal parameterization used to produce
the candidates.  The planner therefore retains its original output interface
and planning loss across generator families.

The calibration strength is interpreted through candidate-set diagnostics.
Different generators can organize candidates differently even when their
nominal budget $K$ is identical.  The role of the auxiliary weights is thus to
control the strength of the corresponding structural correction, based on the
observed mismatch between the planner affinity $S^{P}$ and the reference
affinity $S^{R}$.  This diagnosis is structural and is not defined by
optimizing a planning metric.

\begin{table}[t]
  \centering
  \caption{\textbf{Diagnostic-driven calibration principle.} The diagnostics
  determine which structural component supplies the corrective signal; the
  planning objective remains the primary supervision.}
  \label{tab:cross_generator_calibration}
  \small
  \setlength{\tabcolsep}{4pt}
  \renewcommand{\arraystretch}{1.12}
  \begin{tabular}{@{}p{0.27\linewidth}p{0.30\linewidth}p{0.34\linewidth}@{}}
    \toprule
    Candidate-set issue & Diagnostic & Corresponding AMC component \\
    \midrule
    Excessive candidate concentration
      & $K_{\mathrm{eff}}^{P}<K_{\mathrm{eff}}^{R}$
      & Effective-capacity calibration in $\mathcal{L}_{\mathrm{AMC}}$ \\
    Topology inconsistency
      & $\rho^{P}\neq\rho^{R}$ after capacity alignment
      & Topology-aware term within $\mathcal{L}_{\mathrm{AMC}}$ \\
    Cross-mode contraction
      & $B_{\times}^{P}>B_{\times}^{R}$
      & One-sided cross-mode guard $\mathcal{L}_{\mathrm{guard}}$ \\
    \bottomrule
  \end{tabular}
\end{table}

The table specifies the calibration logic rather than a new loss or a new
experiment: capacity mismatch activates the effective-capacity correction,
topology mismatch directs the topology-aware correction, and excessive
cross-mode burden activates the guard.  The numerical hyperparameters,
training commands, and implementation configurations for the released NAVSIM
v1 run are provided in the public code repository. The cross-generator run
bundles are not part of that release. This appendix records the shared
structural principle and the diagnostic rule used to interpret the auxiliary
terms.

\begin{table}[t]
  \centering
  \caption{\textbf{Cross-generator provenance audit.} ``Official + local AMC''
  means that the cited official implementation was deployed locally, Base was
  run, and +AMC was trained/evaluated with the AMC branch under the same
  NAVSIM v1 protocol.  The raw per-row bundles are not included in this paper
  repository.}
  \label{tab:cross_generator_provenance}
  \small
  \setlength{\tabcolsep}{3pt}
  \begin{tabular}{@{}lccccc@{}}
    \toprule
    Generator & Base value & +AMC value & Implementation & Pairing & Protocol \\
    \midrule
    DiffusionDrive & local & local & official + local AMC & paired & same NAVSIM v1 \\
    ReCogDrive     & local & local & official + local AMC & paired & same NAVSIM v1 \\
    GoalFlow       & local & local & official + local AMC & paired & same NAVSIM v1 \\
    Drive-JEPA     & local & local & official + local AMC & paired & same NAVSIM v1 \\
    \bottomrule
  \end{tabular}
\end{table}

\begin{table}[t]
\centering
\caption{
Comparison with representative recent multimodal planning methods on
NAVSIM v1. We report the overall PDMS for a concise comparison.
}
\label{tab:amc_transfer}
\small
\setlength{\tabcolsep}{6pt}

\begin{tabular}{llN}
\toprule
Method & Planning Paradigm & PDMS $\uparrow$ \\
\midrule

DiffusionDrive
& Diffusion
& 88.1 \\

GoalFlow
& Flow-based
& 90.3 \\

ReCogDrive
& VLM + Diffusion + RL
& 90.8 \\

DriveSuprim
& Proposal / Selection
& 93.5 \\

Drive-JEPA
& V-JEPA + Proposal
& 93.3 \\

CLEAR
& JEPA-backed Generative
& 93.7 \\

Drive-HWM
& V-JEPA-backed World Model
& 93.8 \\

\midrule

\rowcolor{methodrow}
\textbf{\method{}}
& \textbf{AC-JEPA + AMC}
& \textbf{93.9} \\

\bottomrule
\end{tabular}
\end{table}

The scope of the candidate-level interface is summarized in
Figure~\ref{fig:applications} in the main text. The comparisons below describe
the interface and the supplied values.  The four driving rows are labelled as
matched paired reproductions on the basis of the author-run protocol above;
the remaining limitation is artifact-level independent reconstruction from
the current archive.

\subsection{Robotic Manipulation on Push-T}
\label{app:pusht}

We also instantiate a Push-T-adapted variant of AMC in a diffusion policy on
the manipulation task. The supplied visualization labels this variant
``AMC-RTI Dedup''; it is an application-specific realization of the same
candidate-level calibration interface, not a separate AMC method. The paired
scores are reported in Table~\ref{tab:amc_generalization}: the reproduced
Diffusion Policy baseline obtains $76.0$, and the corresponding configuration
with the adapted AMC variant obtains $77.8$.

\begin{table}[t]
    \centering
    \caption{
    \textbf{Generalization across candidate generators and tasks.}
    $\Delta$ reports the numerical difference between the two values.
    For the four driving rows, Base and +AMC are paired local reproductions:
    we deployed the corresponding official implementation, ran its baseline,
    then added the AMC training branch and re-trained/evaluated under the same
    NAVSIM v1 split and evaluator.  Thus no source-reported baseline is paired
    with a local AMC run.  ``R'' denotes this author-run reproduced protocol.
    The raw per-row checkpoint/log bundles are not included in the present
    paper repository, so ``R'' records provenance rather than an independently
    re-executed audit.  For Push-T, ``U'' retains its previous meaning of
    unverified paired provenance.
    For Push-T, \texttt{+AMC} denotes the Push-T-adapted variant described in
    Appendix~\ref{app:pusht}.
    }
    \label{tab:amc_generalization}
    \small
    \setlength{\tabcolsep}{3.2pt}
    \renewcommand{\arraystretch}{1.03}
    \begin{tabular}{@{}llcrMrr@{}}
        \toprule
        Planner
        & Generator
        & Metric
        & Base
        & \cellcolor{methodrow}+AMC
        & $\Delta$
        & Prov.
        \\
        \midrule

        \multicolumn{7}{@{}l}{\textit{Autonomous driving}} \\
        DiffusionDrive
        & Diffusion
        & PDMS
        & 88.1
        & \cellcolor{methodrow}90.2
        & 2.1
        & R
        \\
        ReCogDrive
        & Diffusion
        & PDMS
        & 90.8
        & \cellcolor{methodrow}\textbf{91.1}
        & 0.3
        & R
        \\
        GoalFlow
        & Flow
        & PDMS
        & 90.3
        & \cellcolor{methodrow}90.4
        & 0.1
        & R
        \\
        Drive-JEPA
        & Proposal
        & PDMS
        & 93.3
        & \cellcolor{methodrow}93.9
        & 0.6
        & R
        \\

        \midrule
        \multicolumn{7}{@{}l}{\textit{Robotic manipulation}} \\

        %
        %
        %

        Diffusion Policy
        & Diffusion
        & Push-T Score
        & 76.0
        & \cellcolor{methodrow}\textbf{77.8}
        & 1.8
        & U
        \\

        \bottomrule
    \end{tabular}

    \vspace{4pt}
    \parbox{0.98\linewidth}{
    \footnotesize
    \textit{Push-T score.}
    The score measures the maximum normalized target coverage achieved
    by the T-shaped block during a rollout, averaged over test episodes.
    A coverage of at least $95\%$ corresponds to successful task completion.
    Higher is better.
    \textit{Provenance.} R = author-run reproduced paired protocol: the
    official generator implementation was deployed locally, Base was run,
    and +AMC was trained/evaluated after adding the AMC branch under the same
    NAVSIM v1 protocol.  U = paired provenance not independently verifiable
    from the current archive (used here only for Push-T).  The cross-generator
    R labels do not claim that the raw per-row checkpoint/log bundles are
    already included in this paper repository.
    }
\end{table}

\begin{figure}[!htbp]
  \centering
  \includegraphics[
    width=\linewidth,
    height=0.40\textheight,
    keepaspectratio
    ]{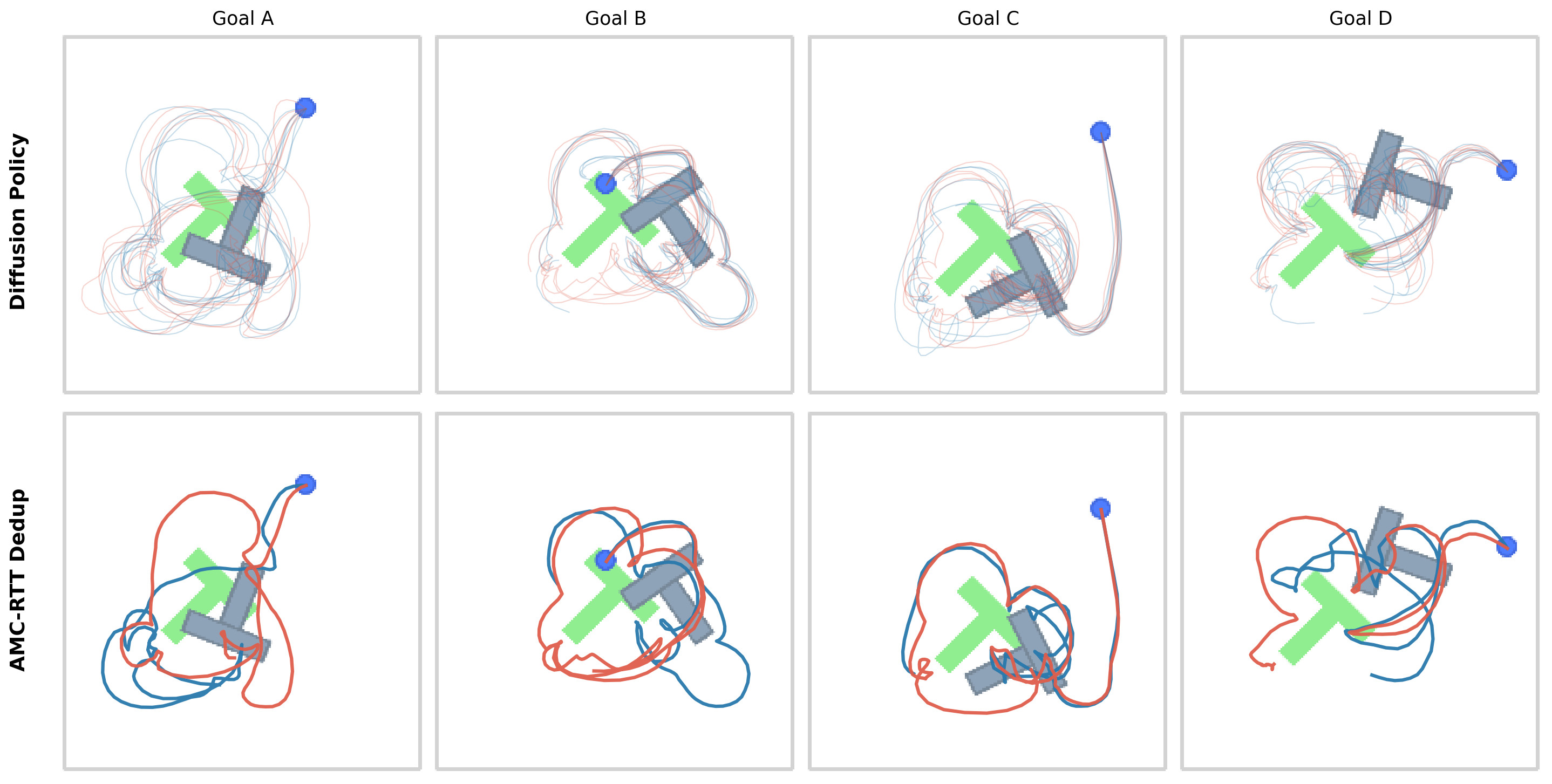}
  \caption{\textbf{Candidate trajectories on Push-T.}
  Each column corresponds to one target condition. The upper row shows
  Diffusion Policy candidates, and the lower row visualizes the
  Push-T-adapted AMC variant labelled ``AMC-RTI Dedup'' in the supplied
  figure. The apparent reduction in trajectories is a display-only redundancy
  filter; the policy retains its original output cardinality and inference
  interface. The panels provide a qualitative view of candidate organization;
  the matched Push-T scores are reported in Table~\ref{tab:amc_generalization}.}
  \label{fig:robo_pusht}
\end{figure}

Figure~\ref{fig:robo_pusht} shows four target conditions. In the upper row,
the displayed Diffusion Policy candidates form dense, overlapping trajectory
bundles around the object. The lower row visualizes the Push-T-adapted AMC
variant under the supplied ``AMC-RTI Dedup'' label. Its smaller-looking sets
are produced by a display-only redundancy filter for readability; no
candidate is removed from the policy output, and the output cardinality and
inference interface remain unchanged. The figure therefore illustrates
qualitative organization and efficient allocation within a fixed candidate
budget. The Push-T score is the quantitative comparison, and the image alone
does not establish coverage, task completion, or transfer to other robotic
tasks.

\FloatBarrier

\FloatBarrier
\section{Additional Qualitative Visualizations}
\label{app:qualitative}

This appendix provides visual diagnostics for the claims in Sec.~\ref{sec:experiments}.
We first inspect candidate relations and action-conditioned responses, then show
how candidate banks are filtered and compared, and finally collect representative
scene and closed-loop examples. The panels document observable geometry and
rendering behavior; numerical performance, redundancy reduction, and safety
claims are taken from the matched tables and experiments in the main text.

\subsection{Scope of the Title and the Effective-Multimodality Claim}
\label{app:effective_multimodality_scope}

The title refers to how much \emph{effective} multimodality is represented
within a fixed candidate budget, not to a procedure that searches for or changes
the nominal number of candidates. In the reported driving experiments, the
planner outputs $K=32$ trajectories and AMC-Drive leaves both this cardinality
and the inference interface unchanged. The quantity that varies is
$K_{\mathrm{eff}}$, computed from pairwise relations between candidate futures
in the frozen AC-JEPA predictive space. AC-JEPA therefore supplies a
scene-conditioned representation for deciding which candidates describe
distinct predictive futures and which are redundant; it does not generate
additional candidates or remove candidates at inference time.

This distinction also limits what can be inferred from the present appendix.
We do not report a nominal-budget sweep such as $K\in\{4,8,16,32,64\}$, so the
figures should not be read as identifying an optimal candidate count or as
showing a performance curve versus $K$. Likewise, the scene montages include
straight or junction scenes, merges, and dense interactions to expose variation
in candidate relations, but they are selected qualitative examples. They are
not a stratified estimate of $K_{\mathrm{eff}}$ by scene type and do not by
themselves establish a systematic monotonic relationship between scene
complexity and $K_{\mathrm{eff}}$. Any such claim would require matched
scene-level aggregation, including the nominal $K$, the distribution of
$K_{\mathrm{eff}}$, sample counts, and uncertainty for each scene category.

\subsection{Mechanism Diagnostics}
\label{app:mechanism_diagnostics}

\begin{figure}[!htbp]
  \centering
  \includegraphics[width=\linewidth,height=0.53\textheight,keepaspectratio]{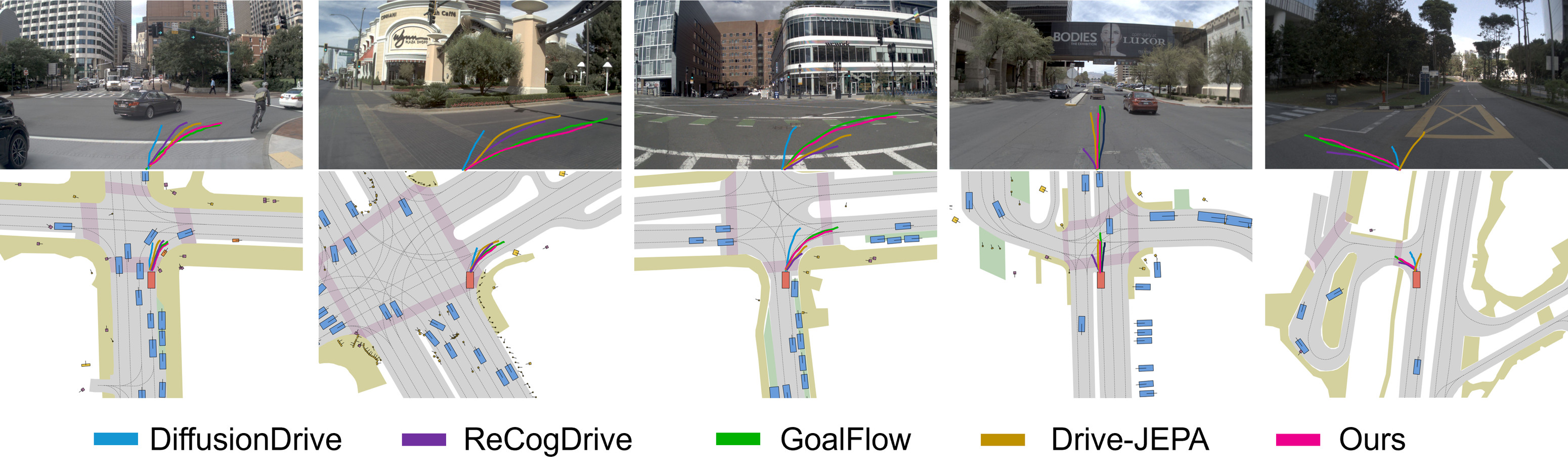}
  \caption{\textbf{NAVSIM trajectory context.} Each column pairs a camera view
  with a bird's-eye map and overlays the compared planners and reference path.
  The montage supplies scene context for the aggregate NAVSIM results; it is
  not a substitute for the quantitative table.}
  \label{fig:performance}
\end{figure}

\paragraph{NAVSIM trajectory context.}
The montage in \cref{fig:performance} gives a compact visual reference for the
trajectory comparisons used in the main text. Each column pairs a front-camera
view with its bird's-eye map, so the reader can inspect both road context and
trajectory geometry without treating a single scene as a ranking experiment.

\begin{figure}[!htbp]
  \centering
  \begin{minipage}[t]{0.49\linewidth}\centering
    \includegraphics[width=\linewidth,height=0.22\textheight,keepaspectratio]{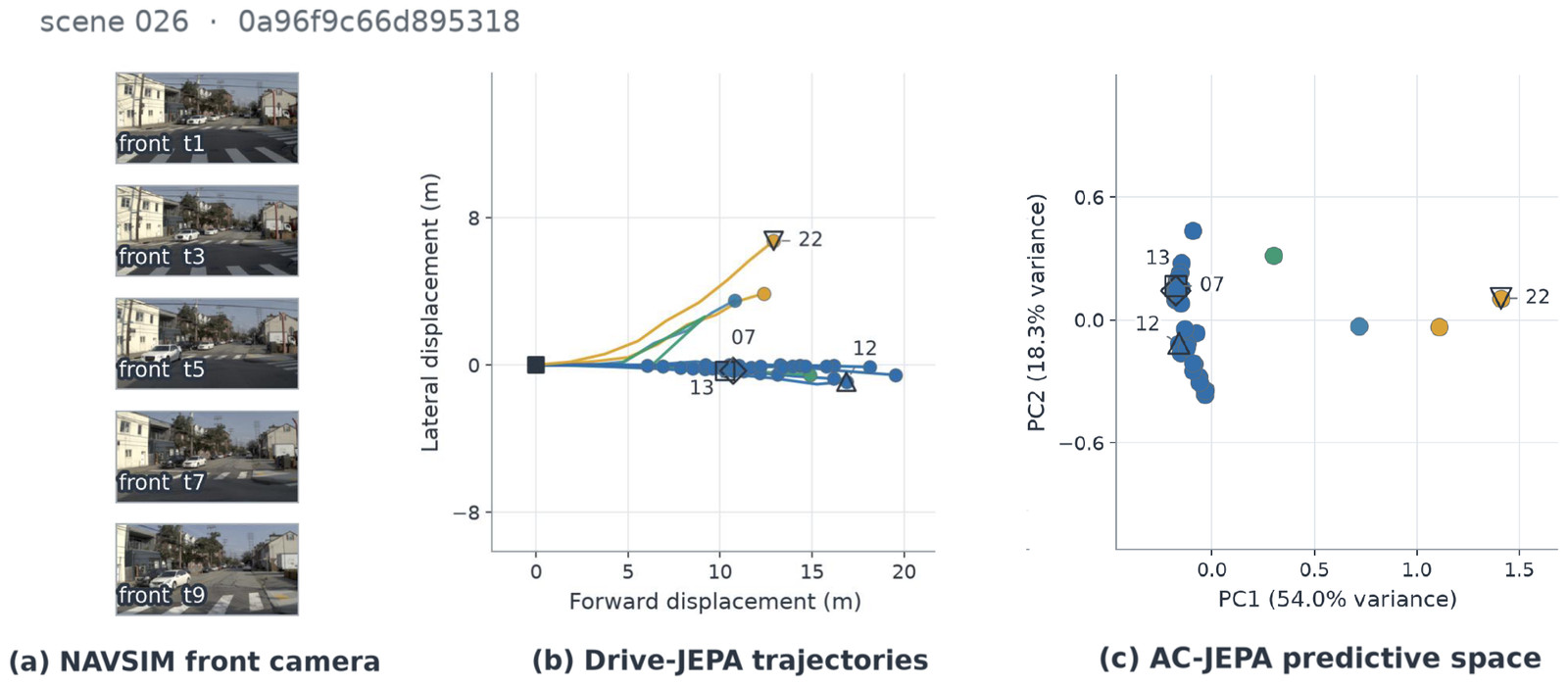}
  \end{minipage}\hfill
  \begin{minipage}[t]{0.49\linewidth}\centering
    \includegraphics[width=\linewidth,height=0.22\textheight,keepaspectratio]{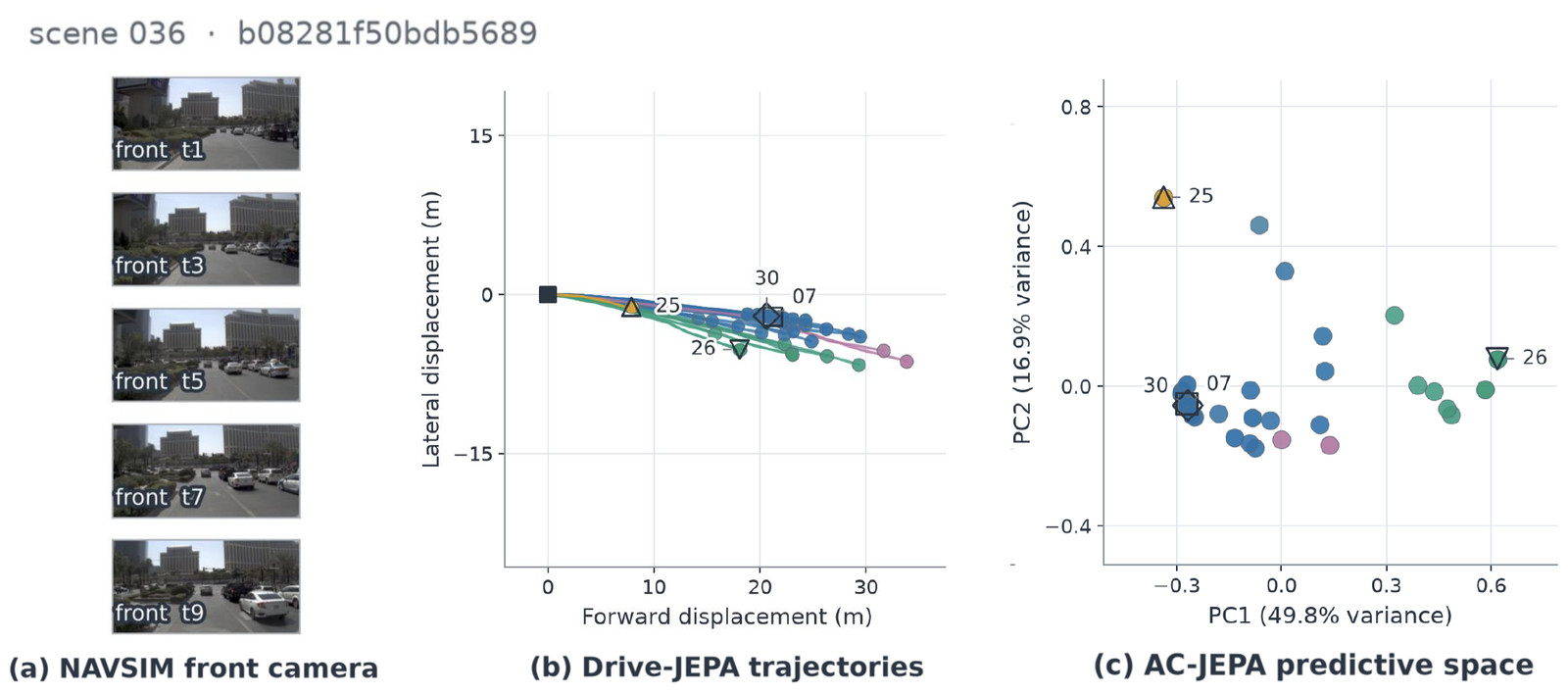}
  \end{minipage}

  \vspace{0.35em}

  \begin{minipage}[t]{0.49\linewidth}\centering
    \includegraphics[width=\linewidth,height=0.22\textheight,keepaspectratio]{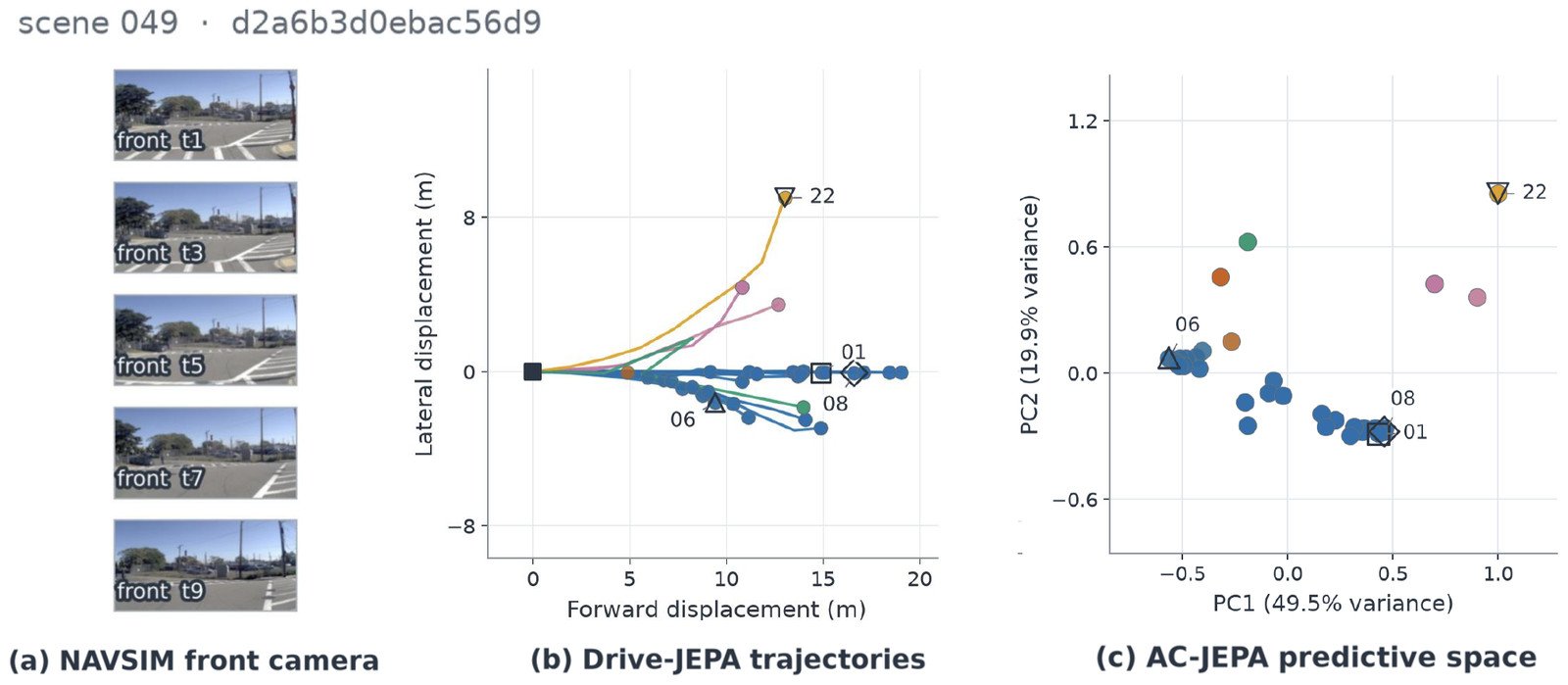}
  \end{minipage}\hfill
  \begin{minipage}[t]{0.49\linewidth}\centering
    \includegraphics[width=\linewidth,height=0.22\textheight,keepaspectratio]{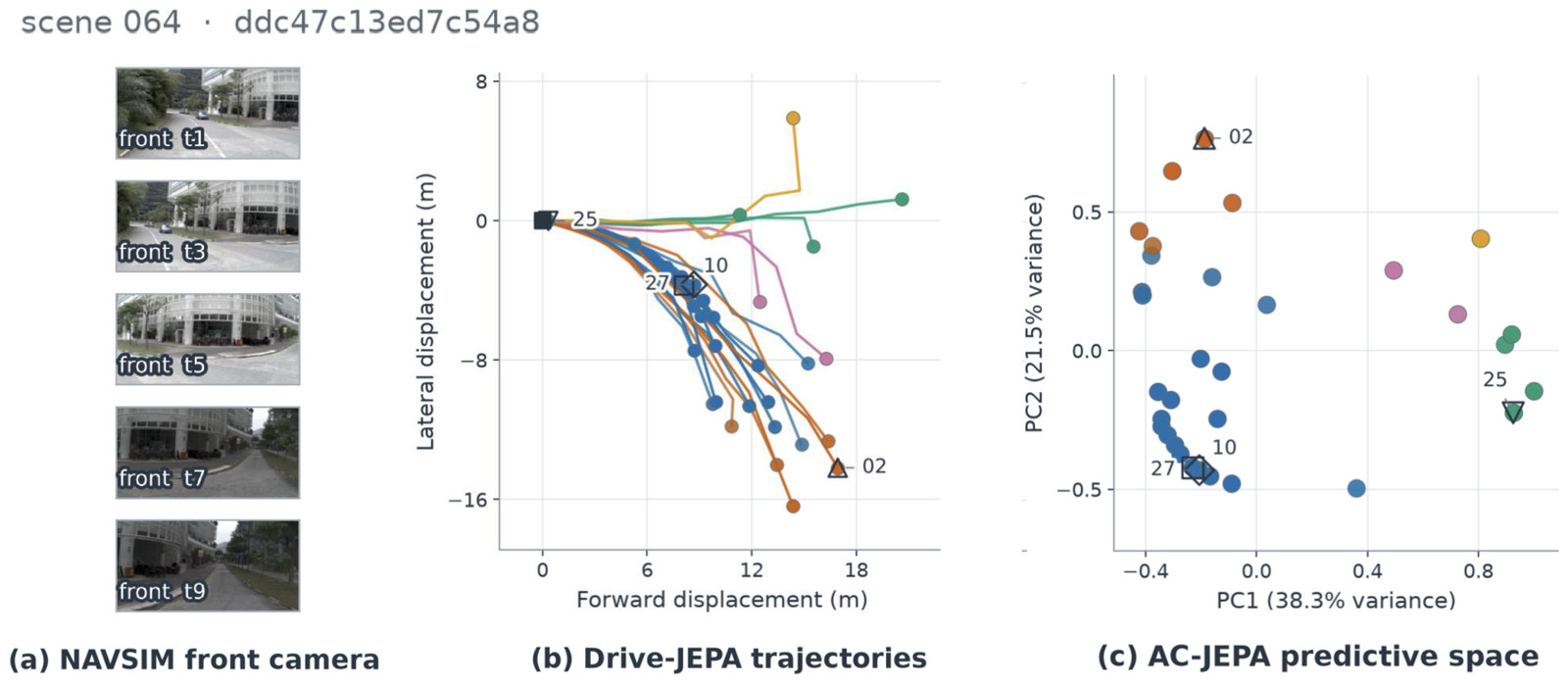}
  \end{minipage}
  \caption{\textbf{Representative latent-space views.} The four scenes show
  front-camera context, candidate trajectories, and AC-JEPA predictive-space
  coordinates. The montage illustrates the relations used to characterize
  effective multimodality under the fixed candidate budget; it does not by
  itself measure a budget sweep, scene-stratified $K_{\mathrm{eff}}$, effective
  capacity, or mode coverage.}
  \label{fig:latent_space_partition}
\end{figure}

\paragraph{Latent-space partition.}
The four selected scenes in \cref{fig:latent_space_partition} expose the two
inspection axes used by AMC-Drive: candidate geometry and the corresponding
predictive-space relation. They are sampled from the supplied sequence to show
variation across scenes while avoiding repeated near-identical pages.
The AC-JEPA coordinates are used here as a qualitative view of the relation
structure that defines $K_{\mathrm{eff}}$; the montage is not a candidate-count
ablation or a scene-stratified $K_{\mathrm{eff}}$ evaluation.

\paragraph{Action-conditioned responses.}
The source visualizations contain camera context, counterfactual trajectories,
and a left--right latent-response heatmap. We retain two cropped views so that
the displayed action buckets remain legible while making the evidence boundary
explicit: these are diagnostics of predictive response, not a quantitative test
of action identifiability.

\begin{figure}[p]
    \centering

    \includegraphics[
        width=0.96\textwidth,
        height=0.78\textheight,
        keepaspectratio
      ]{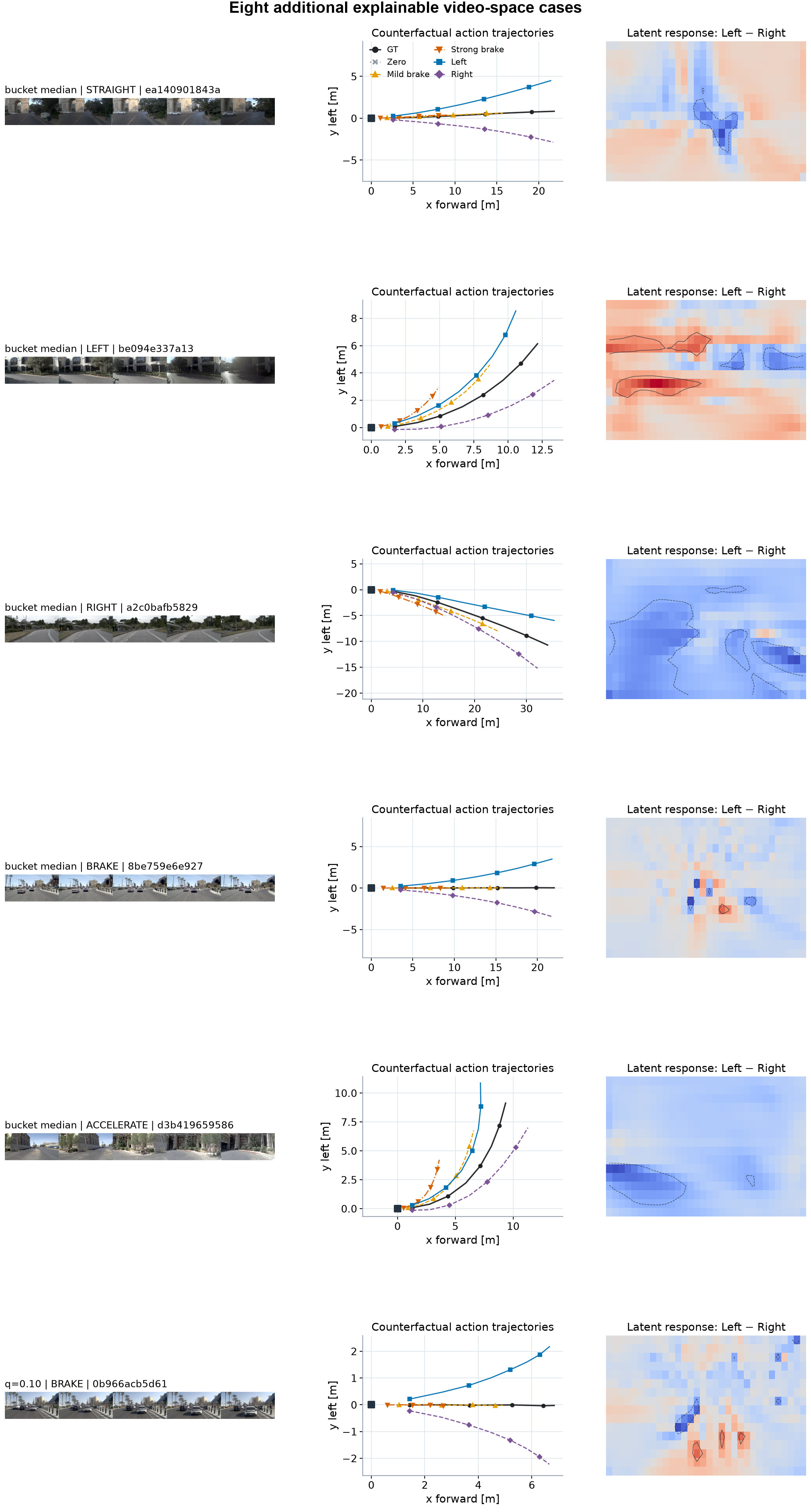}

    \caption{
    \textbf{Action-conditioned responses.}
    The upper rows show straight, left, right, and brake examples;
    the lower rows show accelerate and selected quantile examples.
    Each combines camera context, counterfactual trajectories,
    and a latent-response heatmap. Opposite steering examples show
    directional separation, while response extent and sign vary across scenes.
    }
    \label{fig:wm_explainable_cases}
\end{figure}

\subsection{Quantitative Action-Sensitivity Test}
\label{app:action_sensitivity}

The evidence package now includes a paired intervention on 256 fixed
validation samples. Replacing the ground-truth action by a fixed-seed
derangement increases future-latent MSE at every tested horizon, with positive
bootstrap intervals and consistent effects across the four behavior buckets.
The two panels below provide the most compact quantitative view; the formal
definitions and scope of the test follow afterward.

\begin{figure}[!htbp]
  \centering
  \begin{minipage}[t]{0.58\linewidth}\centering
    \includegraphics[width=\linewidth]{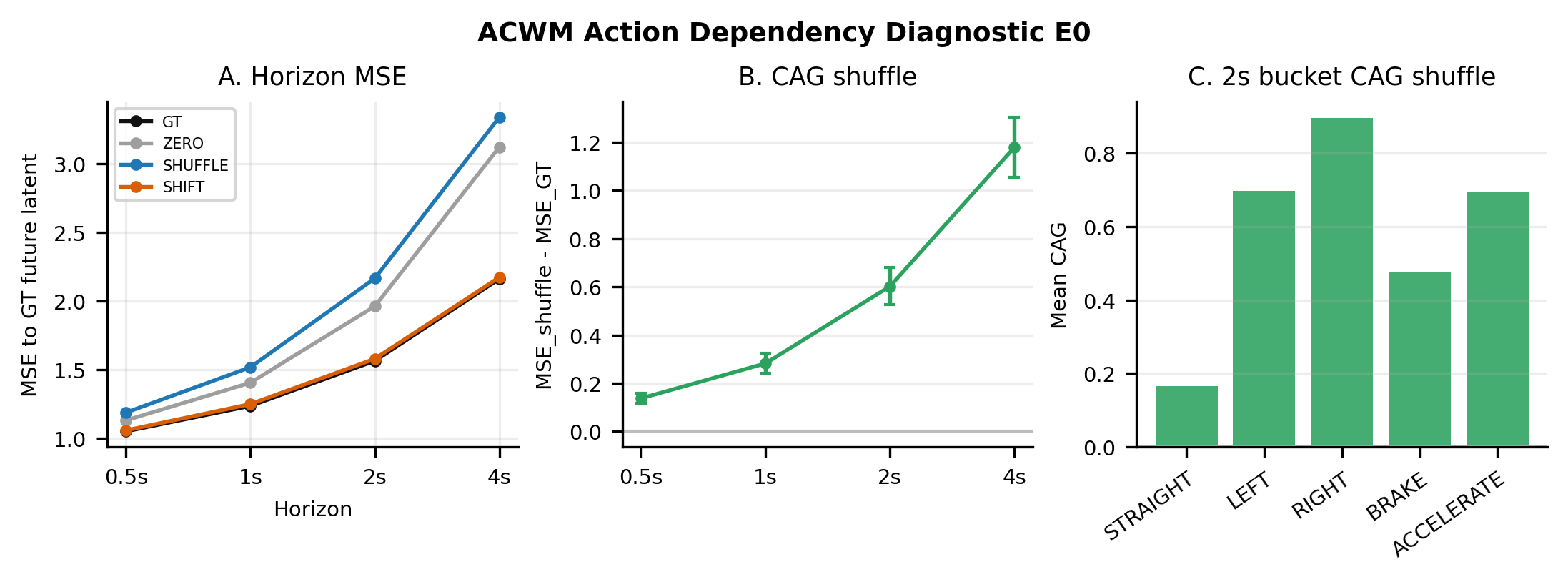}
  \end{minipage}\hfill
  \begin{minipage}[t]{0.39\linewidth}\centering
    \includegraphics[width=\linewidth]{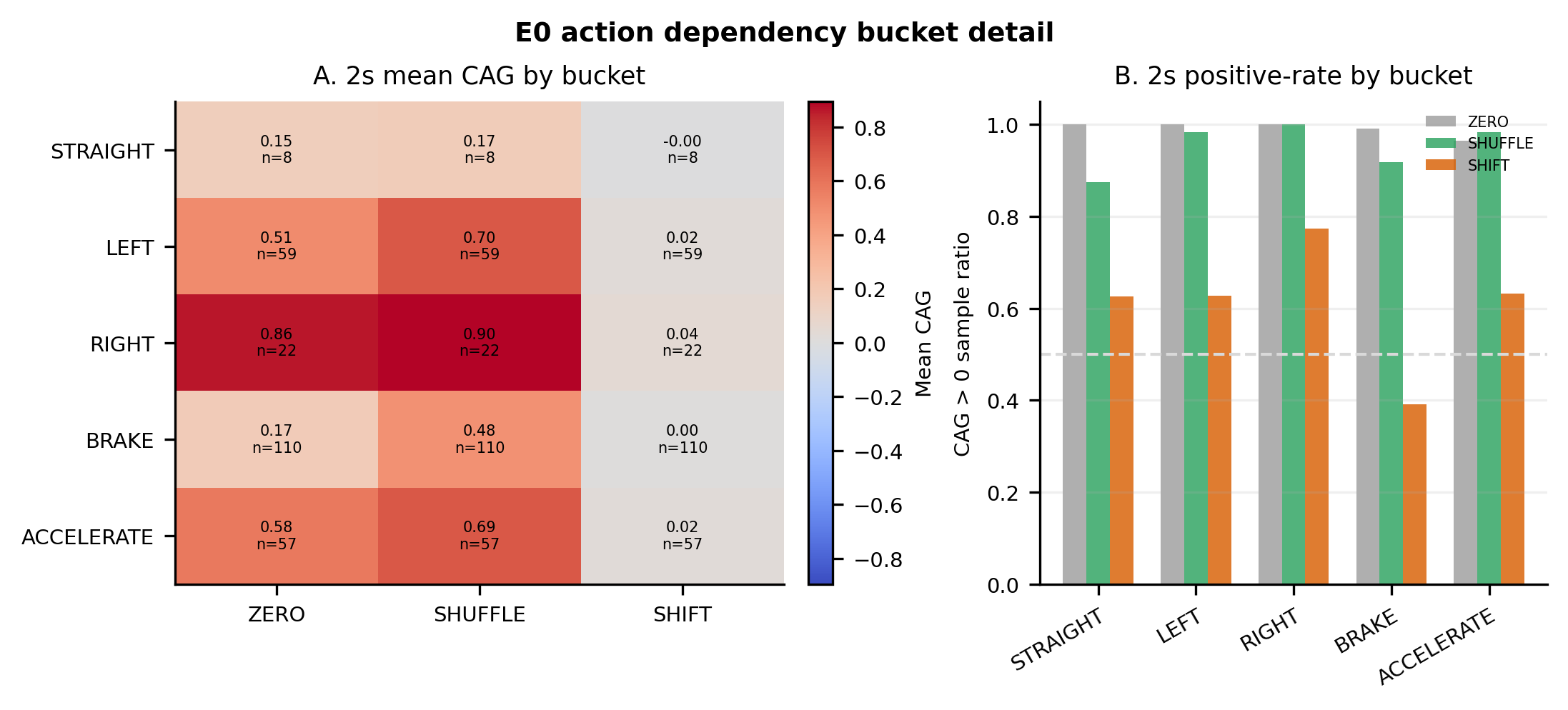}
  \end{minipage}
  \caption{\textbf{Paired action intervention (E0).} Fixed-history action
  replacement yields positive CAG at 0.5, 1, 2, and 4 s; the bucket panel
  shows the same direction for left, right, brake, and accelerate conditions.
  These results support action sensitivity of the evaluated checkpoint, not
  planner improvement or safety certification.}
  \label{fig:e0_action_sensitivity}
\end{figure}

For reference, the Shuffle CAG values are $0.136983$, $0.282042$,
$0.601921$, and $1.178901$ at 0.5, 1, 2, and 4 s, respectively; all reported
95\% bootstrap intervals are strictly positive. At 2 s, the bucket CAGs are
$0.696542$ (left), $0.895216$ (right), $0.476396$ (brake), and $0.694297$
(accelerate). We retain the distinction between this checkpoint-level
action-sensitivity result and the stronger, untested claim of end-to-end task
benefit.

The test follows the fixed-history intervention used for action-conditioned
JEPA models~\citep{assran2025vjepa2} and the displacement-based sensitivity
principle of Delta-JEPA~\citep{zhang2026deltajepa}.

\paragraph{Action-only intervention.}
For a scene history $h_n$, let $a_{n,c}$ denote the action sequence in bucket
$c\in\{\mathrm{straight},\mathrm{left},\mathrm{right},\mathrm{brake}\}$ and
let
\begin{equation}
    z_{n,c}=F_{\mathrm{AC}}(h_n,a_{n,c})
\end{equation}
be the frozen AC-JEPA predictive representation.  All representations are
pooled and normalized exactly as in the relation computation used by AMC.
The intervention keeps $h_n$ fixed and replaces only $a_{n,c}$; thus a
positive response cannot be explained by a changed camera observation.  The
primary latent-sensitivity score is
\begin{equation}
    \mathrm{AS}
    = \frac{1}{N}\sum_{n=1}^{N}
      \frac{1}{|\mathcal P_n|}
      \sum_{(c,c')\in\mathcal P_n}
      \frac{\lVert z_{n,c}-z_{n,c'}\rVert_2}
           {\lVert z_{n,c}\rVert_2+\varepsilon},
    \qquad
    \mathcal P_n=\{(c,c'):c\neq c'\}.
    \label{eq:action_sensitivity}
\end{equation}
We compare this value with an action-shuffled null, denoted
$\mathrm{AS}_{\mathrm{perm}}$, obtained by independently permuting actions
across histories while preserving the bucket frequencies.  We report the
paired difference $\Delta_{\mathrm{AS}}=\mathrm{AS}-\mathrm{AS}_{\mathrm{perm}}$
with a bootstrap 95\% confidence interval over scenes.

\paragraph{Action discrimination and counterfactual consistency.}
To test whether the response contains the action identity rather than only a
non-specific change, we fit centroids $\mu_c$ on the training scenes and
classify each held-out response by
\begin{equation}
    \widehat c_{n,c}
    = \arg\min_{c'}\lVert (z_{n,c}-z_{n,\mathrm{straight}})-\mu_{c'}\rVert_2 .
    \label{eq:action_discrimination}
\end{equation}
We report balanced accuracy for the four buckets and, separately, the
left--right balanced accuracy.  As a directional check, let
$v_{\mathrm{LR}}$ be the normalized difference between the left and right
training centroids and let $\ell_{n,c}$ be the signed terminal lateral
displacement of the corresponding candidate trajectory.  Counterfactual
consistency is
\begin{equation}
    \mathrm{CC}_{\mathrm{LR}}
    = \frac{1}{N_{\mathrm{LR}}}\sum_{(n,c)}
      \mathbf 1\!\left[
      \operatorname{sign}\big((z_{n,c}-z_{n,\mathrm{straight}})^\top
      v_{\mathrm{LR}}\big)
      = \operatorname{sign}(\ell_{n,c})
      \right].
    \label{eq:counterfactual_consistency}
\end{equation}
The centroids and $v_{\mathrm{LR}}$ are estimated on a disjoint set of
scenes, preventing the directional statistic from becoming a training-set
fit.  A Spearman correlation between pairwise latent distances and pairwise
trajectory distances is reported as a secondary, continuous alignment
measure.

\paragraph{Identifiability proposition.}
\begin{proposition}[Action sensitivity under fixed-history intervention]
\label{prop:action_sensitivity}
If $F_{\mathrm{AC}}(h,a)$ is independent of $a$ for every fixed history $h$,
then $\mathrm{AS}=0$; bucket discrimination is at chance under balanced
sampling; and the population action--trajectory distance correlation is zero.
Consequently, a positive paired sensitivity gap with a confidence interval
strictly above zero, together with above-chance held-out discrimination,
rejects the action-insensitive null for the tested action family.
\end{proposition}

\emph{Proof.}
Under the null, $z_{n,c}=z_{n,c'}$ for every pair, so every summand in
Eq.~\ref{eq:action_sensitivity} is zero.  The response then contains no
information about the bucket label, and the Bayes accuracy is the class prior,
which is chance for balanced buckets.  Since the latent distance is constant
with respect to the action while the trajectory distance varies with the
intervention, their population rank correlation is zero.  The contrapositive
gives the stated rejection criterion.  The proposition establishes functional
action dependence, not causal sufficiency or safety.

\paragraph{Evidence boundary for the current release.}
The E0 intervention is a quantitative checkpoint-level test: it supports
action sensitivity under fixed history, but it does not establish causal
sufficiency, planner improvement, or safety. The downstream ``+ Predictive''
backbone-substitution ablation (PDMS $91.7\!\rightarrow93.5$), specified in
Appendix~\ref{app:predictive_ablation}, remains a separate planning result and
is not used to inflate the action-sensitivity claim. The package also reports
an A3.2 frozen-path gradient smoke test; its zero AC-JEPA parameter gradients
confirm that the encoder is frozen, while its non-zero action-input gradient
confirms the differentiable path. We do not conflate either diagnostic with
an end-to-end task endpoint.

\subsection{Candidate Selection and Planner Comparison}
\label{app:selection_comparison}

\begin{figure}[!htbp]
  \centering
  \begin{minipage}[t]{0.49\linewidth}\centering
    \includegraphics[width=\linewidth,height=0.23\textheight,keepaspectratio]{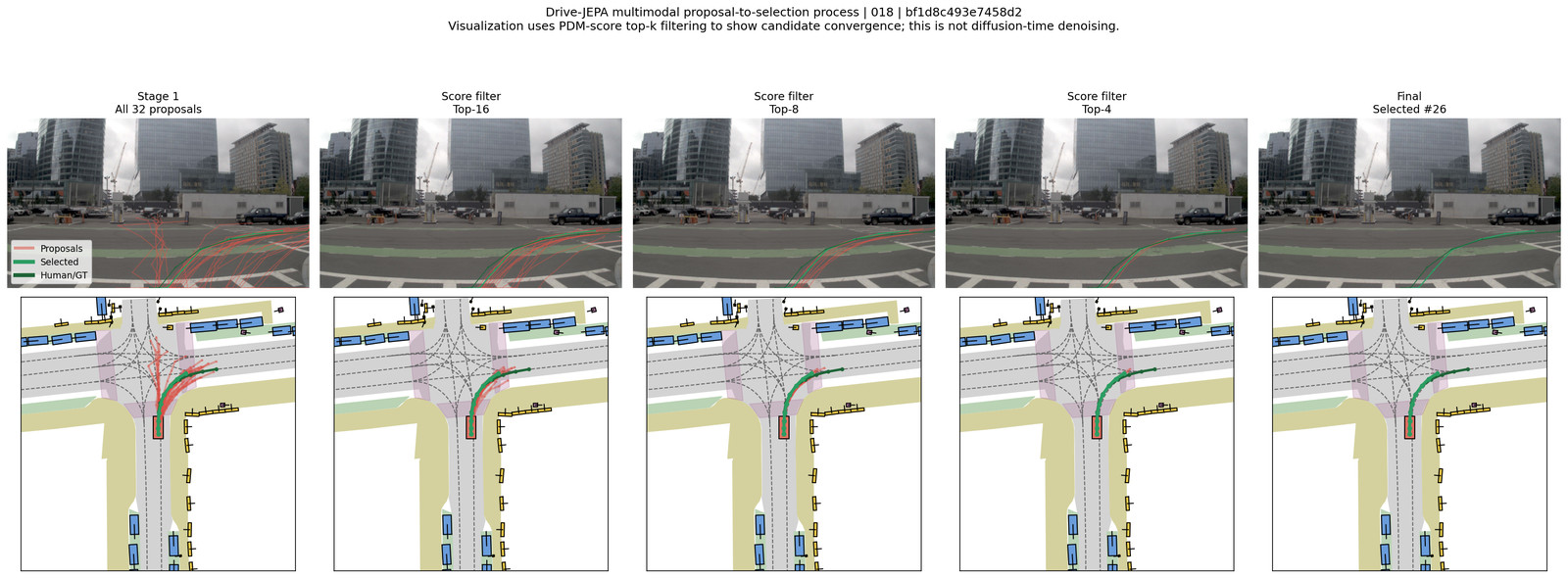}
  \end{minipage}\hfill
  \begin{minipage}[t]{0.49\linewidth}\centering
    \includegraphics[width=\linewidth,height=0.23\textheight,keepaspectratio]{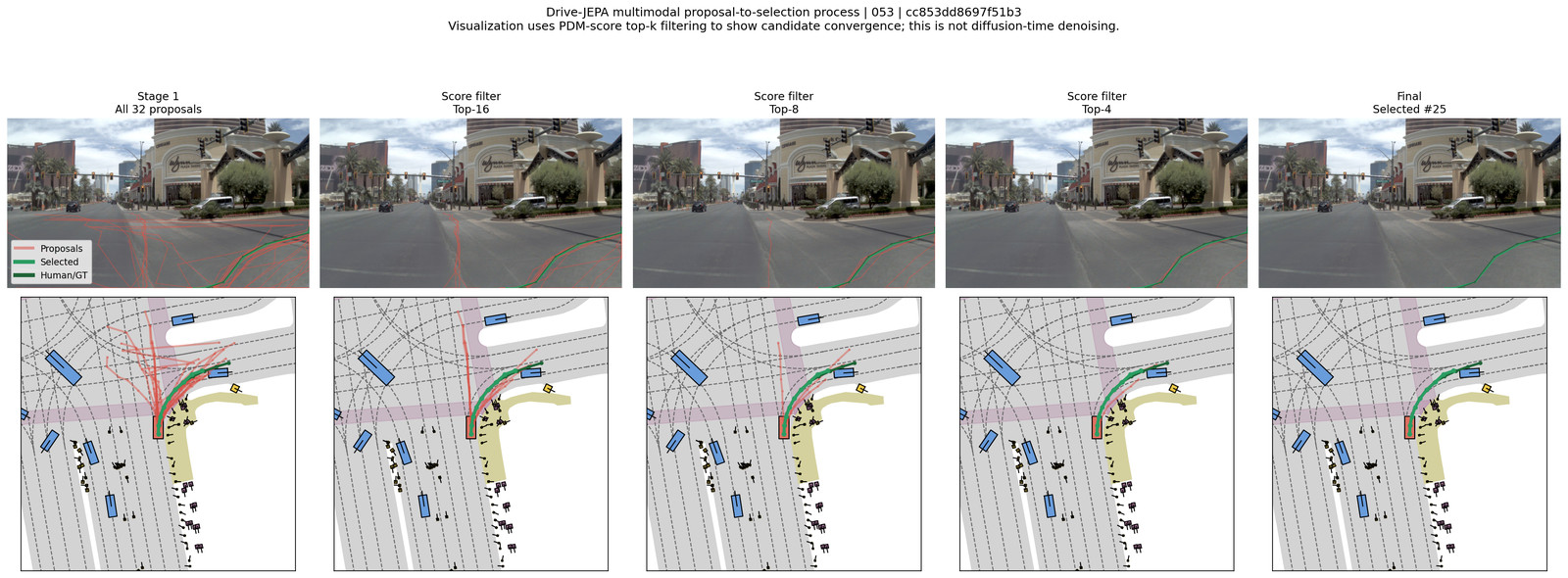}
  \end{minipage}

  \vspace{0.35em}

  \begin{minipage}[t]{0.49\linewidth}\centering
    \includegraphics[width=\linewidth,height=0.23\textheight,keepaspectratio]{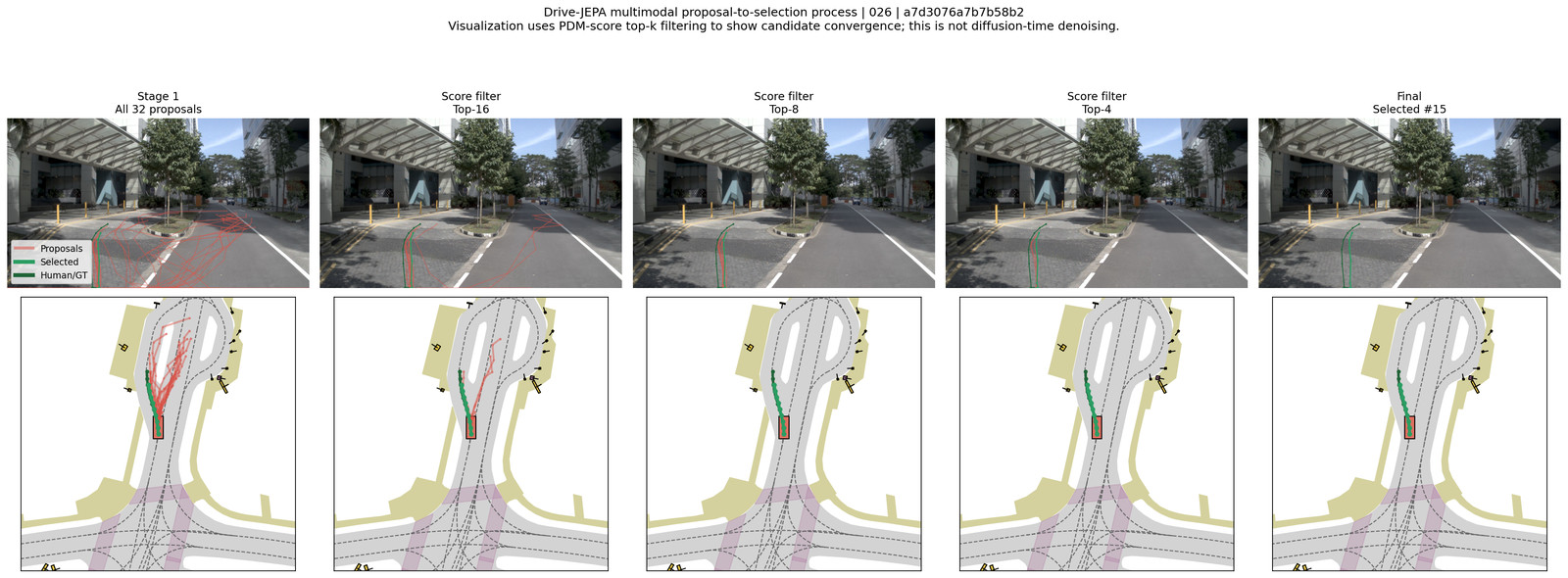}
  \end{minipage}\hfill
  \begin{minipage}[t]{0.49\linewidth}\centering
    \includegraphics[width=\linewidth,height=0.23\textheight,keepaspectratio]{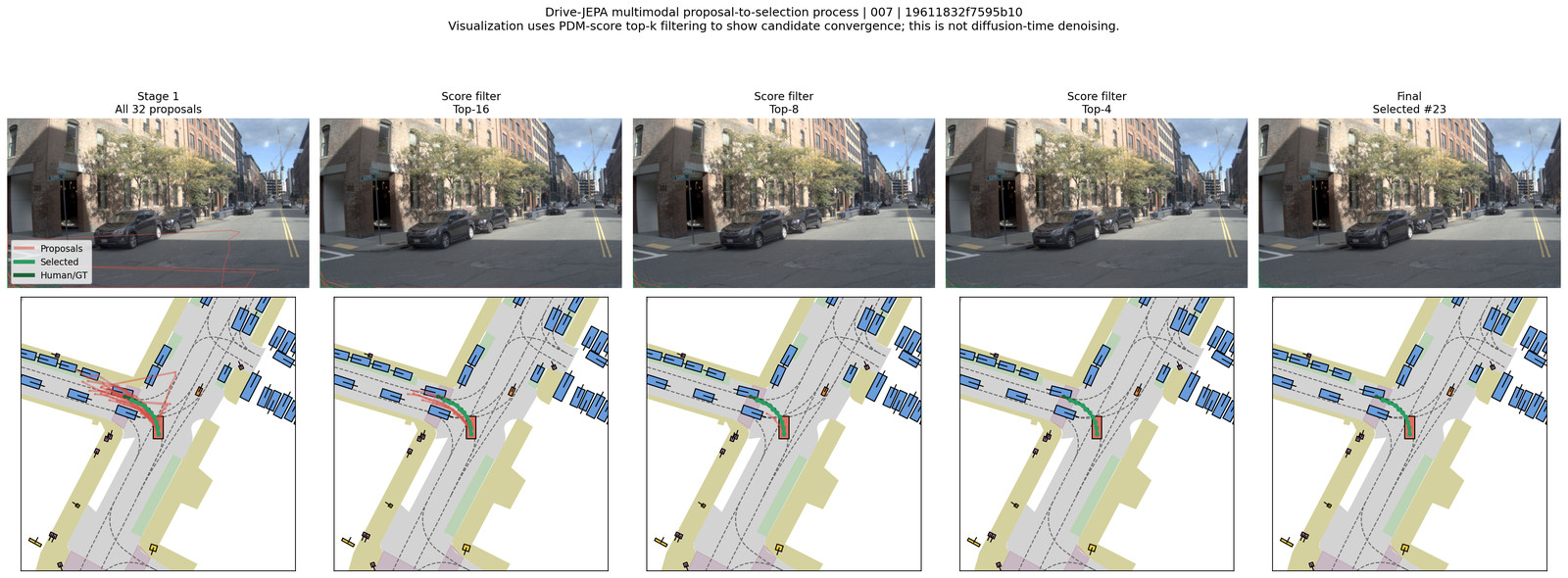}
  \end{minipage}
  \caption{\textbf{Selection from proposal banks.} Four scenes show the initial
  candidates, successive score-ranked subsets, and the final selected path.
  Colors and column order follow the legends embedded in the source panels.
  The montage illustrates selection behavior and does not establish a change
  in effective multimodal capacity.}
  \label{fig:app_selection}
\end{figure}

\paragraph{Proposal selection.}
The four examples in \cref{fig:app_selection} follow the same display order:
the full proposal bank, the top-ranked subsets, and the selected trajectory.
Grouping them in one montage makes clear that the operation is score-based
selection from an existing bank, rather than diffusion denoising or an
inference-time change to the candidate count.

\begin{figure}[!htbp]
  \centering
  \includegraphics[width=0.8\linewidth,height=0.52\textheight,keepaspectratio]{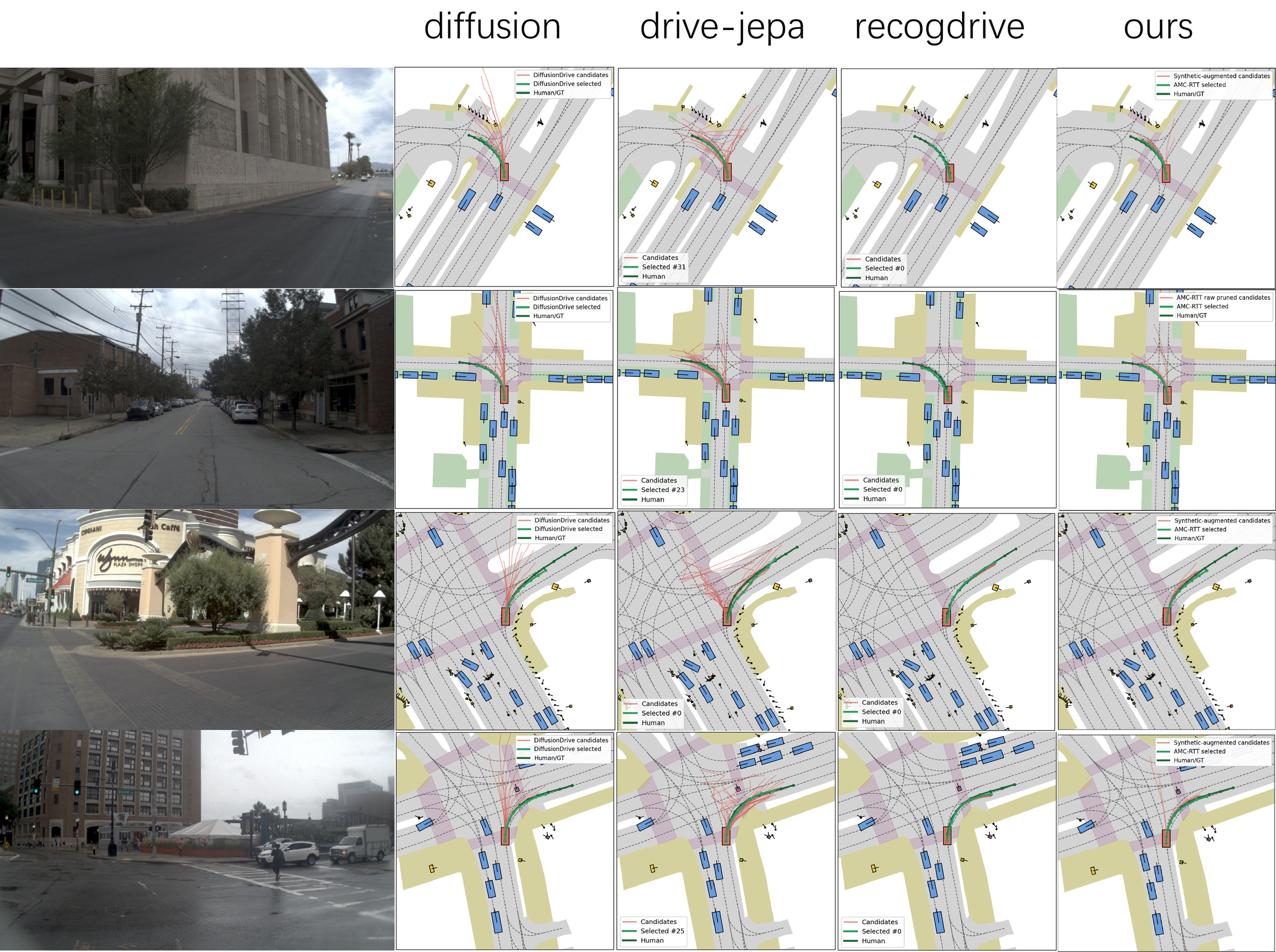}
  \caption{\textbf{Candidate-bank comparison.} The rows compare candidate
  spread, overlap, and selected trajectories across planners under a shared
  camera observation and bird's-eye layout. The quantitative comparison remains
  the main-text table.}
  \label{fig:app_methods}
\end{figure}

\paragraph{Planner comparison.}
\Cref{fig:app_methods} places the camera observation beside candidate-bank
visualizations for the compared planners. Candidate spread and overlap can be
inspected under the same displayed road layout; no numerical ranking is inferred
from these selected examples.

\FloatBarrier
\subsection{Representative Scenes and Closed-Loop Plans}
\label{app:scenes_closed_loop}

\paragraph{Driving scenes.}
We retain four scene groups that cover an unprotected turn, a signalized
intersection, a merge or ramp, and dense interaction. Each group contains
multiple camera views with trajectory overlays. The overlays expose local road
geometry and visibility; they do not establish maneuver completion or collision
avoidance from a single frame. In relation to the title, these groups are
qualitative probes of how the AC-JEPA predictive-space organization can differ
across contexts while $K$ remains fixed; no category-wise $K_{\mathrm{eff}}$
summary is claimed here.

\begin{figure}[!htbp]
  \centering
  \begin{minipage}[t]{0.49\linewidth}\centering
    \includegraphics[width=\linewidth,height=0.20\textheight,keepaspectratio]{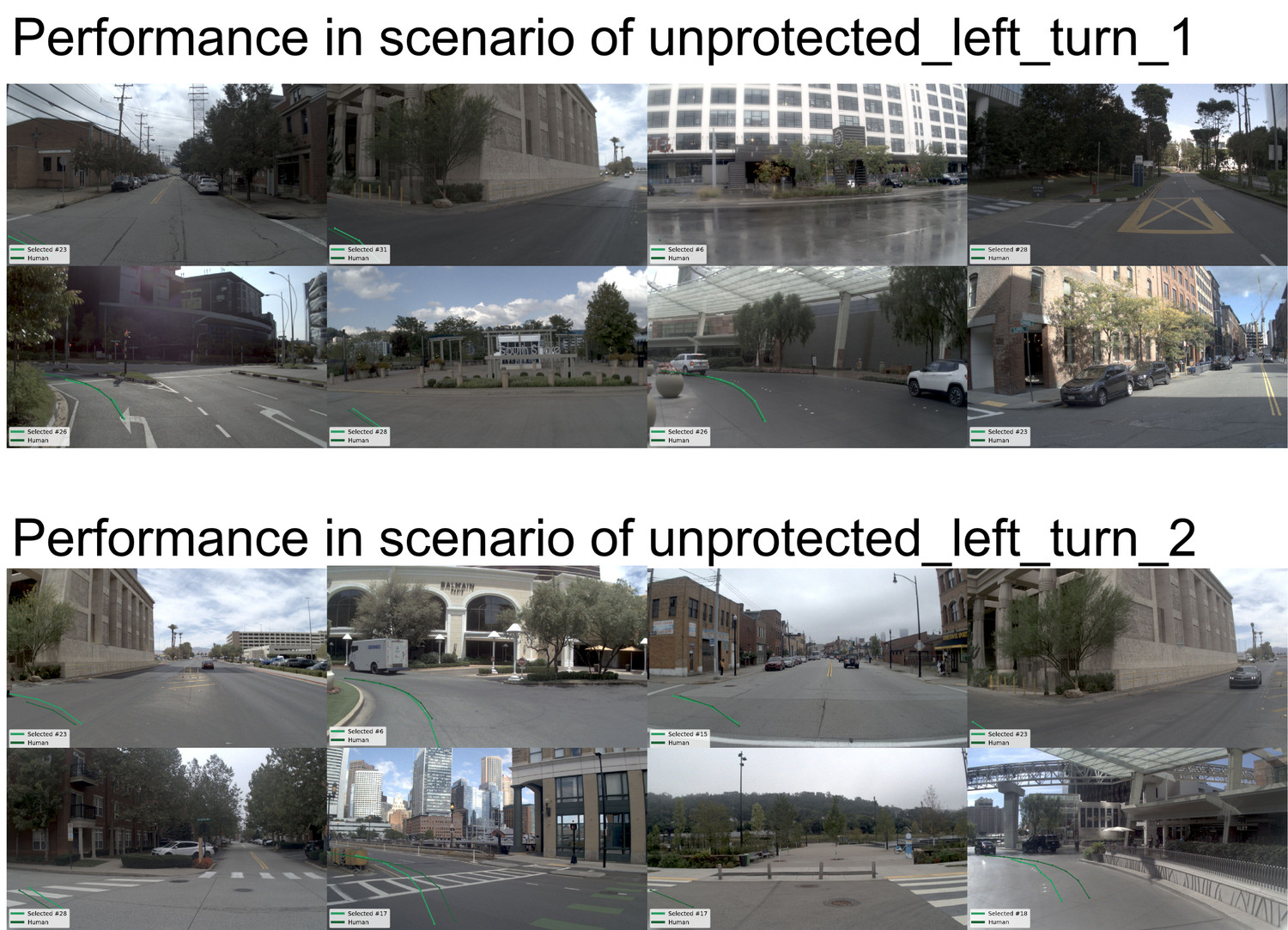}
  \end{minipage}\hfill
  \begin{minipage}[t]{0.49\linewidth}\centering
    \includegraphics[width=\linewidth,height=0.20\textheight,keepaspectratio]{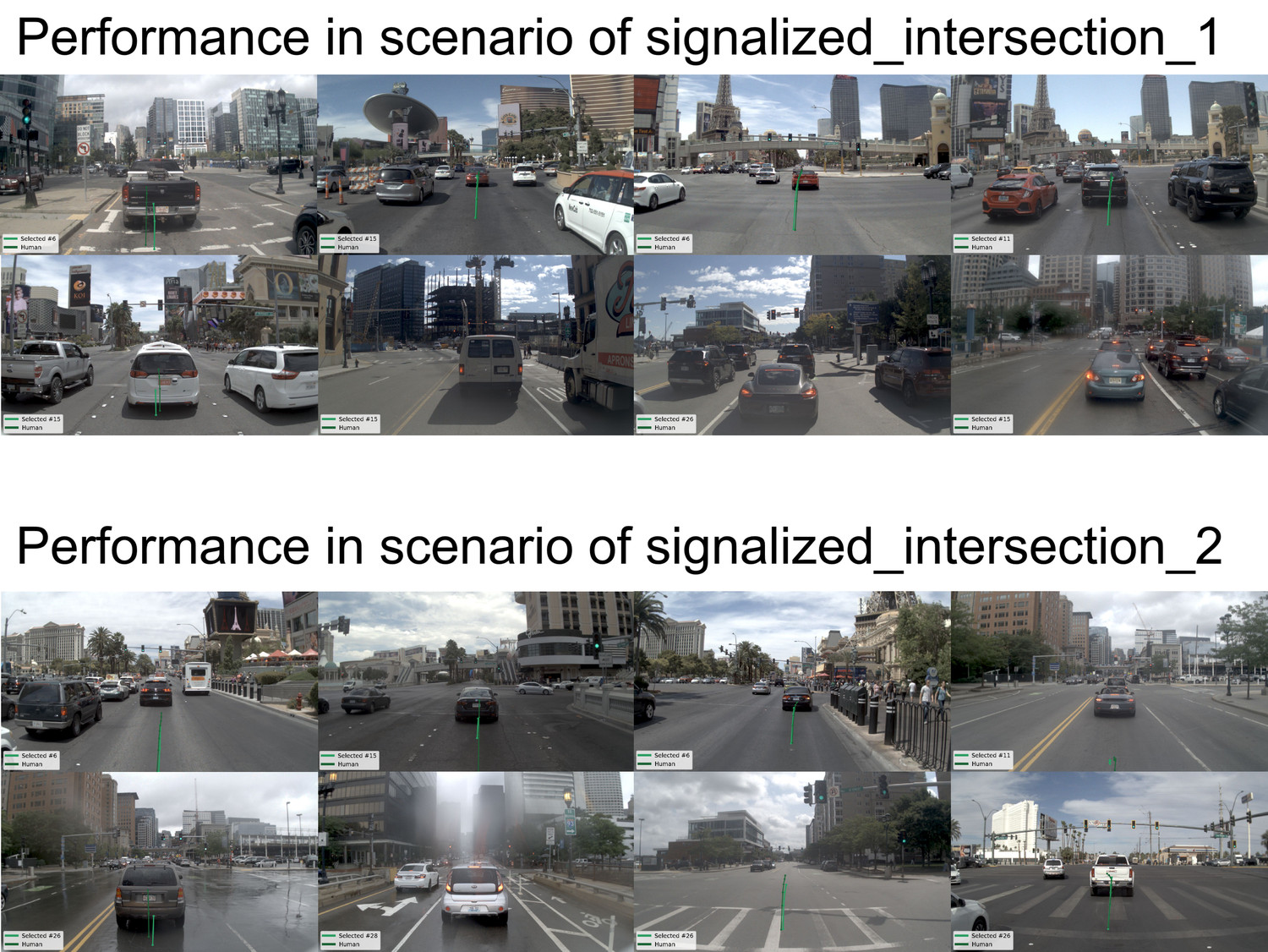}
  \end{minipage}

  \vspace{0.35em}

  \begin{minipage}[t]{0.49\linewidth}\centering
    \includegraphics[width=\linewidth,height=0.20\textheight,keepaspectratio]{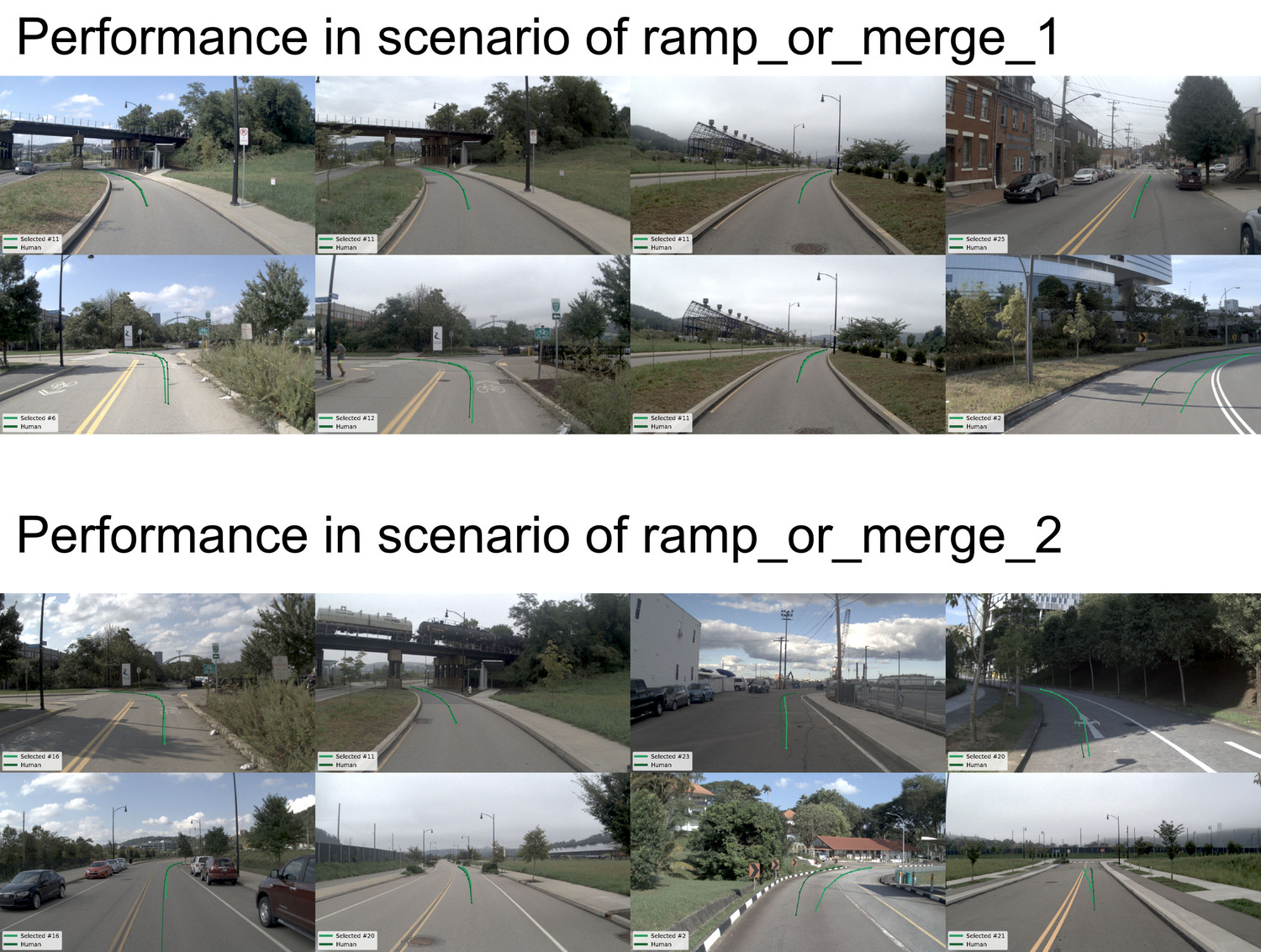}
  \end{minipage}\hfill
  \begin{minipage}[t]{0.49\linewidth}\centering
    \includegraphics[width=\linewidth,height=0.20\textheight,keepaspectratio]{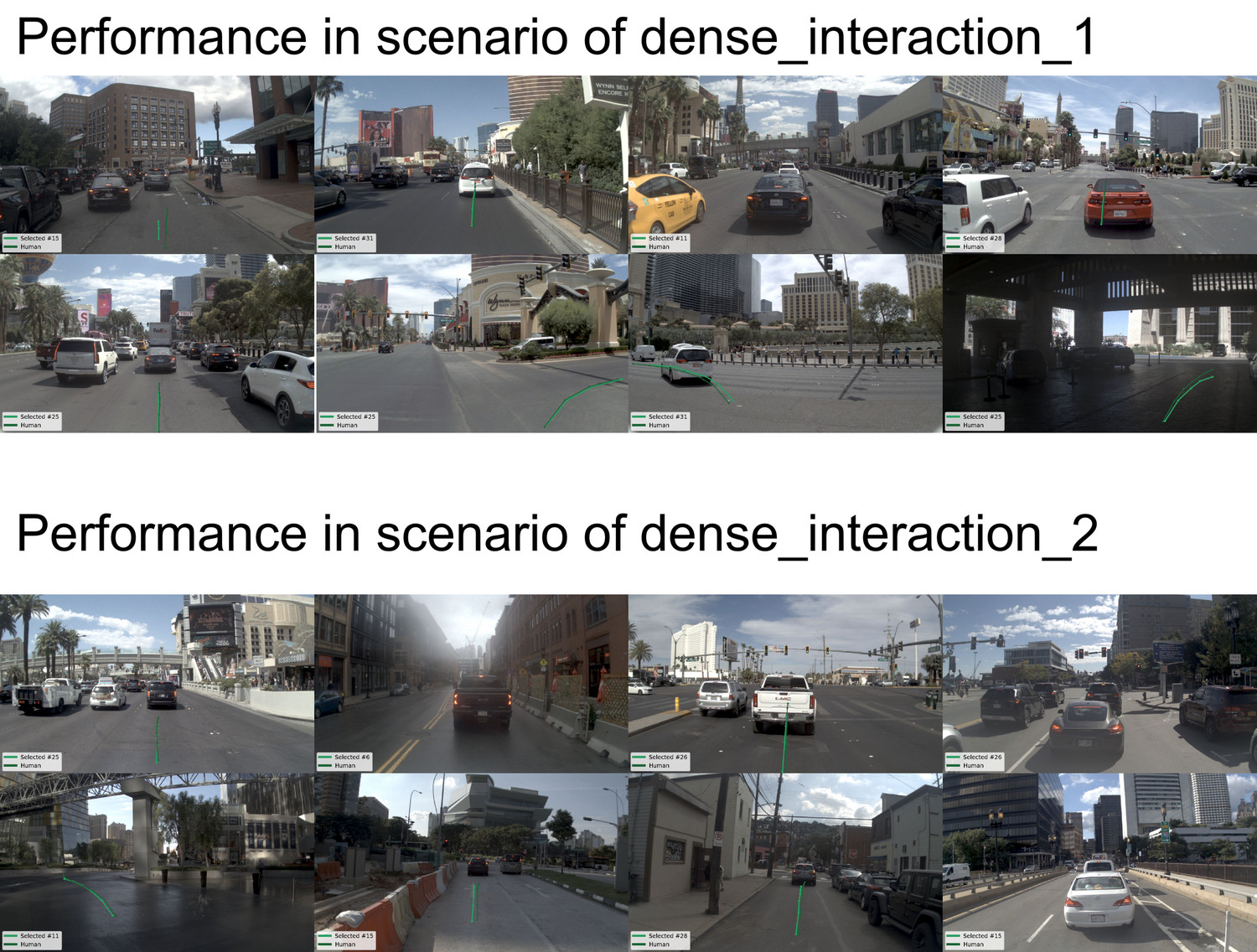}
  \end{minipage}
  \caption{\textbf{Representative driving scenes.} The four groups cover
  junction turning, signalized traffic, merge geometry, and dense interaction.
  Green overlays show the paths present in the source assets; the panels provide
  scene context rather than an additional safety evaluation.}
  \label{fig:app_scenes}
\end{figure}

\paragraph{Closed-loop plans.}
The six selected closed-loop panels cover varied roadway context, nearby agents,
and changing illumination. They are grouped in two montages to keep each panel
large enough for inspection while avoiding a separate float for every sample.

\begin{figure}[!htbp]
  \centering
  \begin{minipage}[t]{0.49\linewidth}\centering
    \includegraphics[width=\linewidth,height=0.20\textheight,keepaspectratio]{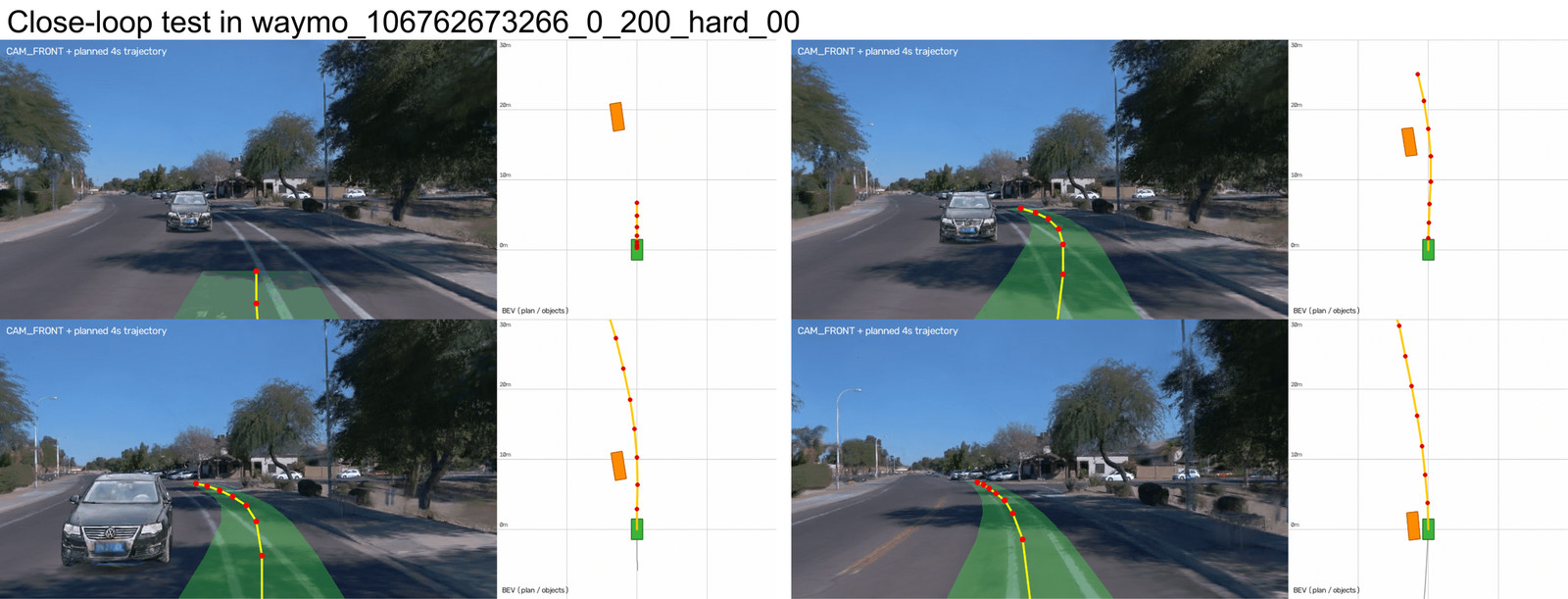}
  \end{minipage}\hfill
  \begin{minipage}[t]{0.49\linewidth}\centering
    \includegraphics[width=\linewidth,height=0.20\textheight,keepaspectratio]{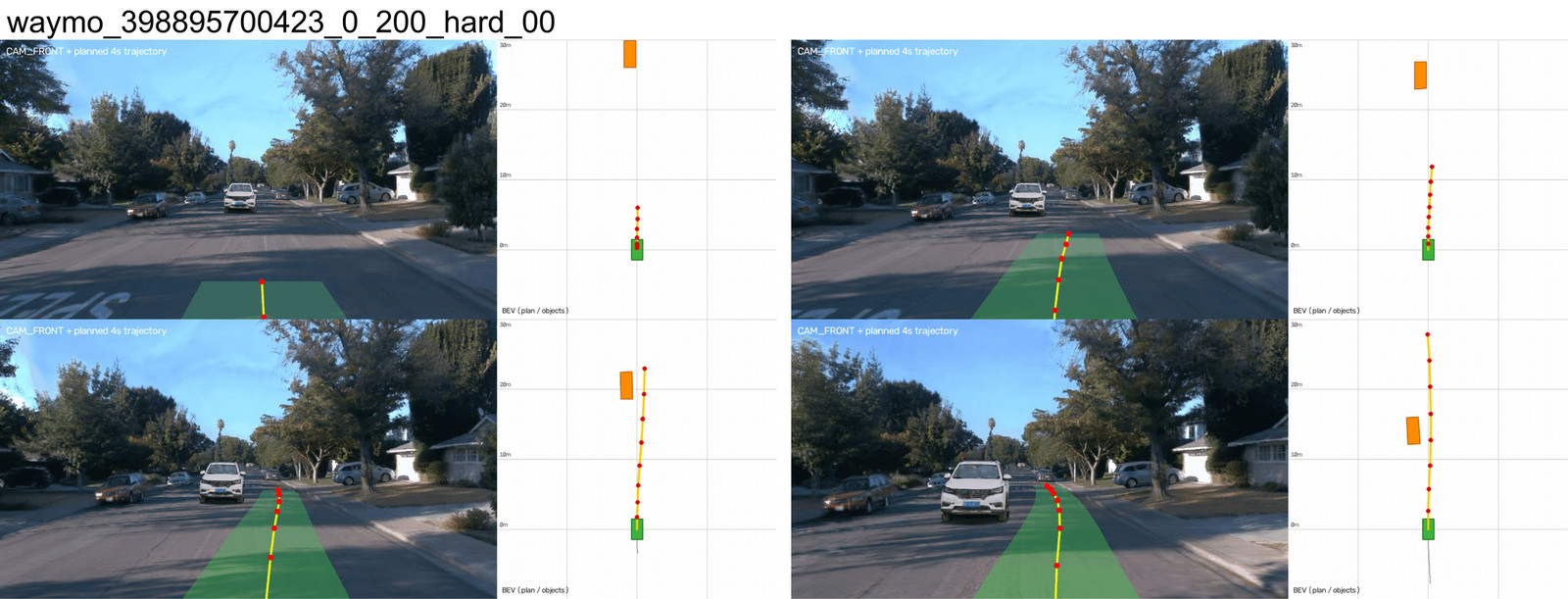}
  \end{minipage}

  \vspace{0.35em}

  \begin{minipage}[t]{0.49\linewidth}\centering
    \includegraphics[width=\linewidth,height=0.20\textheight,keepaspectratio]{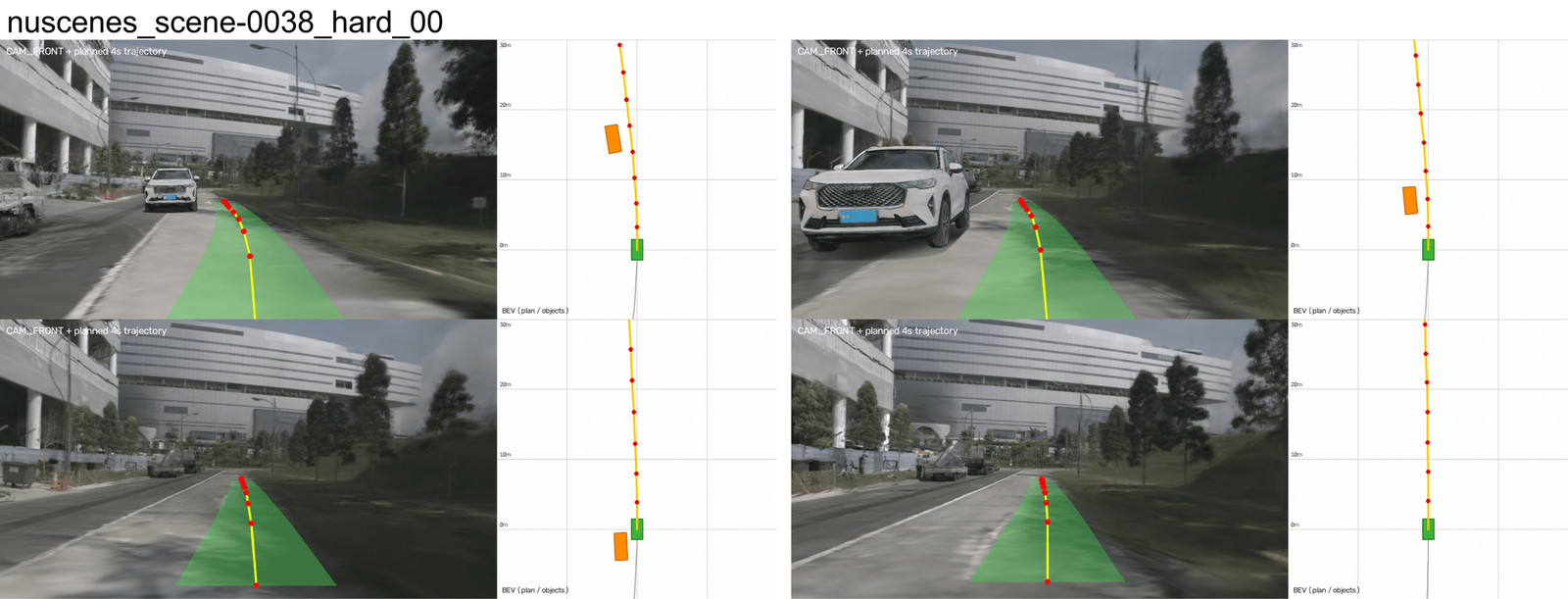}
  \end{minipage}\hfill
  \begin{minipage}[t]{0.49\linewidth}\centering
    \includegraphics[width=\linewidth,height=0.20\textheight,keepaspectratio]{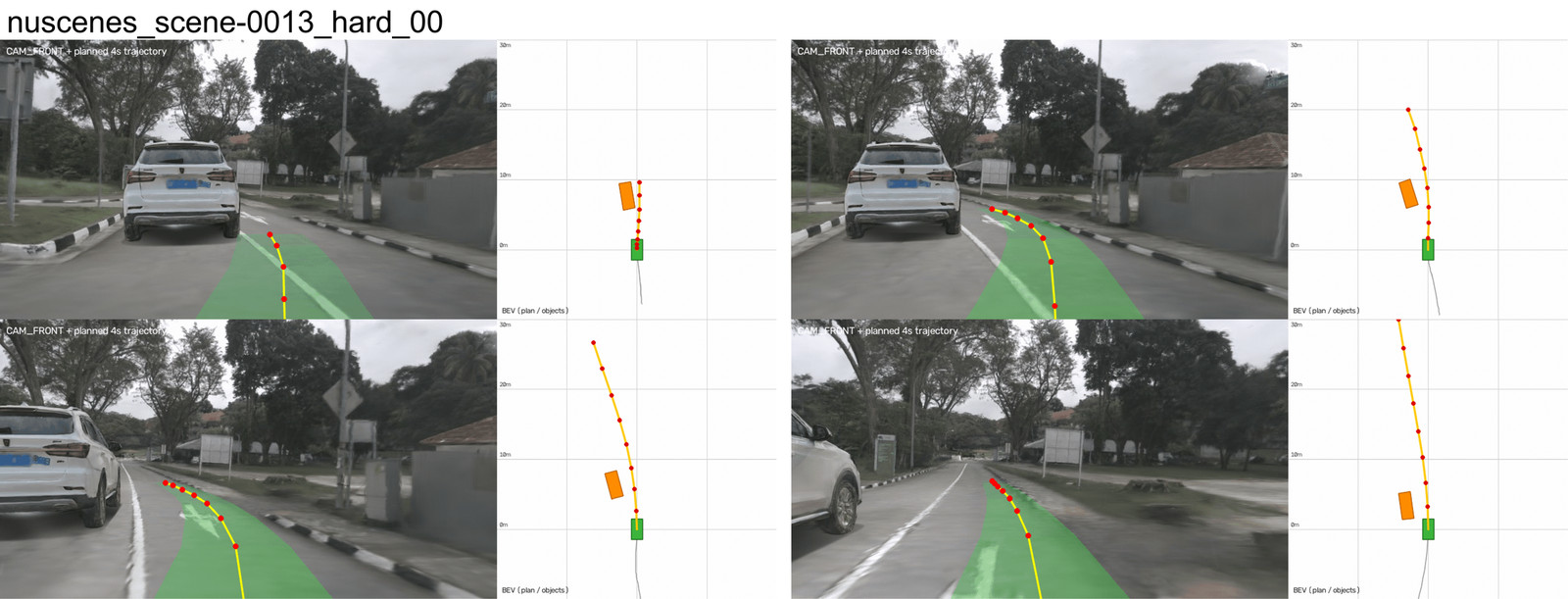}
  \end{minipage}
  \caption{\textbf{Closed-loop plans, montage I.} Camera and bird's-eye panels
  show local plan geometry under different roadway and traffic contexts. The
  montage supports visual inspection of rendered plans, not a replacement for
  the HUGSIM metrics.}
  \label{fig:app_closeloop_a}
\end{figure}

\begin{figure}[!htbp]
  \centering
  \begin{minipage}[t]{0.49\linewidth}\centering
    \includegraphics[width=\linewidth,height=0.20\textheight,keepaspectratio]{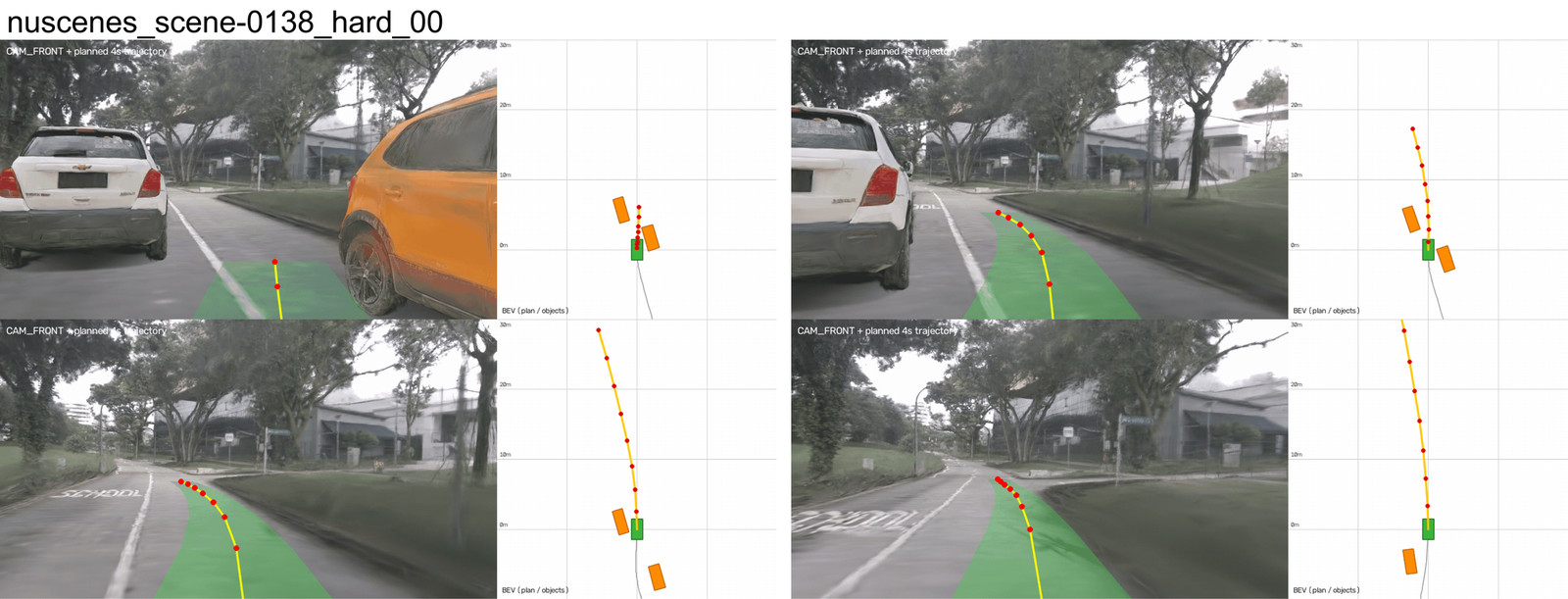}
  \end{minipage}\hfill
  \begin{minipage}[t]{0.49\linewidth}\centering
    \includegraphics[width=\linewidth,height=0.20\textheight,keepaspectratio]{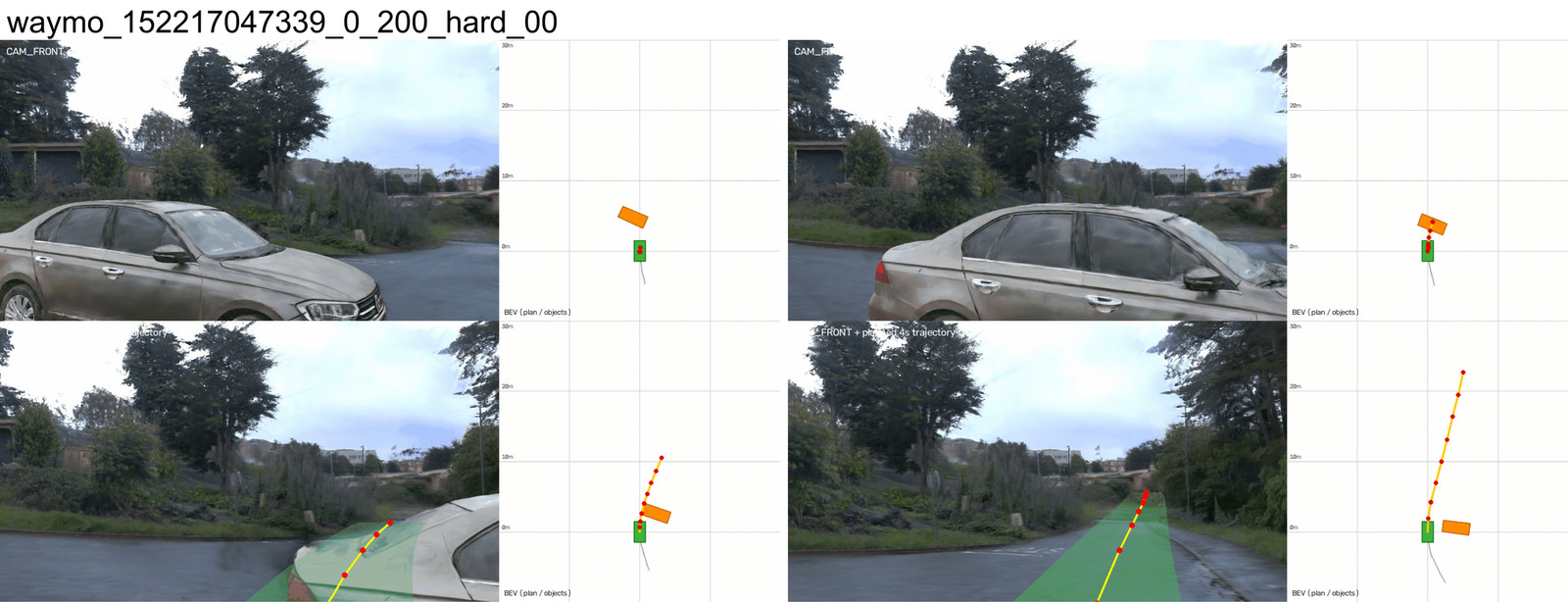}
  \end{minipage}
  \caption{\textbf{Closed-loop plans, montage II.} Additional scenes show
  planned waypoints relative to nearby objects and road boundaries. These
  renderings document qualitative behavior only; success and collision claims
  require the matched closed-loop evaluation.}
  \label{fig:app_closeloop_b}
\end{figure}

\FloatBarrier

\end{document}